\documentclass[11pt, letterpaper, logo, copyright, nonumbering]{seed}

\usepackage{natbib}

\usepackage{CJKutf8}

\usepackage{xargs}  

\usepackage{todonotes}  

\usepackage{multirow}

\usepackage{cleveref}

\usepackage{amsmath}
\usepackage{dsfont}

\usepackage{subcaption}

\usepackage{svg}

\usepackage{listings}
\lstdefinestyle{mystyle}{
    commentstyle=\color{green!50!black},
    numberstyle=\tiny\color{gray},
    basicstyle=\ttfamily\fontsize{5}{5}\selectfont,
    breakatwhitespace=false,         
    breaklines=true,                 
    captionpos=b,                    
    keepspaces=true,                 
    numbers=none,                    
    numbersep=5pt,                  
    showspaces=false,                
    showstringspaces=false,
    showtabs=false,                  
    tabsize=2
}

\lstdefinestyle{mystyle2}{
    commentstyle=\color{green!50!black},
    numberstyle=\tiny\color{gray},
    basicstyle=\centering\ttfamily\fontsize{7}{7}\selectfont,
    breakatwhitespace=false,         
    breaklines=true,                 
    captionpos=b,                    
    keepspaces=true,                 
    numbers=none,                    
    numbersep=5pt,                  
    showspaces=false,                
    showstringspaces=false,
    showtabs=false,                  
    tabsize=2
}

\usepackage{makecell}
\usepackage{tabularx}
\usepackage{array}
\usepackage{xspace}
\usepackage{wrapfig}
\usepackage{subcaption}
\usepackage{multirow}
\usepackage{multicol}
\usepackage{setspace}
\usepackage[frozencache,cachedir=minted-cache]{minted}
\usepackage{booktabs}
\usepackage{inconsolata}
\usepackage{float}
\usepackage{graphicx}
\usepackage{url}
\usepackage{adjustbox}

\definecolor{mydarkblue}{rgb}{0,0.08,0.45}
\definecolor{themecolor}{HTML}{A468FE}

\hypersetup{
 colorlinks=true,
 citecolor=mydarkblue,
 linkcolor=mydarkblue,
 urlcolor=mydarkblue
}

\newcommand{\ourmodel}{Seed-Coder\xspace}
\newcommand{\graybg}{\rowcolor{themecolor!15}}
\newcommand{\grayt}{\color{black}}
\newcommand{\fst}[1]{\textbf{#1}}

\renewcommand{\today}{}

\newcommand{\creditsectionheader}[1]{\parbox{\columnwidth}{\centering \textbf{\normalsize #1}}\\}
\newcommand{\creditlistheader}[1]{\noindent\textbf{#1}\\}

\newcommand{\corecontributor}[1]{#1\\}

\newcommand{\huggingface}{\raisebox{-1.5pt}{\includegraphics[height=1.0em]{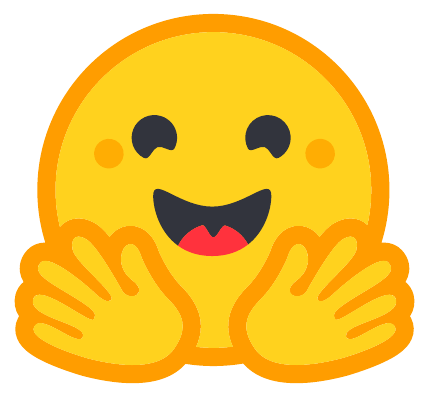}}\xspace}
\newcommand{\github}{\raisebox{-1.5pt}{\includegraphics[height=1.0em]{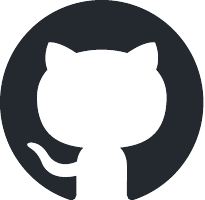}}\xspace}
\newcommand{\website}{\raisebox{-1.5pt}{\includegraphics[height=1.0em]{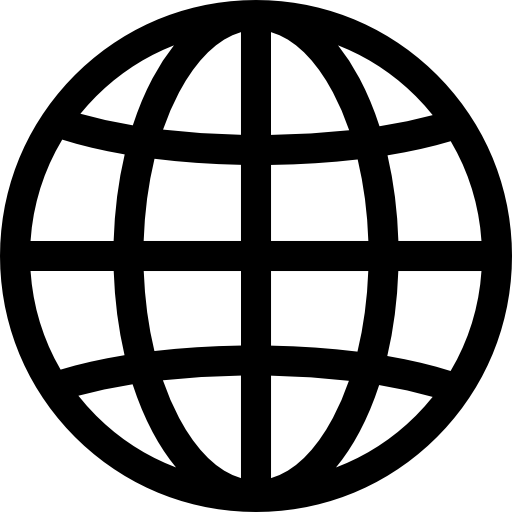}}\xspace}
\newcommand{\doubao}{\raisebox{-1.5pt}{\includegraphics[height=1.0em]{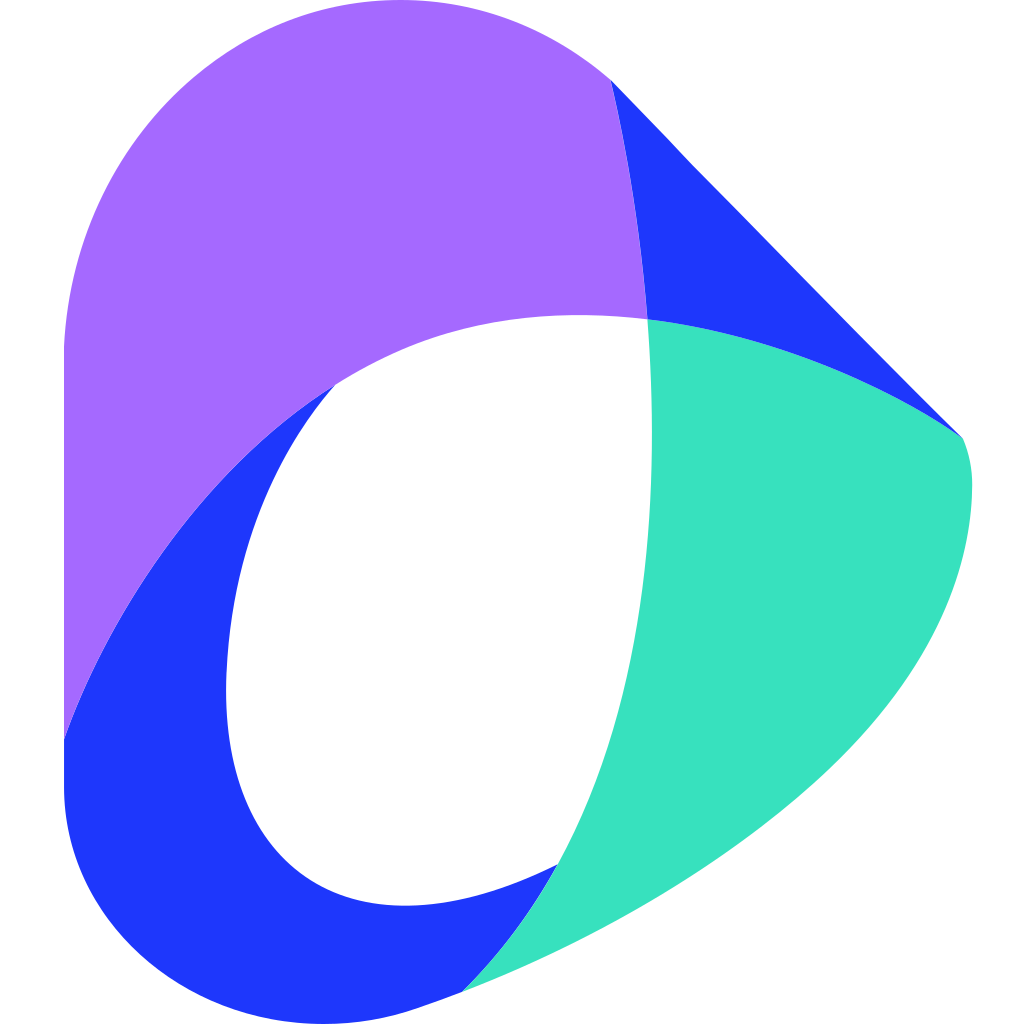}}\xspace}

\newcommandx{\info}[2][1=]{\todo[linecolor=red,backgroundcolor=red!25,bordercolor=red,#1]{#2}}

\title{
\centering \doubao \ourmodel: Let the Code Model Curate Data for Itself}

\author[*]{\textbf{ByteDance Seed}
\\
\begin{center}

\ \ \ \website \ \ \normalfont{\href{https://bytedance-seed-coder.github.io}{Homepage}} \ \ \ \ \quad \github \ \ \normalfont{\href{https://github.com/ByteDance-Seed/Seed-Coder}{GitHub}} \ \ \quad \huggingface \ \ \normalfont{\href{https://huggingface.co/collections/ByteDance-Seed/seed-coder-680de32c15ead6555c75b0e4}{Hugging Face}}

\end{center}
\vspace{-20pt}
}

\begin{abstract}
\vspace{-10pt}
Code data in large language model (LLM) pretraining is recognized crucial not only for code-related tasks but also for enhancing general intelligence of LLMs. Current open-source LLMs often heavily rely on human effort to produce their code pretraining data, such as employing hand-crafted filtering rules tailored to individual programming languages, or using human-annotated data to train quality filters. However, these approaches are inherently limited in scalability, prone to subjective biases, and costly to extend and maintain across diverse programming languages. To address these challenges, we introduce \ourmodel, a series of open-source LLMs comprising base, instruct and reasoning models of 8B size, minimizing human involvement in data construction. Our code pretraining data is produced by a model-centric data pipeline, which predominantly leverages LLMs for scoring and filtering code data. The instruct model is further trained via supervised fine-tuning and preference optimization, and the reasoning model leverages Long-Chain-of-Thought (LongCoT) reinforcement learning to improve multi-step code reasoning. \ourmodel achieves state-of-the-art results among open-source models of similar size and even surpasses some much larger models, demonstrating superior performance in code generation, code completion, code editing, code reasoning, and software engineering tasks.
\end{abstract}

\begin{document}

\maketitle
\begin{figure}[H]
    \vspace{5pt}
    \centering
    \includegraphics[width=\linewidth]{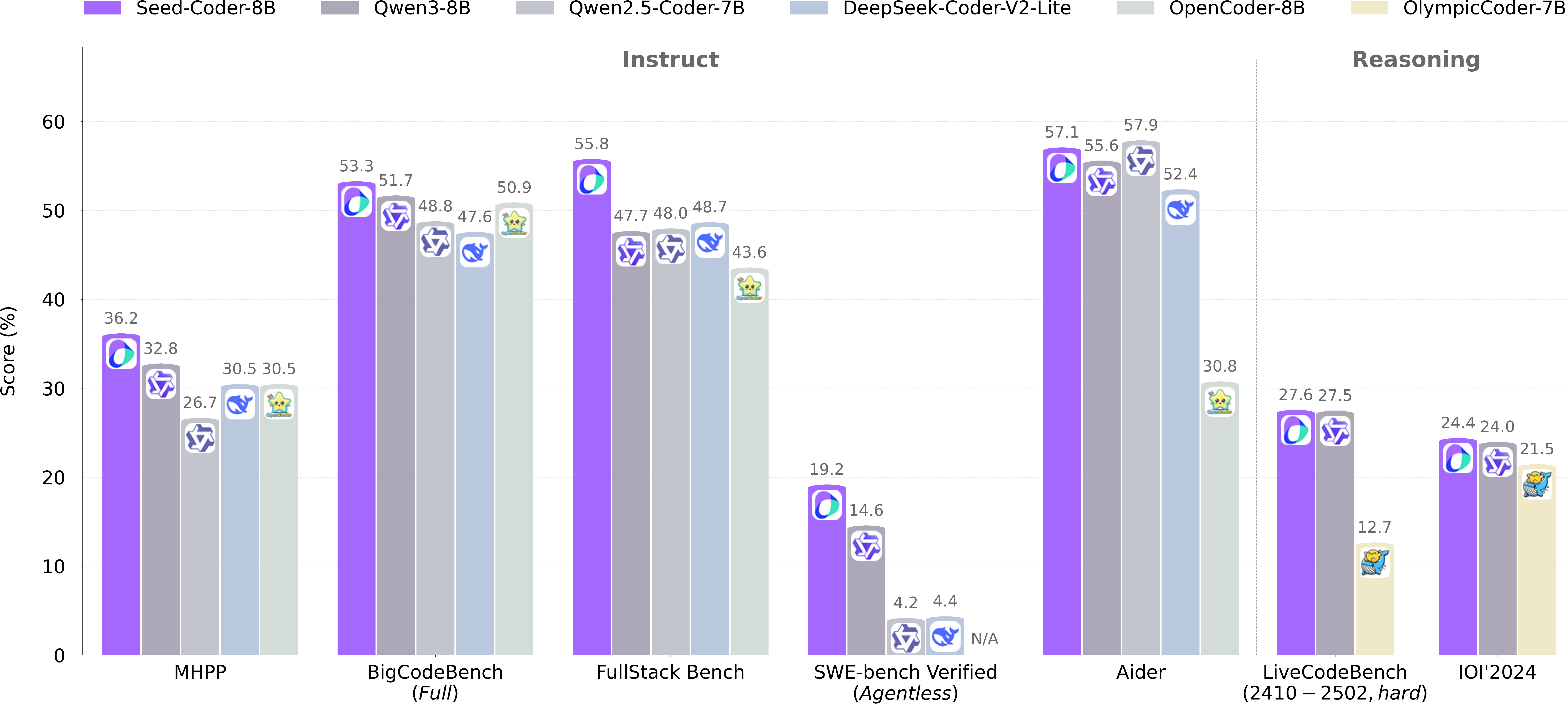}
    \label{fig:teaser}
    \vspace{-17pt}
    \caption{Benchmark performance of instruct and reasoning variants of \ourmodel-8B.}
\end{figure}
\let\oldthefootnote\thefootnote
\renewcommand*{\thefootnote}{\fnsymbol{footnote}}
\let\thefootnote\oldthefootnote

\newpage

\tableofcontents

\newpage

\section{Introduction}

Large language models (LLMs) have achieved superior performance in a wide range of coding tasks, showing their potential to revolutionize the entire ecosystem of software development. Current state-of-the-art models, such as Claude 3.7 Sonnet~\citep{anthropic2025claude} and OpenAI o3~\citep{openai2025o3}, have demonstrated unprecedented capability in a wide range of code-related tasks, including code generation, code explanation, code debugging, code editing, and real-world software engineering tasks. These models can largely enhance developers' productivity and provide substantial support for the software industry, showing the potential to gradually automate the process of software development. However, these cutting-edge models remain proprietary and have disclosed very little about their training data.

\noindent State-of-the-art open-source LLMs, such as DeepSeek-R1~\citep{deepseekai2025deepseekr1} and Qwen2.5-Coder~\citep{qwen25coder}, have invigorated the research community by offering self-deployable models that demonstrate competitive performance on coding tasks. While their technical reports provide valuable insights into both pretraining and post-training techniques, many offer only a high-level overview of their data processing methodologies. Among these ideas, one broad consensus for data quality control lies in that leveraging human expertise to curate and refine datasets is an intuitive and effective approach to enhancing the quality of code pretraining data. For example, DeepSeek-Coder~\citep{deepseek_coder} and DeepSeek-Coder-V2~\citep{deepseekai2024deepseekv2strongeconomicalefficient} apply a set of filtering rules as the initial data processing step, following the filtering rules used in StarCoder~\citep{li2023starcoder}. Qwen2.5-Coder~\citep{qwen25coder} also utilizes a series of rule-based filtering methods similar to DeepSeek-Coder. Moreover, OpenCoder~\citep{huang2024opencoder} incorporates over 130 hand-crafted filtering rules with customized weights in their pretraining data pipeline.

\noindent However, hand-crafted rules are prone to conflicts among themselves and may incur high costs to extend and maintain across diverse programming languages. Existing human-centric data filtering approaches that heavily rely on hand-crafted rules often introduce subjective biases and inherently lack scalability. We believe that these limitations align with insights from the widely cited essay ``The Bitter Lesson''~\citep{sutton2019bitter}: AI researchers often favor human-centric methods due to their short-term advantages, yet these approaches inevitably plateau and even inhibit long-term progress; the breakthrough progress eventually arrives by the opposing approach via scaling computation and data. In the context of code pretraining data, the human-centric approaches appear particularly favorable, since many AI researchers are skilled programmers themselves and feel confident in evaluating code quality. Nevertheless, human-centric methods would ultimately prove restrictive and tend to impede advancements of code LLMs in the long run.

\noindent In this report, we present a model-centric data pipeline that minimizes human involvement in constructing code pretraining data. Our data pipeline predominantly leverages LLMs, instead of hand-crafted rules, for scoring and filtering code data. Compared to human-centric approaches, the use of LLMs is particularly advantageous, as they effectively capture nuanced standards of code quality that are difficult to quantify explicitly, and simultaneously provide scalable, self-contained evaluations capable of processing billions of samples consistently. We applied these LLM filters during the preprocessing of our data -- including GitHub code, GitHub commits, and code-related web data -- and curated a code pretraining corpus comprising a total of 6 trillion tokens.

\noindent Building on this foundation, we introduce \ourmodel, a family of state-of-the-art open-source code LLMs at the 8B scale, including a base model, an instruct model, and a reasoning model. Starting from the base model trained with the proposed pretraining pipeline, the instruct model is further fine-tuned on large-scale synthetic data generated and filtered by LLMs, followed by direct preference optimization (DPO) to enhance the instruction-following capabilities. Meanwhile, the reasoning model improves multi-step reasoning in complex coding tasks by applying Long-Chain-of-Thought (LongCoT) reinforcement learning. Extensive evaluations on well-established benchmarks on a diverse set of coding tasks were conducted to showcase the advanced capabilities of the three variants of \ourmodel. We release this lightweight yet powerful model family with the intention of providing research insights while maintaining a manageable scale to facilitate further exploration within the open-source community.

\section{Pretraining}

\subsection{Data Pipeline}
\begin{figure}
    \centering
    \includegraphics[width=\linewidth]{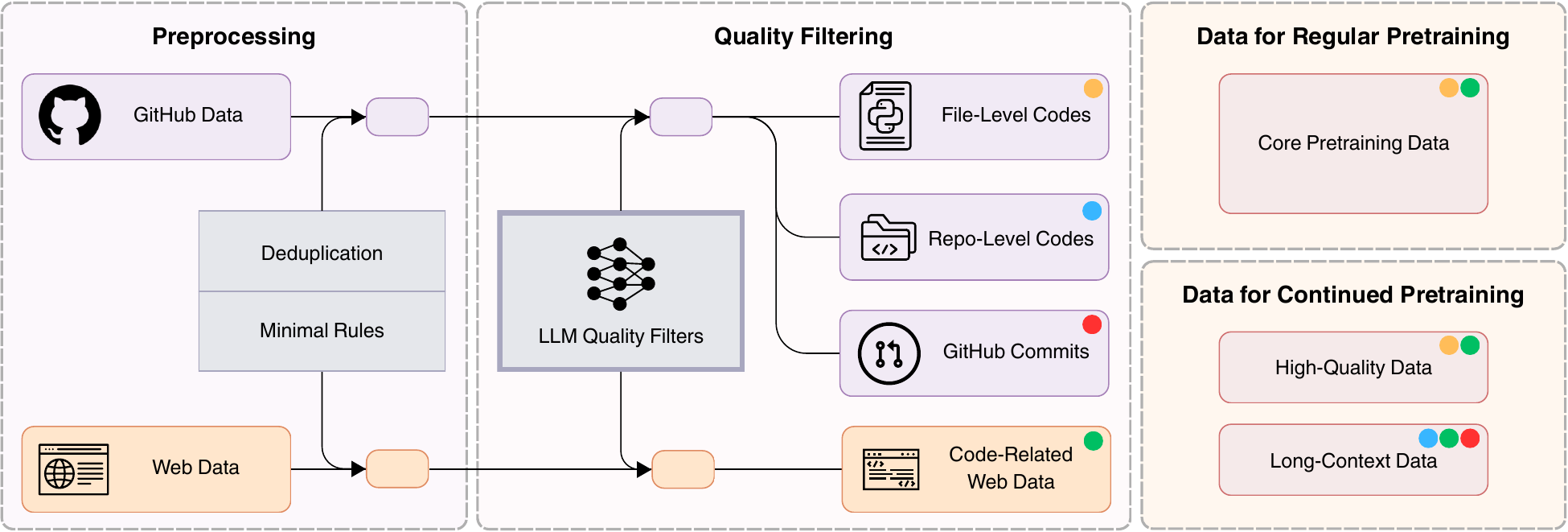}
    \captionsetup{justification=justified, singlelinecheck=false}
    \caption{Processing pipeline for pretraining data. We collected data from GitHub and web archives. The raw data were processed into four categories: file-level codes (yellow), repository-level codes (blue), GitHub commits (red) and code-related web data (green). For each phase in pretraining, we combined and reorganized the processed data from the four categories, indicated by colors on the top-right of the blocks.}
    \label{fig:main_pipeline}
\end{figure}

\noindent Producing data for LLM pretraining typically requires complex data pipelines and intricate data lineage. To ensure efficiency and reduce computation and storage costs, we propose a parallel design, which decouples the sequential dependency of various filters so as to streamline our data pipeline. As illustrated in Figure~\ref{fig:main_pipeline}, our end-to-end data pipeline takes as input the raw code data crawled from GitHub and web archives, and outputs the final pretraining data after a few steps of data processing. All the preprocessing and filtering modules were disentangled so that each can run individually to avoid re-running the entire pipeline for better support of incremental data expansion and flexible pipeline manipulation.

\noindent Our data pipeline starts with the standard exact- and near-deduplication to get a holistic set of code-related files without duplication. Then we applied basic filters with minimal rules such as language inferring to get rid of irrelevant and non-code data. On top of the preprocessed data, we developed advanced quality filters powered by LLMs, which capture general standards and exhibit strong generalizability to address the wide variety across the large-scale data. Our filtered data falls into four categories: 
\begin{itemize} [leftmargin=*, topsep=0pt, itemsep=5pt]
    \item \textbf{File-level codes:} Individual code files from GitHub Data.
    \item \textbf{Repository-level codes:} Code files structured based on the repositories.
    \item \textbf{Commits data:} Snapshots of GitHub Commits, including commit messages, repository metadata, all relevant files, and code patches.
    \item \textbf{Code-related web data:} Documents from web archives that contains code blocks or highly code-related.
\end{itemize}

\noindent Based on the four categories above, we designed our pretraining recipes mainly comprising two phases: In the regular pretraining phase, we used file-level codes and code-related web data to build the fundamental capabilities of our model; in the continued pretraining phase, we expanded to all four categories to enhance performance and alignment, while also stimulating the model's ability to understand long-context data.

\noindent In the next section, we detail the pipeline over individual data ingredients. Section~\ref{sec:pretraining_quality_filter} introduces the pipeline for GitHub data, with an emphasis on the general design and implementation of our LLM quality filters. Sections~\ref{sec:commits_data} and~\ref{sec:web_data} present the processing of GitHub commits and web archives, respectively. Sections~\ref{sec:hq_data} and~\ref{sec:lctx_data} describe the construction of the continued pretraining data. Finally, we conclude the pretraining section with discussions on the Fill-in-the-Middle (FIM) format applied to both phases.

\subsection{Data Ingredients}

\subsubsection{GitHub Data}
\label{sec:pretraining_quality_filter}
GitHub is widely regarded as the most abundant and valuable data source for training LLM coders. In our work, it serves as the primary component of our pretraining corpus. The raw data was collected in the form of repositories, and we adopted the two-stage processing pipeline illustrated in Figure~\ref{fig:main_pipeline} to extract a high-quality subset.

\noindent\textbf{Preprocessing.} We commenced by implementing deduplication at both the repository and file levels. For each level, we performed exact-deduplication using SHA256 hashes of contents and near-deduplication via the MinHash algorithm~\citep{mapreduce_minhash,lee-etal-2022-deduplicating}. This two-tier strategy yielded two variants of the code corpus: the file-level variant offered flexibility for training with short context windows, while the repository-level variant preserved the project structure, enabling more coherent long-context learning. Following the deduplication, we checked the remaining files via syntax parsers such as Tree-sitter~\citep{tree-sitter} and discarded those with syntax errors. Overall, the preprocessing stage reduced the raw data volume by approximately 98\%, resulting in a manageable dataset for downstream quality filtering.

\noindent\textbf{Quality Filtering.} Rule-based filters are widely used to enhance the quality of pretraining datasets. The rules defined by coding experts provide strong controllability and interpretability in high-quality document selection. However, these filters are inherently limited to rules that can be precisely defined, making it difficult to incorporate empirical heuristics. They also lack flexibility in adjusting filtering strength. More fundamentally, rule creation relies on expert consensus, a stringent requirement that may not always be met. A notable example in code datasets is class templates with extensive comments. Some may argue that detailed comments help the model understand the semantic meanings and usage of class functions and attributes. Others, however, may see them as diluting code-related content, potentially hindering coding capabilities by overemphasizing plain text rather than actual code lines. Similarly, a short document consisting only of API calls may be dismissed for its brevity and lack of logical statements, yet it can also be valued as a fundamental building block in software development.

\noindent Beyond the challenges of constructing rule-based filters, heuristics derived from human expertise can also be unreliable, as fully assessing a code document requires a comprehensive grasp of its entire context, which is an exhaustive and inherently unscalable task. Figure~\ref{fig:minor_logicall_error} illustrates this issue with a function that appears heuristically sound at first glance, featuring well-commented usage and a structured pipeline. However, it harbors subtle logical errors that are easily overlooked. Figure~\ref{fig:ascii_art} presents an even more extreme case, where human experts struggle to decipher the script's meaning unless viewed from a significant distance.

\begin{figure}
    \begin{center}
        \begin{adjustbox}{center}
\begin{lstlisting}[language=Python,style=mystyle2]
def main(temp):
    """
    Display the message according to the following conditions
    Temp < 0: "Freezing"
    Temp 1-10: "Very Cold"
    Temp 11-20: "Cold"
    Temp 21-30: "Normal"
    Temp 31-40: "Hot"
    Temp > 40: "Very Hot"
    Args:
        temp: integer
    Returns:
        string: the message to print
    """
    if temp < 0 and temp > 1:
        print("Freezing")
    if temp > 0 and temp < 11:
        print("Very Cold")
    if temp > 10 and temp < 21:
        print("Cold")
    if temp > 20 and temp < 31:
        print("Normal")
    if temp > 30 and temp < 41:
        print("Hot")
    if temp > 40:
        print("Very Hot")
    return
\end{lstlisting}
        \end{adjustbox}
    \end{center}
    \caption{Sample Python script with decent structure but logical errors.}
\label{fig:minor_logicall_error}
\end{figure}

\begin{figure}
    \centering
    \begin{subfigure}[b]{0.45\textwidth}
        \centering
\begin{lstlisting}[language=Python, style=mystyle]
t3 = [
    [0, 0, 7, 0, 0, 0, 0, 0, 0, 0, 0, 0, 0, 7, 0, 0],
    [0, 7, 7, 7, 0, 0, 0, 0, 0, 0, 0, 0, 7, 7, 7, 0],
    [0, 7, 7, 7, 0, 0, 0, 0, 0, 0, 0, 0, 7, 7, 7, 0],
    [0, 0, 7, 7, 7, 0, 0, 0, 0, 0, 0, 7, 7, 7, 0, 0],
    [0, 0, 0, 7, 7, 0, 0, 0, 0, 0, 0, 7, 7, 0, 0, 0],
    [0, 0, 0, 7, 7, 7, 0, 0, 0, 0, 7, 7, 0, 0, 0, 0],
    [0, 0, 0, 7, 7, 7, 7, 7, 7, 7, 7, 7, 7, 0, 0, 0],
    [0, 0, 7, 7, 7, 7, 7, 7, 7, 7, 7, 7, 7, 7, 0, 0],
    [0, 7, 7, 7, 7, 7, 7, 7, 7, 7, 7, 7, 7, 7, 7, 0],
    [0, 7, 7, 7, 7, 7, 7, 7, 7, 7, 7, 7, 7, 7, 7, 0],
    [0, 7, 7, 7, 7, 7, 7, 7, 7, 7, 7, 7, 7, 7, 7, 0],
    [0, 7, 7, 7, 0, 7, 7, 7, 7, 7, 7, 0, 7, 7, 7, 0],
    [0, 7, 7, 7, 7, 7, 7, 7, 7, 7, 7, 7, 7, 7, 7, 0],
    [0, 7, 7, 7, 7, 7, 7, 7, 7, 7, 7, 7, 7, 7, 7, 0],
    [0, 7, 7, 7, 7, 7, 7, 0, 0, 7, 7, 7, 7, 7, 7, 0],
    [0, 0, 7, 7, 7, 7, 7, 7, 7, 7, 7, 7, 7, 7, 0, 0],
    [0, 0, 0, 7, 7, 7, 7, 7, 7, 7, 7, 7, 7, 0, 0, 0],
    [0, 0, 0, 7, 7, 7, 7, 7, 7, 7, 7, 7, 7, 0, 0, 0],
    [0, 0, 0, 7, 7, 7, 7, 7, 7, 7, 7, 7, 7, 0, 0, 0],
    [0, 0, 0, 7, 7, 7, 7, 7, 7, 7, 7, 7, 7, 0, 0, 0],
    [0, 0, 0, 7, 7, 7, 7, 7, 7, 7, 7, 7, 7, 0, 0, 0],
    [0, 0, 0, 7, 7, 7, 7, 7, 7, 7, 7, 7, 7, 0, 0, 0],
    [0, 0, 0, 7, 7, 7, 7, 7, 7, 7, 7, 7, 7, 0, 0, 0],
    [0, 0, 0, 7, 7, 7, 7, 7, 7, 7, 7, 7, 7, 0, 0, 0],
    [0, 0, 0, 7, 7, 7, 7, 7, 7, 7, 7, 7, 7, 0, 0, 0],
    [0, 0, 0, 7, 7, 7, 7, 7, 7, 7, 7, 7, 7, 0, 0, 0],
    [0, 0, 0, 7, 7, 7, 7, 7, 7, 7, 7, 7, 7, 0, 0, 0],
    [0, 0, 0, 0, 7, 7, 7, 7, 7, 7, 7, 7, 0, 0, 0, 0],
    [0, 0, 0, 0, 0, 7, 7, 0, 0, 7, 7, 0, 0, 0, 0, 0],
    [0, 0, 0, 0, 0, 7, 7, 0, 0, 7, 7, 0, 0, 0, 0, 0],
    [0, 0, 0, 0, 7, 7, 7, 0, 0, 7, 7, 7, 0, 0, 0, 0],
    [0, 0, 0, 0, 0, 0, 0, 0, 0, 0, 0, 0, 0, 0, 0, 0]
]
\end{lstlisting}
    \end{subfigure}
    \hfill
    \begin{subfigure}[b]{0.45\textwidth}
        \centering
\begin{lstlisting}[language=Python, style=mystyle]
t3 = [
    [      7                                7      ] 
    [   7  7  7                          7  7  7   ] 
    [   7  7  7                          7  7  7   ] 
    [      7  7  7                    7  7  7      ] 
    [         7  7                    7  7         ] 
    [         7  7  7              7  7            ] 
    [         7  7  7  7  7  7  7  7  7  7         ] 
    [      7  7  7  7  7  7  7  7  7  7  7  7      ] 
    [   7  7  7  7  7  7  7  7  7  7  7  7  7  7   ] 
    [   7  7  7  7  7  7  7  7  7  7  7  7  7  7   ] 
    [   7  7  7  7  7  7  7  7  7  7  7  7  7  7   ] 
    [   7  7  7     7  7  7  7  7  7     7  7  7   ] 
    [   7  7  7  7  7  7  7  7  7  7  7  7  7  7   ] 
    [   7  7  7  7  7  7  7  7  7  7  7  7  7  7   ] 
    [   7  7  7  7  7  7        7  7  7  7  7  7   ] 
    [      7  7  7  7  7  7  7  7  7  7  7  7      ] 
    [         7  7  7  7  7  7  7  7  7  7         ] 
    [         7  7  7  7  7  7  7  7  7  7         ] 
    [         7  7  7  7  7  7  7  7  7  7         ] 
    [         7  7  7  7  7  7  7  7  7  7         ] 
    [         7  7  7  7  7  7  7  7  7  7         ] 
    [         7  7  7  7  7  7  7  7  7  7         ] 
    [         7  7  7  7  7  7  7  7  7  7         ] 
    [         7  7  7  7  7  7  7  7  7  7         ] 
    [         7  7  7  7  7  7  7  7  7  7         ] 
    [         7  7  7  7  7  7  7  7  7  7         ] 
    [         7  7  7  7  7  7  7  7  7  7         ] 
    [            7  7  7  7  7  7  7  7            ] 
    [               7  7        7  7               ] 
    [               7  7        7  7               ] 
    [            7  7  7        7  7  7            ] 
    [                                              ]
]
\end{lstlisting}
    \end{subfigure}
    \captionsetup{justification=justified, singlelinecheck=false}
    \caption{Sample code snippet from a Python script for LED display controlling. The original code (left) and a visually intuitive format by replacing zeros and commas with space (right) are presented.}
\label{fig:ascii_art}
\end{figure}

\begin{figure}[]
    \centering
    \includegraphics[width=\textwidth]{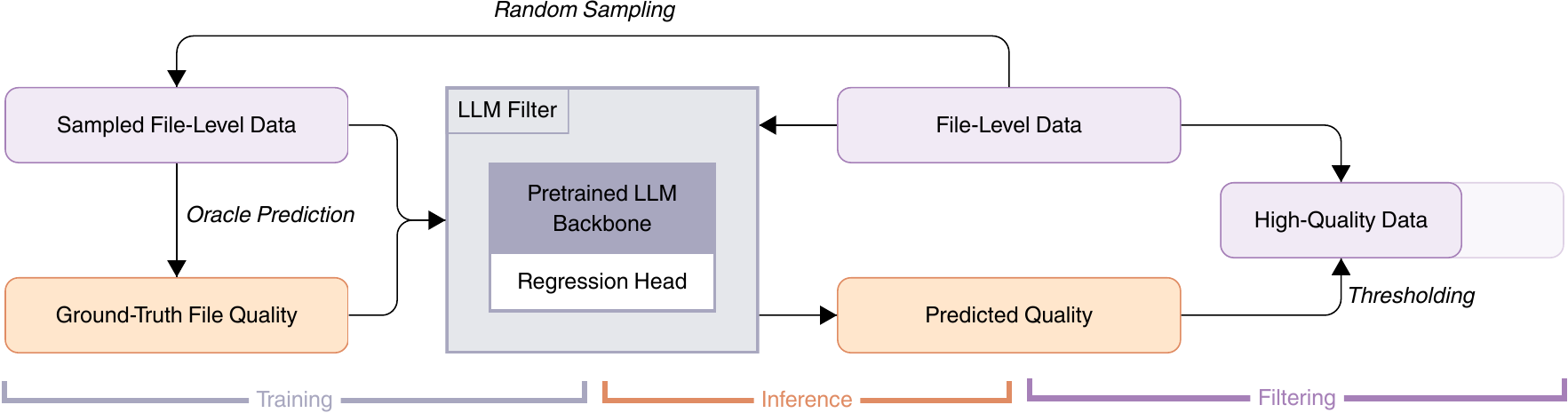}
    \caption{Pipeline of our data quality scorer.}
    \label{fig:scorer_pipeline}
\end{figure}

\noindent To overcome the above shortness of human-oriented filter designs, we propose a file-level scoring model to filter out low-quality code files from GitHub data in one shot, freeing us from fabricating the complicated standards one by one. The pipeline for constructing and using the scorer as a quality filter is shown in Figure~\ref{fig:scorer_pipeline}. To construct the specific training set for this filter, we randomly sampled $222,066$ code files from the most commonly used programming languages and queried DeepSeek-V2-Chat~\citep{deepseekai2024deepseekv2strongeconomicalefficient} (conceptually referred to as ``the oracle'') to assess four key aspects: readability, modularity, clarity, and reusability.
\begin{itemize} [leftmargin=*, topsep=0pt, itemsep=5pt]
\item \textbf{Readability:} High-quality code should include a reasonable amount of comments, follow consistent naming conventions, and adhere to common formatting and structural practices.
\item \textbf{Modularity:} High-quality code should be well-structured, avoiding overly complex or lengthy functions by leveraging modularization. Each module or component should serve a distinct, coherent purpose with a clear separation of logic and functionality.
\item \textbf{Clarity:} High-quality code should minimize redundancy, such as excessive function calls and large blocks of commented-out code or debugging print statements. It should clearly convey the intent behind each code block.
\item \textbf{Reusability:} High-quality code should be free of syntax and logical errors, avoid excessive hard-coded data, and be designed for easy integration into other projects with complete and meaningful functionality.
\end{itemize}

\noindent These standards primarily apply to substantial code files. However, randomly sampled code files may also contain incidental content such as configurations, data, or documents. To mitigate this, we applied high-level constraints to exclude such cases. Additionally, we filtered out auto-generated code, which often contains repetitive blocks and may introduce rigidity during pretraining. The final prompt for evaluating the quality of GitHub code files can be found in Appendix~\ref{appd:quality_prompt}.

\noindent The oracle was required to give an overall score ranging from 0 to 10 that evaluates the quality of code files (higher indicates better quality), with detailed explanations to support the score. Subsequently, only the score itself was extracted from the oracle's response to serve as the ground-truth label for each file. To scale the scorer to the full GitHub data, we opted for a regression model rather than a binary classifier. This approach allows for finer-grained quality evaluation, avoiding the rigidity of rule-based pass-or-fail filtering. Prioritizing efficiency for large-scale inference, we rescaled the ground-truth scores to the range [0,1] and fine-tuned a pretrained 1.3B model of Llama 2~\citep{touvron2023llama2} structure with a regression head for one epoch as the quality scorer. To balance the quality and diversity of code pretraining data, we filtered out from the entire GitHub dataset the bottom \textasciitilde10\% files, aggregating to \textasciitilde1T unique tokens based on this quality scoring method. This corpus supports 89 programming languages, forged into both the repository-level and file-level code data shown in Figure \ref{fig:main_pipeline}. The list of supported languages can be found in Appendix~\ref{appd:supported_langs}. Figure~\ref{fig:llm-filter-ablation} offers a glimpse into the effectiveness of LLM-based filters by presenting a snapshot of benchmark performance during pretraining. Further details on oracle selection, ground-truth generation, model architecture, and ablation studies can be found in Appendix~\ref{appd:pretraining_quality_filter}.

\begin{figure} %
    \centering %
    \begin{subfigure}[b]{0.34\textwidth}
        \centering
        \includegraphics[width=\textwidth]{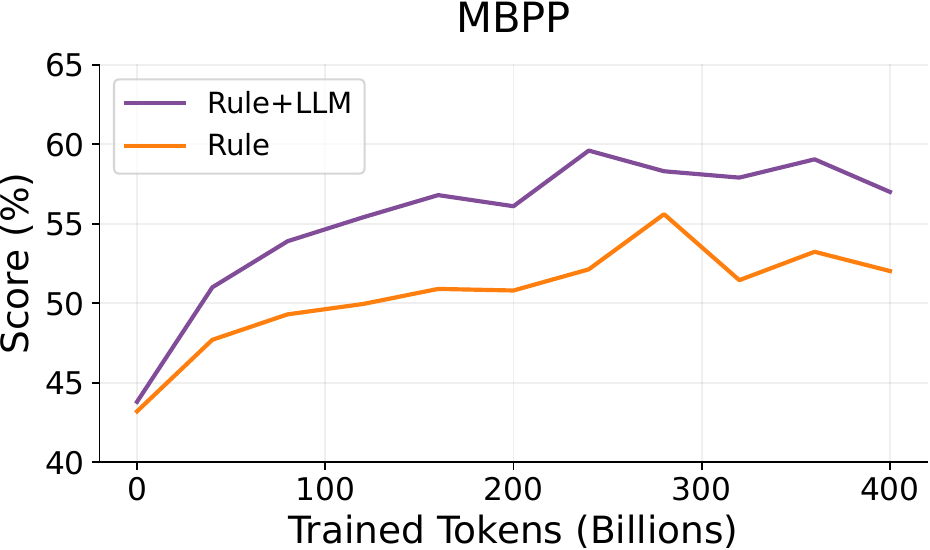}
        \label{fig:mbpp}
    \end{subfigure}
    \hfill %
    \begin{subfigure}[b]{0.31\textwidth}
        \centering
        \includegraphics[width=\textwidth]{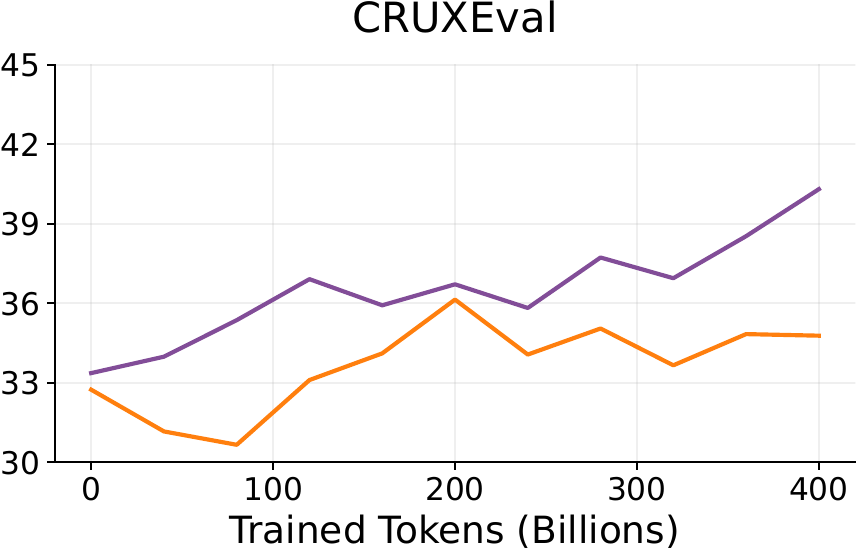}
        \label{fig:cruxeval}
    \end{subfigure}
    \hfill %
    \begin{subfigure}[b]{0.31\textwidth}
        \centering
        \includegraphics[width=\textwidth]{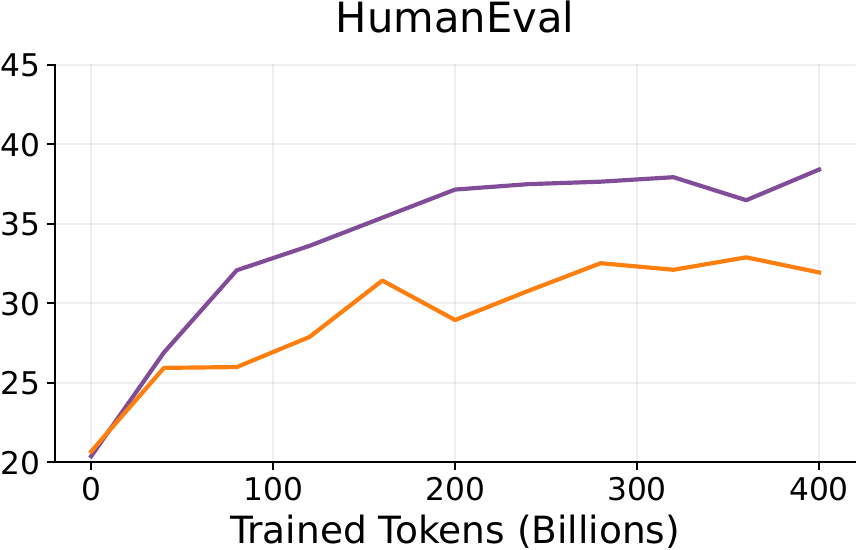}
        \label{fig:humaneval}
    \end{subfigure}
    \captionsetup{justification=justified, singlelinecheck=false}
    \vspace{-12pt}
    \caption{On-the-fly performance over benchmarks during pretraining, from data with minimal rules only (orange) and LLM-based filters involved (purple).}
    \label{fig:llm-filter-ablation}
\end{figure}

\subsubsection{Commits Data}
\label{sec:commits_data}

Besides GitHub code files, GitHub commits data encapsulates the collective wisdom of developers, capturing how code evolves through bug fixes, feature updates, and iterative refinement. To better align \ourmodel with practical development workflows, we incorporated large-scale GitHub commits data into our pretraining.
Specifically, we collected $74$ million commits from $140$K high-quality repositories, selected based on the following criteria: at least $100$ stars, $10$ forks, $100$ commits, and $100$ days of maintenance activity.
Each commit includes metadata such as the commit message, patch, merge status, and pre-commit code snapshot.

\noindent To utilize GitHub commits data for pretraining, we format each sample as a code change prediction task:
given a commit message and its associated context, the model predicts the modified file paths and the corresponding code changes.
The context includes the pre-commit code snapshot's README, directory structure, and top 5 relevant files retrieved via BM25~\citep{bm25}.
After deduplication and preprocessing, we obtained a curated corpus of approximately $100$ billion tokens of commits for pretraining, providing dense supervision for learning real-world code change patterns.

\subsubsection{Code-Related Web Data}
\label{sec:web_data}

\begin{figure}
    \centering
    \includegraphics[width=\textwidth]{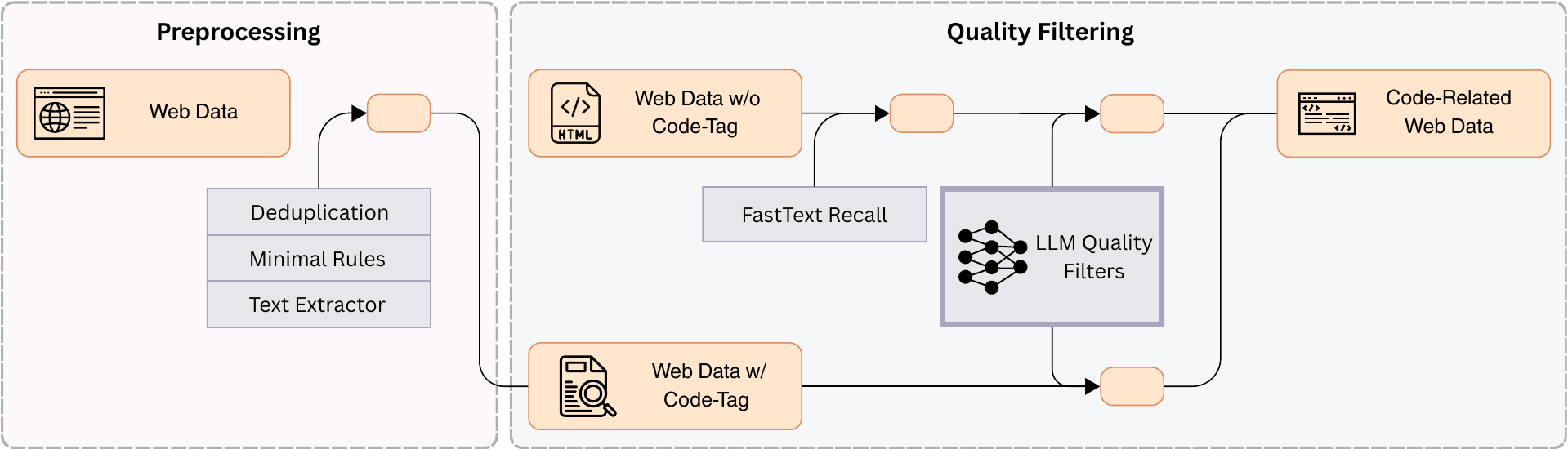}
    \caption{Pipeline of extracting high-quality code-related data from web archives.}
    \label{fig:web_data_pipeline}
\end{figure}

In this section, we propose an optimized framework for extracting code-related data from extensive web archives like Common Crawl\footnote{\url{https://commoncrawl.org/}}. As shown in Figure~\ref{fig:web_data_pipeline}, our extraction framework incorporates preprocessing and quality filtering mechanisms, thereby ensuring the harvested code-related data exhibits both high quality and diversity.

\noindent \textbf{Preprocessing.}
The framework commences with efficient preprocessing of large-scale web archives. We implemented text extraction procedures on Common Crawl and identified two distinct categories of raw data:  1) web pages with explicit code tags (such as $\text{<code>\ldots</code>}$) in HTML that are readily extractable using standard rules, and 2) non-explicit code-tag data potentially containing code or related knowledge, which  presents extraction challenges due to its volume and complexity. Paralleling our GitHub data processing approach, we implemented exact- and near-deduplication techniques, and developed heuristics to eliminate low-quality documents (e.g., less than 10 words) during the preprocessing stage. 

\noindent \textbf{Quality Filtering.} The framework implements dual complementary strategies to ensure data quality: first, identifying content with code relevance, and second, assessing the intrinsic quality of the identified content.

\begin{itemize} [leftmargin=*, topsep=0pt, itemsep=5pt]
\item \textbf{FastText Recall:} To handle the complexity and scale of non-explicit code-tag data, we extracted and scored 10 million candidate web pages from Common Crawl data to establish gold-standard datasets for evaluation. From the annotated dataset, we reserved 70\% data as the initial seed corpus while preserving the remainder for validation. Using the seed corpus, we trained a fastText model~\citep{joulin2016fasttext, joulin-etal-2017-bag} designed to identify and retrieve additional code-related content. In validation dataset, our model demonstrated a recall rate of 99\% alongside a precision rate of 45\%. Through this methodology, we successfully identified approximately 3\% of Common Crawl data as code-related candidates.
\item \textbf{LLM Quality Filters:} We employed a scoring system ranging from 0 to 10 to assess the quality of code-related files using an LLM-based evaluator. During our analysis of the collected code-related web content, we observed significant variations in quality scores across different website categories. This variation revealed biases in our initial scoring methodology that required careful mitigation to ensure a balanced and representative dataset. Notably, e-commerce platforms, color-related content, and documentation sites consistently received higher scores due to their standardized formatting and structured presentation, while forums and community discussion platforms typically scored lower despite containing valuable content, primarily attributable to their heterogeneous and less structured formats. To rectify this imbalance and enhance dataset representativeness, we implemented category-specific filtering protocols that adjusted quality thresholds and sampling rates for each category, thereby preventing over representation of individual content types.
\end{itemize}
The pipeline generated a robust corpus of approximately 1.2 trillion tokens, ensuring balanced representation across diverse web content categories while maintaining both quality standards and categorical diversity.

\subsubsection{High-Quality Data for Continued Pretraining}
\label{sec:hq_data}
To further improve model performance and better align the distribution between the pretraining model and post-training data, we constructed high-quality datasets by combining the quality score with iteratively trained fastText~\citep{joulin2016fasttext, joulin-etal-2017-bag} models.

\noindent These high-quality datasets were obtained from multiple sources, including major programming languages, algorithms, application development, Jupyter notebooks, and general code data. From each source, we initially curated a small yet diverse high-quality seed dataset (approximately 100K samples), selected based on specific data characteristics such as quality scores, programming language, comment ratios, imported packages, etc. This seed dataset served as positive samples for training a fastText model. Negative samples consisted of two parts: randomly selected samples and carefully constructed hard negative samples. Random negative samples were obtained by random selection from the original data after removing the high-quality seeds. Hard negative samples were deliberately designed to resemble high-quality seeds closely, such as code samples with high filter scores but lacking comments or docstrings, or data recalled by the first-round fastText model but scoring poorly according to the quality filter. Hard negative samples were crucial for effectively training the fastText model, as they prevented the model from overfitting positive examples, which otherwise might lead to uniformly high scores and reduced discriminative capability. 

\noindent Finally, after training 2--3 rounds, each round expanded the seed dataset by incorporating newly identified positive samples, we obtained an effective high-quality data recall model and constructed approximately 130 billion tokens of high-quality data for continued pretraining.

\subsubsection{Long-Context Data for Continued Pretraining}
\label{sec:lctx_data}
We introduce long-context training by supporting sequences of up to 32K tokens during the continued pretraining phase. We conducted two stages, progressively expanding the context window from the original 8K to 32K. This stage utilized approximately 1 trillion training tokens. Training data were organized into two complementary granularities:

\begin{itemize} [leftmargin=*, topsep=0pt, itemsep=5pt]
\item \textbf{File-level:} Long context data filtered from GitHub repositories and code-related web data using LLM filtering techniques.
\item \textbf{Repository-level:} We selected high-quality repositories based on average file quality scores. For mainstream programming languages (e.g., Python, Java, and C), we implemented topological concatenation based on file dependencies. For HTML, SQL, and Shell, we used random concatenation. Each repository was mapped to a single string sequence, with exceptionally large repositories (e.g., PyTorch) being decomposed into multiple independent subgraphs to avoid oversized sequences while preserving logical coherence.
\end{itemize}

\subsubsection{Fill-in-the-Middle (FIM)}
Beyond the objective of next-token prediction, \citet{fim} introduces the Fill-in-the-Middle (FIM) training to enhance context-aware completion. Due to the flexibility of prompt reformatting in real-world cases, the Suffix-Prefix-Middle (SPM) and Prefix-Suffix-Middle (PSM) modes are considered contextually equivalent. Consequently, most recent works~\citep{qwen25coder, zhu2024deepseek} favor a single-mode FIM approach in their training strategies. However, in contrast to the more commonly adopted PSM mode, we found that SPM performs slightly better during the training process, which is consistent with~\citep{bavarian2022efficient}. One possible explanation for this outperformance is the positional bias in attention mechanisms~\citep{yu2024unveiling, hsieh2024found}: in SPM mode, the beginning of the suffix and the end of the prefix are positioned at opposite ends of the input token sequence, both crucial for middle-content prediction.

\noindent During pretraining, we employed character-level random splitting to support intra-word completion. Each FIM sample was converted from the original single-file document into the following format:
\begin{equation*}
\texttt{<[fim-suffix]>}SUFFIX\texttt{<[fim-prefix]>}PREFIX\texttt{<[fim-middle]>}MIDDLE
\end{equation*}
where \texttt{<[fim-suffix]>}, \texttt{<[fim-prefix]>} and \texttt{<[fim-middle]>} were added as special tokens, and $SUFFIX$, $PREFIX$, $MIDDLE$ were the corresponding parts split from the document.
For the regular pretraining phase, the FIM ratio was set to $0.5$. For the continued pretraining phase, the FIM ratio was set to $0.1$.

\subsection{Pretraining Policy}
Our pretraining model architecture follows the widely adopted Llama 3~\citep{meta2024llama3} structure and is configured as follows: the model has 8.2 billion parameters, 36 layers, with a hidden size of $4,096$ and an intermediate size of $14,336$, and employs Grouped Query Attention (GQA) with 32 heads for queries and 8 heads for keys and values. We do not apply tied embeddings. The context length during regular pretraining was set to 8K tokens, which was later expanded to 32K tokens during the continued pretraining and post-training phases.

\noindent Our pretraining consumed 6 trillion tokens in total. Initially, we pretrained the model with a learning rate of $3\mathrm{e}{-4}$ on a mixture of code-related web data and math-related web data for the first 1 trillion tokens. This was followed by an additional 4 trillion tokens of training on curated code data. During the continued pretraining phase, we switched to high-quality and long-context datasets: the learning rate was first reduced by a factor of $\sqrt{10}$, and the model was trained for 400 billion tokens. Subsequently, the learning rate was further decreased to $3\mathrm{e}{-5}$, and continued training for an additional 600 billion tokens.

\section{Post-training}
\subsection{Instruct Model}
\begin{figure}
    \centering
    \includegraphics[width=\textwidth]{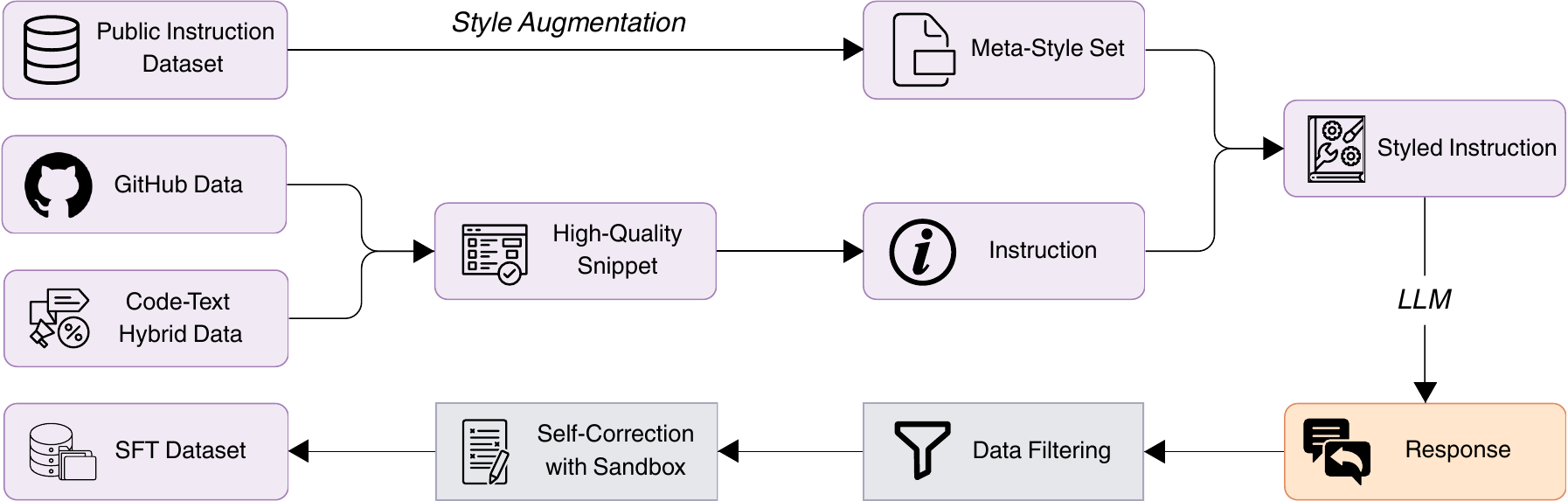}
    \caption{Pipeline of synthetic data curation for our instruct model training.}
    \label{fig:sft_pipeline}
\end{figure}

To build \ourmodel-8B-Instruct, we first applied supervised fine-tuning (SFT) to adapt our base models to a wide range of real-world tasks, followed by direct preference optimization (DPO; \citet{rafailov2023direct}) to further enhance specific capabilities such as code generation and reasoning. For SFT, we constructed an instruction tuning dataset containing millions of samples. In curating this dataset, we focused on three primary aspects: diversity, quality, and difficulty, which are critical for improving the model's robustness and generalization. The pipeline for SFT dataset curation is illustrated in Figure~\ref{fig:sft_pipeline}. Specifically, we first collected high-quality code snippets from curated GitHub repositories and code-text hybrid datasets, and built a meta-style set through style augmentation over public instruction data. These sources were then used to prompt LLMs to synthesize diverse, styled instructions \citep{wei2024magicoderempoweringcodegeneration} and corresponding responses. The instruction-response pairs were filtered through a combination of rule-based and model-based techniques to ensure high quality and appropriate difficulty levels. Finally, we applied a sandbox-based self-correction mechanism to iteratively refine the outputs. The finalized high-quality examples were compiled into the SFT dataset for training. We detail this process in the following subsections.

\subsubsection{Data Construction: Diversity}
We primarily employed synthetic data generation to construct an instruction dataset emphasizing diversity, difficulty, scalability, and quality. During prompt synthesis, we prioritized prompt diversity to enhance the LLM's robustness across diverse scenarios and edge cases. We leveraged high-quality seed snippets to promote diversity in two dimensions: seed snippet diversity and style diversity.

\noindent \textbf{Seed Snippet Diversity.}  
To ensure broad coverage, we collected code snippets from multiple high-quality sources. Specifically, we extracted curated GitHub code snippets, applying rigorous filtering based on file quality scores. We then utilized OSS-Instruct \citep{wei2024magicoderempoweringcodegeneration} to efficiently generate additional code snippets. While isolated code snippets provide valuable supervision signals, they may not sufficiently expose the model to complex real-world scenarios and user interaction patterns. To address this, we expanded our data sources to include code-text hybrid datasets such as Markdown files, Jupyter notebooks, and StackExchange discussions. These sources offer not only executable codes but also accompanying explanations, usage examples, and problem-solving discussions; thus, they better approximate real-world user interactions. By incorporating both codes and surrounding contexts, we enable the model to generate instructions that closely align with how developers write, explain, and seek assistance.

\noindent \textbf{Style Diversity.}  
Although synthetic data offers an effective and cost-efficient alternative to real-world data \citep{liu2024bestpracticeslessonslearned}, it often lacks the stylistic diversity observed in actual user prompts. For instance, model-generated instructions are typically well-structured and complete, whereas real-world prompts tend to be more casual and varied. To bridge this gap, we built a \textit{meta-style set} by collecting diverse instructions from public datasets. To further enrich this set, we applied \textit{style augmentation}, wherein two styles were randomly selected and blended to synthesize new styles. During instruction generation, prompts were reformatted to match randomly selected styles from the meta-style set, thereby enhancing the model's robustness against different prompting behaviors.

\noindent To further improve the model's ability to respond to real-world queries, we additionally incorporated code-related data sampled from WildChat~\citep{zhao2024wildchat1mchatgptinteraction} into our SFT dataset.

\subsubsection{Data Filtering: Quality and Difficulty}
We employed a combination of rule-based and model-based techniques to filter out low-quality SFT pairs from the dataset.

\noindent \textbf{Quality Filtering.}  
To ensure data quality, we applied both rule-based and model-based filters. For rule-based filtering, we used Tree-sitter~\citep{tree-sitter} to eliminate responses containing syntax errors. For model-based filtering, we prompted an evaluation model to score responses based solely on correctness, discarding those with low scores.

\noindent \textbf{Difficulty Filtering.} 
We first classified the topic of each instance using tagging techniques \citep{lu2023instaginstructiontagginganalyzing}. Subsequently, we prompted a model to assess the difficulty level of each instance within its domain. Instances receiving a difficulty score lower than 3 out of 10 were discarded from the dataset.

\subsubsection{Self-Correction with Sandbox Verification}
In our experiments, we observed that examples with higher difficulty scores often exhibit higher error rates, resulting in many challenging instances being filtered out during quality filtering. To mitigate this and recall more high-difficulty examples in the SFT dataset, we implemented a sandbox verification and self-correction framework. Specifically, we prompted the model to generate solutions along with corresponding unit tests, evaluate the outputs within a sandbox environment, and iteratively refine any failed solutions until all tests passed or a maximum number of revision attempts was reached. This iterative refinement process allows us to retain more challenging examples and enrich the difficulty distribution of the training corpus.

\subsubsection{Direct Preference Optimization}
To further enhance the model's capabilities in code generation and reasoning, we constructed on-policy preference data for DPO. We first selected task-relevant prompts and sampled hundreds of candidate responses for each. These responses were then evaluated in a sandbox environment using generated code and corresponding unit tests. Based on the evaluation results, we formed preference pairs for DPO training to refine the model's behavior.

\subsubsection{Training Recipe for Instruct Model}
We fine-tuned \ourmodel-8B-Instruct in two stages: In the first stage, we performed SFT on the instruction dataset described above. The dataset contained approximately 3 million high-quality instruction-response pairs spanning diverse tasks and domains. To enhance the model's ability to handle challenging tasks, we employed difficulty-aware sampling to prioritize higher-difficulty examples during training. All prompts were reformatted using a system message and a chat-style template to better align the model's outputs with instruction-following conventions. The model was trained for 3 epochs using a learning rate of $2\mathrm{e}{-5}$ and enabled sequence packing to improve training efficiency. In the second stage, we applied DPO to further strengthen the model's capabilities in code generation and reasoning. We constructed approximately $20,000$ high-quality preference pairs by focusing on challenging examples from these domains.

\noindent This two-stage training strategy enables the model to learn both general instruction-following behaviors and fine-grained preferences in complex domains, ultimately leading to improved robustness, reasoning ability, and code generation performance.

\subsection{Reasoning Model}
Recently, models employing Long-Chain-of-Thought (LongCoT) reasoning have demonstrated significant performance gains on complex and verifiable tasks, such as mathematics and coding problems. To explore the upper limits of our 8B-sized base model, we conducted LongCoT reinforcement learning (RL) training and evaluated the model's performance on difficult problems. Starting from our base model, we first performed a LongCoT warmup phase, followed by GRPO-based RL training to further enhance its capabilities.

\subsubsection{Data}
We collected challenging real-world coding problems, including CodeContests~\citep{li2022competition} and ICPC problems\footnote{\url{https://icpc.global/worldfinals/past-problems}}, and gathered model-generated solutions from DeepSeek-R1~\citep{deepseekai2025deepseekr1} on these tasks. Additionally, we incorporated open-source chain-of-thought datasets such as open-r1/codeforces-cots~\citep{penedo2025codeforcescots} for a distilled model. Note that, for CodeContests and ICPC data, we applied sandbox-based rejection sampling during collection, retaining only correct model generations. This ensures that our warmup set provides stronger supervision compared to other distilled models. For the RL training, we utilized the aforementioned datasets combined with LiveCodeBench data collected prior to August 2024, excluding any unverifiable samples.

\subsubsection{Warmup Step}
Before proceeding to reinforcement learning (RL) training, we fine-tuned \ourmodel-8B-Base on thousands of collected warmup LongCoT samples using a learning rate of $2\mathrm{e}{-5}$. We initiated the warmup process from the base model rather than the instruct model, despite the latter achieving higher scores immediately following warmup completion. This decision stemmed from our observation that the instruct model frequently collapses into typical SFT training data patterns during RL, ultimately degrading its post-RL performance.


\noindent During the warmup phase, we observed that model performance improvements were positively correlated with the amount of distillation data used. However, as pointed out by \citet{yue2025doesreinforcementlearningreally}, a large volume of distillation data may exceed the inherent capabilities of the base model and thus fail to accurately reflect its true potential. To preserve the model's own exploration space, we chose not to further scale up the distillation data. Instead, we employed a relatively small amount of data to encourage the base model to learn diverse thinking patterns, relying on RL for subsequent self-improvement in order to better explore the model's upper bound.

\subsubsection{Training Recipe for Reasoning Model}

\begin{figure}[h]
    \centering
    \includegraphics[height=4cm]{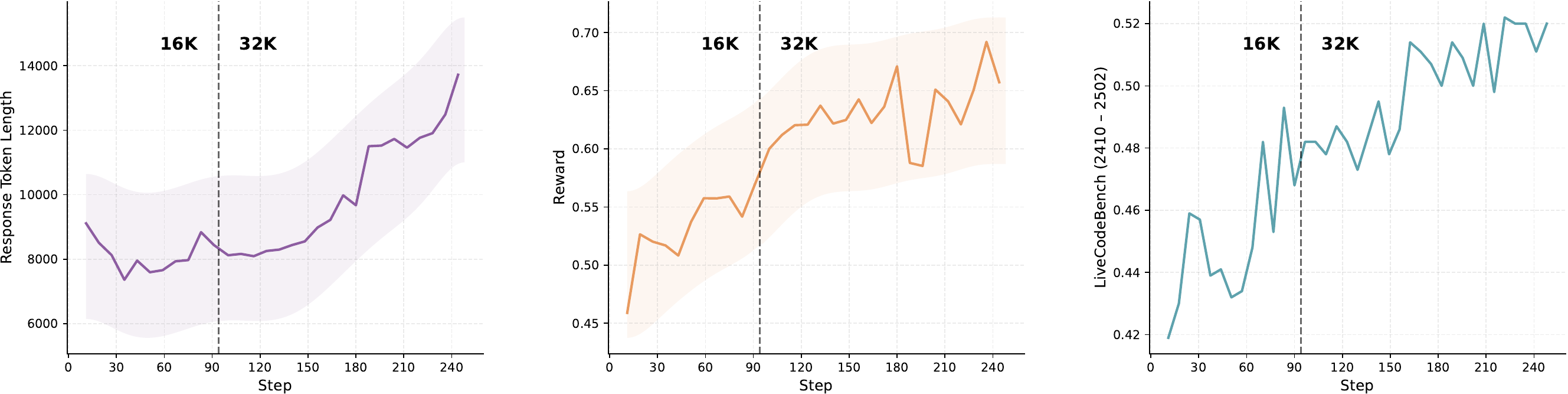}
        \captionsetup{justification=justified, singlelinecheck=false}
    \caption{RL training trajectories: smoothed average response token length (left), smoothed average rewards (middle), LiveCodeBench pass@1 performance (right). The area to the left of the dashed line corresponds to a 16K window, and the area to the right corresponds to a 32K window.}
    \label{fig:reasoning-rewards}
\end{figure}

During the RL training process, similar to DeepCoder~\citep{deepcoder2025}, we used the open-source verl framework~\citep{Sheng_2025} for GRPO training~\citep{deepseek-math}, and adopted optimization techniques similar to DAPO~\citep{yu2025dapoopensourcellmreinforcement,seed2025seed15thinkingadvancingsuperbreasoning}. The training configuration included a batch size of 128, an initial learning rate of $1\mathrm{e}{-6}$, a temperature of 0.6, and we removed the KL loss. We set the clip ratio high at 0.28, filtered overlong samples, and applied the token-wise loss during training. In Figure~\ref{fig:reasoning-rewards}, we present the training trajectories of response length, rewards, and benchmark performance of our reasoning model.

\noindent In addition, based on several empirical observations, we applied further modifications:

\noindent\textbf{Optimized Curriculum Learning.}
While GRPO inherently incorporates curriculum learning by filtering entirely correct or incorrect samples, we identified two areas for improvement: For simple problems, the model's thinking process contained significant redundancy, including repetitive examples and superficial logic. Since thinking patterns play a crucial role in scaling capabilities, we further filtered simple problems -- those with correctness rates above $87.5\%$ -- to encourage more efficient utilization of thinking tokens. Moreover, due to the addition of format rewards, situations arose with groups containing positive-only examples and format errors, which would not be automatically filtered out by GRPO. As these also qualify as simple problems, we filtered them out as well.

\noindent\textbf{Progressive Exploration Strategy.}
We adopted a gradually expanding strategy for sequence length and rollout number during training, further extending the progressive approach introduced in \citep{deepcoder2025}. Specifically, we first trained for 90 steps with a 16K sequence length and 16 samples per prompt, followed by 160 steps with a 32K sequence length and 32 samples per prompt, totaling 250 steps. Through empirical comparison, we found that expanding the sequence length and the rollout number yielded comparable final results, but the progressive schedule significantly improved training efficiency during the early stages. Preliminary observations also suggest that further increasing the rollout size to 64 or higher could potentially lead to even better performance. We encourage future work to explore more systematic strategies for allocating computational resources across different learning stages.

\section{Decontamination}

To ensure that our training data is not affected by potential test data leakage, we performed a decontamination process on the entire training data, including our pretraining and post-training data. Specifically, we employed the widely adopted 10-gram filtering method~\citep{deepseek_coder, qwen25coder}, which removed all data that had any 10-gram word overlap with key benchmark datasets, including HumanEval, MBPP, MHPP, BigCodeBench, LiveCodeBench, Aider, FullStack Bench, etc.

\section{Results}

We performed extensive series evaluations to investigate the performance of 1) \ourmodel-8B-Base: the pretrained base model, 2) \ourmodel-8B-Instruct: the instruction fine-tuned model, and 3) \ourmodel-8B-Reasoning: the reasoning model trained by reinforcement learning. We present the results of these evaluations in separate sections below. We compare with previous state-of-the-art open-source LLMs as follows.
\begin{itemize} [leftmargin=*, topsep=0pt, itemsep=5pt]
    \item \textbf{StarCoder2}~\citep{lozhkov2024starcoder2} is the successor of StarCoder~\citep{li2023starcoder}, a classic code LLM with 15.5B parameters trained on 1 trillion tokens sourced from the Stack~\citep{li2023starcoder}, which consists of 6.4TB of permissively licensed source code in 384 programming languages. The fully upgraded version StarCoder2 consists of 3B, 7B, and 15B parameters models trained on 3.3 to 4.3 trillion tokens of the Stack v2~\citep{lozhkov2024starcoder2}, with 67.5TB code data spanning 619 programming languages.
    \item \textbf{Codestral}~\citep{mistral2024codestral} is an open-weight 22B parameter model released by Mistral AI, which is explicitly designed for code generation tasks. It is trained on a dataset of 80+ programming languages.
    \item \textbf{CodeLlama}~\citep{codellama} is a family of code LLMs developed based on Llama 2~\citep{touvron2023llama2} models, by firstly training on 500B to 1T tokens from a code-heavy dataset and then fine-tuning with additional 5B tokens to better follow human instructions. CodeLlama is available in four sizes: 7B, 13B, 34B, and 70B parameters.
    \item \textbf{DeepSeek-Coder}~\citep{deepseek_coder} is a series of code LLMs with 1.3B, 6.7B, and 33B parameters, each trained from scratch on 2 trillion tokens, with a composition of 87\% code and 13\% natural language in both English and Chinese. These models are pretrained on a project-level code corpus and an additional fill-in-the-blank task, which enables project-level code completion and infilling.
    \item \textbf{CodeQwen1.5}~\citep{codeqwen1.5} is a specialized code LLM with 7B parameters built upon the Qwen1.5 language model~\citep{qwen1.5}, via training on around 3 trillion tokens of code-related data that covers 92 programming languages.
    \item \textbf{Yi-Coder}~\citep{yi-coder} is a series of code LLMs in two sizes (1.5B and 9B), trained on top of 2.4 trillion high-quality tokens sourced from a repository-level GitHub code corpus that covers 52 programming languages and code-related data filtered from CommonCrawl.
    \item \textbf{Llama 3.1}~\citep{meta2024llama31} is a suite of general LLMs with 8B, 70B, and 405B model parameters, which is an upgrade of the previously released Llama 3~\citep{meta2024llama3} series. Llama 3.1 is pretrained on a corpus of about 15 trillion multilingual tokens, compared to 1.8 trillion tokens for its predecessor Llama 2~\citep{touvron2023llama2}.
    \item \textbf{OpenCoder}~\citep{huang2024opencoder} is a series of code LLMs which includes 1.5B and 8B parameter models, trained from scratch on 2.5 trillion tokens composed of 90\% raw code and 10\% code-related web data.
    \item \textbf{DeepSeek-Coder-V2}~\citep{deepseek_coder} is a series of Mixture-of-Experts (MoE) code LLMs with 16B and 236B parameters, which has 2.4B and 21B activation parameters, respectively. These models are further pretrained from a checkpoint at 4.2 trillion tokens of DeepSeek-V2~\citep{zhu2024deepseek} with additional 6 trillion tokens comprising 60\% source code, 10\% math corpus, and 30\% natural language corpus.
    \item \textbf{Qwen2.5-Coder}~\citep{qwen25coder} is a series of code LLMs that has a rich variety of model sizes, including 0.5B, 1.5B, 3B, 7B, 14B, and 32B parameter models, which are further pretrained from Qwen2.5~\citep{qwen2.5} on over 5.5 trillion tokens of code-related data. As the flagship model of this series, Qwen2.5-Coder-32B-Instruct has achieved the best overall performance among open-source models on various coding benchmarks, and even has competitive performance with powerful proprietary LLMs such as GPT-4o.
    \item \textbf{Qwen3}~\citep{qwen3} is the most recent addition to the Qwen family of open-source LLMs. The Qwen3 series includes two MoE models -- Qwen3-235B-A22B (235B total parameters, with 22B activated parameters) and Qwen3-30B-A3B (30B total parameters, with 3B activated parameters) -- as well as six dense models: Qwen3-32B, Qwen3-14B, Qwen3-8B, Qwen3-4B, Qwen3-1.7B, and Qwen3-0.6B. The Qwen3 models introduce a hybrid approach consisting of a ``non-thinking mode'' and a ``thinking mode'': In non-thinking mode, the model provides quick responses, whereas in thinking mode, the model reasons in a step-by-step manner before generating a final response. To ensure fair evaluation that accounts for both model size and modeling type, we compare the performance of Qwen3-8B in non-thinking mode against standard instruct models, and its performance in thinking mode against dedicated reasoning models.
\end{itemize}

\subsection{Evaluation of Base Models}

In this section, we report the evaluation results of our pretrained base model \ourmodel-8B-Base, compared with various previous state-of-the-art models to assess their capabilities in code generation, code completion, and code reasoning. To ensure the reproducibility of our results, we use the open-source evaluation suites released by Qwen2.5-Coder~\citep{qwen25coder}, adopting their reported performance metrics when available and conducting our own evaluations for the remaining ones.

\begin{table}[t]
\vspace{-10pt}
\centering
\resizebox{0.75\textwidth}{!}{
\begin{tabular}{lrcccc}
\toprule
\multirow{2}{*}{\textbf{Model}} & \multicolumn{1}{r|}{\multirow{2}{*}{\textbf{Size}}} & \multicolumn{2}{c|}{\textbf{HumanEval}} & \multicolumn{2}{c}{\textbf{MBPP}} \\
 & \multicolumn{1}{r|}{} & \textit{HE} & \multicolumn{1}{c|}{\textit{HE$^+$}} & \textit{MBPP} & \textit{MBPP$^+$} \\ \midrule
\multicolumn{6}{c}{\textasciitilde 8B Models} \\ \midrule
StarCoder2-7B & \multicolumn{1}{r|}{7B} & 35.4 & \multicolumn{1}{c|}{29.9} & 54.4 & 45.6 \\
DeepSeek-Coder-6.7B-Base & \multicolumn{1}{r|}{6.7B} & 47.6 & \multicolumn{1}{c|}{39.6} & 70.2 & 56.6 \\
CodeQwen1.5-7B & \multicolumn{1}{r|}{7B} & 51.8 & \multicolumn{1}{c|}{45.7} & 72.2 & 60.2 \\
OpenCoder-8B-Base & \multicolumn{1}{r|}{8B} & 66.5 & \multicolumn{1}{c|}{63.4} & 79.9 & \fst{70.4} \\
Qwen2.5-Coder-7B & \multicolumn{1}{r|}{7B} & 72.0 & \multicolumn{1}{c|}{67.1} & 79.4 & 68.3 \\
\graybg \ourmodel-8B-Base & \multicolumn{1}{r|}{8B} & \fst{77.4} & \multicolumn{1}{c|}{\fst{68.3}} & \fst{82.0} & 69.0 \\ \midrule
\multicolumn{6}{c}{\grayt 13B+ Models} \\ \midrule
\grayt StarCoder2-15B & \multicolumn{1}{r|}{\grayt 15B} & \grayt 46.3 & \multicolumn{1}{c|}{\grayt 37.8} & \grayt 66.2 & \grayt 53.1 \\
\grayt CodeLlama-70B-Base & \multicolumn{1}{r|}{\grayt 70B} & \grayt 52.4 & \multicolumn{1}{c|}{\grayt 50.6} & \grayt 71.0 & \grayt 65.6 \\
\grayt Llama-3.1-70B-Base & \multicolumn{1}{r|}{\grayt 70B} & \grayt 54.9 & \multicolumn{1}{c|}{\grayt 51.2} & \grayt 81.4 & \grayt 67.2 \\
\grayt DeepSeek-Coder-33B-Base & \multicolumn{1}{r|}{\grayt 33B} & \grayt 54.9 & \multicolumn{1}{c|}{\grayt 47.6} & \grayt 74.2 & \grayt 60.7 \\
\grayt DeepSeek-Coder-V2-Lite-Base & \multicolumn{1}{r|}{\grayt 2.4B/16B} & \grayt 40.9 & \multicolumn{1}{c|}{\grayt 34.1} & \grayt 71.9 & \grayt 59.4 \\
\grayt Qwen2.5-Coder-14B & \multicolumn{1}{r|}{\grayt 14B} & \grayt \fst{83.5} & \multicolumn{1}{c|}{\grayt \fst{75.6}} & \grayt \fst{83.6} & \grayt \fst{69.8} \\
\bottomrule
\end{tabular}
}
\caption{Performance of various base models on HumanEval($^+$) and MBPP($^+$).}
\label{tab:base_humaneval_mbpp}
\end{table}

\begin{table}[]
\vspace{-10pt}
\centering
\resizebox{\textwidth}{!}{
\begin{tabular}{lrcccccccc|c}
\toprule
\textbf{Model} & \multicolumn{1}{r|}{\textbf{Size}} & Python & C++ & Java & PHP & TS & C\# & Bash & JS & \textbf{Average} \\ \midrule
\multicolumn{11}{c}{\textasciitilde 8B Models} \\ \midrule
StarCoder2-7B & \multicolumn{1}{r|}{7B} & 35.4 & 40.4 & 38.0 & 30.4 & 34.0 & 46.2 & 13.9 & 36.0 & 34.3 \\
DeepSeek-Coder-6.7B-Base & \multicolumn{1}{r|}{6.7B} & 49.4 & 50.3 & 43.0 & 38.5 & 49.7 & 50.0 & 28.5 & 48.4 & 44.7 \\
CodeQwen1.5-7B & \multicolumn{1}{r|}{7B} & 51.8 & 52.2 & 42.4 & 46.6 & 52.2 & 55.7 & 36.7 & 49.7 & 48.4 \\
OpenCoder-8B-Base & \multicolumn{1}{r|}{8B} & 66.5 & 63.4 & 63.9 & 61.5 & 68.6 & 54.3 & 44.3 & 65.8 & 61.0 \\
Qwen2.5-Coder-7B & \multicolumn{1}{r|}{7B} & 72.0 & 62.1 & 53.2 & 59.0 & 64.2 & \fst{60.8} & 38.6 & 60.3 & 58.8 \\
\graybg \ourmodel-8B-Base & \multicolumn{1}{r|}{8B} & \fst{77.4} & \fst{69.6} & \fst{72.8} & \fst{63.9} & \fst{77.4} & 53.8 & \fst{48.1} & \fst{77.6} & \fst{67.6} \\ \midrule
\multicolumn{11}{c}{\grayt 13B+ Models} \\ \midrule
\grayt StarCoder2-15B & \multicolumn{1}{r|}{\grayt 15B} & \grayt 46.3 & \grayt 47.2 & \grayt 46.2 & \grayt 39.1 & \grayt 42.1 & \grayt 53.2 & \grayt 15.8 & \grayt 43.5 & \grayt 41.7 \\
\grayt CodeLlama-70B-Base & \multicolumn{1}{r|}{\grayt 70B} & \grayt 52.4 & \grayt 49.7 & \grayt 44.7 & \grayt 46.6 & \grayt 57.2 & \grayt 46.7 & \grayt 31.6 & \grayt 56.5 & \grayt 48.2 \\
\grayt Llama-3.1-70B-Base & \multicolumn{1}{r|}{\grayt 70B} & \grayt 54.9 & \grayt 41.0 & \grayt 41.1 & \grayt 48.4 & \grayt 57.9 & \grayt 44.2 & \grayt 29.1 & \grayt 55.3 & \grayt 46.5  \\
\grayt DeepSeek-Coder-33B-Base & \multicolumn{1}{r|}{\grayt 33B} & \grayt 56.1 & \grayt 58.4 & \grayt \fst{51.9} & \grayt 44.1 & \grayt 52.8 & \grayt 51.3 & \grayt 32.3 & \grayt 55.3 & \grayt 50.3 \\
\grayt DeepSeek-Coder-V2-Lite-Base & \multicolumn{1}{r|}{\grayt 2.4B/16B} & \grayt 40.9 & \grayt 45.9 & \grayt 34.8 & \grayt 47.2 & \grayt 48.4 & \grayt 41.7 & \grayt 19.6 & \grayt 44.7 & \grayt 40.4 \\ 
\grayt Qwen2.5-Coder-14B & \multicolumn{1}{r|}{\grayt 14B} & \grayt \fst{83.5} & \grayt \fst{69.6} & \grayt 46.8 & \grayt \fst{64.6} & \grayt \fst{69.2} & \grayt \fst{63.3} & \grayt \fst{39.9} & \grayt \fst{61.5} & \grayt \fst{62.3} \\
\bottomrule
\end{tabular}
}
\caption{Performance of base models on MultiPL-E.}
\label{tab:base_multipl-e}
\end{table}

\subsubsection{Code Generation} \label{subsubsec:eval_base_codegen}

\noindent \textbf{HumanEval}~\citep{chen2021evaluating} consists of 164 manually written programming tasks in Python, each of which provides a function signature with a docstring for LLMs to complete a solution, whose functional correctness is judged by a handful of test cases. To rigorously benchmark the correctness of LLM-generated code, EvalPlus~\citep{liu2023evalplus} extends the test cases of the original HumanEval by 80$\times$ to build HumanEval$^+$. We used EvalPlus to evaluate base models on both HumanEval and HumanEval$^+$.

\noindent \textbf{MBPP}~\citep{austin2021program} is created by crowd-sourcing participants to write 974 Python programming problems, each of which is comprised of a function signature with a docstring, as well as a few test cases. We adopt the human-verified version of MBPP in EvalPlus, which is a subset of 399 tasks that are verified to be well-formed. Similarly, to accurately reflect the true performance of LLMs, we also evaluated base models on MBPP$^+$ with 35$\times$ more test cases than the original MBPP.

\noindent The evaluation results of HumanEval($^+$) and MBPP($^+$) are presented in Table~\ref{tab:base_humaneval_mbpp}. When using EvalPlus to evaluate Qwen2.5-Coder series, we found that although these models were supposed to be base models, they came with an official chat template and their performance on HumanEval and MBPP would be significantly boosted when the chat template was applied. Without the chat template applied, EvalPlus reports similar base model performance as in the Qwen2.5-Coder report~\citep{qwen25coder}. To reflect the genuine ability of Qwen2.5-Coder series, we decide to report their higher EvalPlus scores with the chat template applied. For \ourmodel-8B-Base, which has no chat template, we report its EvalPlus scores under the normal base prompt setting. As shown in Table~\ref{tab:base_humaneval_mbpp}, \ourmodel-8B-Base achieves impressive performance among open-source models of similar size, and even surpasses some much larger models.

\noindent \textbf{MultiPL-E.} Besides Python, we also evaluated base models on their code generation abilities across other popular programming languages. MultiPL-E~\citep{cassano2023multiple} extends HumanEval by translating Python tasks and unit tests into 18 additional programming languages, including C++, Java, PHP, TypeScript (TS), C\#, Bash, JavaScript (JS), etc. Evaluating models on MultiPL-E provides a comprehensive overview of how they work across various languages and helps understand the language factors that may affect the coding performance of LLMs. Following Qwen2.5-Coder~\citep{qwen25coder}, we chose eight mainstream programming languages from MultiPL-E for evaluation and present the results in Table~\ref{tab:base_multipl-e}. On 7 out of 8 programming languages, \ourmodel-8B-Base outperforms other open-source models of comparable size, and on some languages, even outperforms models with over 13B parameters. This shows the superior multilingual code generation abilities of \ourmodel-8B-Base.

\subsubsection{Code Completion}
Code completion is one of the most essential features that code LLMs provide to developers, enabling the suggestion of code snippets based on both preceding and succeeding context. We evaluated the code completion performance of base models using three benchmarks -- CrossCodeEval, RepoEval and Single-line MultiPL-HumanEval FIM -- each assessed under the FIM format.

\begin{table}[]
\centering
\resizebox{\textwidth}{!}{
\begin{tabular}{lrcccccccc|cc}
\toprule
\multirow{2}{*}{\textbf{Model}} & \multicolumn{1}{r|}{\multirow{2}{*}{\textbf{Size}}} & \multicolumn{2}{c}{Python} & \multicolumn{2}{c}{Java} & \multicolumn{2}{c}{TypeScript} & \multicolumn{2}{c}{C\#} & \multicolumn{2}{|c}{\textbf{Average}} \\ 
\cmidrule(lr){3-4}\cmidrule(lr){5-6}\cmidrule(lr){7-8}\cmidrule(lr){9-10}\cmidrule(lr){11-12}
 & \multicolumn{1}{r|}{} & \textit{EM} & \textit{ES} & \textit{EM} & \textit{ES} & \textit{EM} & \textit{ES} & \textit{EM} & \textit{ES} & \textit{EM} & \textit{ES} \\ \midrule
\multicolumn{12}{c}{\textasciitilde 8B Models} \\ \midrule
StarCoder2-7B & \multicolumn{1}{r|}{7B} & 10.9 & 63.1 & 8.3 & 71.0 & 6.7 & 76.8 & 7.3 & 72.1 & 8.3 & 70.8 \\
DeepSeek-Coder-6.7B-Base & \multicolumn{1}{r|}{6.7B} & 41.1 & 79.2 & 39.9 & 80.1 & 46.3 & 82.4 & 55.0 & 86.9 & 45.6 & 82.1 \\
CodeQwen1.5-7B & \multicolumn{1}{r|}{7B} & 40.7 & 77.8 & 47.0 & 81.6 & 45.8 & 82.2 & 59.7 & 87.6 & 48.3 & 82.3 \\
Qwen2.5-Coder-7B & \multicolumn{1}{r|}{7B} & 42.4 & 78.6 & 48.1 & 82.6 & 46.8 & 83.4 & 59.7 & 87.9 & 49.3 & 83.1 \\
\graybg \ourmodel-8B-Base & \multicolumn{1}{r|}{8B} & \fst{49.5} & \fst{82.0} & \fst{52.6} & \fst{84.5} & \fst{50.5} & \fst{84.6} & \fst{62.0} & \fst{89.1} & \fst{53.7} & \fst{85.1} \\ \midrule
\multicolumn{12}{c}{\grayt 13B+ Models} \\ \midrule
\grayt StarCoder2-15B & \multicolumn{1}{r|}{\grayt 15B} & \grayt 28.2 & \grayt 70.5 & \grayt 26.7 & \grayt 71.0 & \grayt 24.7 & \grayt 76.3 & \grayt 25.2 & \grayt 74.2 & \grayt 26.2 & \grayt 73.0 \\
\grayt DeepSeek-Coder-33B-Base & \multicolumn{1}{r|}{\grayt 13B} & \grayt 44.2 & \grayt 80.4 & \grayt 46.5 & \grayt 82.7 & \grayt 49.2 & \grayt 84.0 & \grayt 55.2 & \grayt 87.8 & \grayt 48.8 & \grayt 83.7 \\
\grayt DeepSeek-Coder-V2-Lite-Base & \multicolumn{1}{r|}{\grayt 2.4B/16B} & \grayt 41.8 & \grayt 78.3 & \grayt 46.1 & \grayt 81.2 & \grayt 44.6 & \grayt 81.4 & \grayt 58.7 & \grayt 87.9 & \grayt 47.8 & \grayt 82.2 \\
\grayt Qwen2.5-Coder-14B & \multicolumn{1}{r|}{\grayt 14B} & \grayt 47.7 & \grayt 81.7 & \grayt 54.7 & \grayt 85.7 & \grayt 52.9 & \grayt 86.0 & \grayt 66.4 & \grayt 91.1 & \grayt 55.4 & \grayt 86.1 \\
\grayt Qwen2.5-Coder-32B & \multicolumn{1}{r|}{\grayt 32B} & \grayt \fst{49.2} & \grayt \fst{82.1} & \grayt \fst{56.4} & \grayt \fst{86.6} & \grayt \fst{54.9} & \grayt \fst{87.0} & \grayt \fst{68.0} & \grayt \fst{91.6} & \grayt \fst{57.1} & \grayt \fst{86.8} \\
\bottomrule
\end{tabular}
}
\caption{Performance of base models on CrossCodeEval.}
\label{tab:fim_cceval}
\end{table}

\begin{table}[!ht]
\centering
\resizebox{0.85\textwidth}{!}{
\begin{tabular}{lrcccccc|cc}
\toprule
\multirow{2}{*}{\textbf{Model}} & \multicolumn{1}{r|}{\multirow{2}{*}{\textbf{Size}}} & \multicolumn{2}{c}{Line} & \multicolumn{2}{c}{Function} & \multicolumn{2}{c}{API} & \multicolumn{2}{|c}{\textbf{Average}} \\
\cmidrule(lr){3-4}\cmidrule(lr){5-6}\cmidrule(lr){7-8}\cmidrule(lr){9-10}
 & \multicolumn{1}{r|}{} & \textit{EM} & \textit{ES} & \textit{EM} & \textit{ES} & \textit{EM} & \textit{ES} & \textit{EM} & \textit{ES} \\ \midrule
\multicolumn{10}{c}{\textasciitilde 8B Models} \\ \midrule
StarCoder2-7B & \multicolumn{1}{r|}{7B} & 19.5 & 67.6 & 4.0 & 53.5 & 19.1 & 72.8 & 14.2 & 64.7 \\
DeepSeek-Coder-6.7B-Base & \multicolumn{1}{r|}{6.7B} & 63.1 & 85.5 & 9.9 & 53.3 & 52.3 & 81.7 & 41.7 & 73.5 \\
CodeQwen1.5-7B & \multicolumn{1}{r|}{7B} & 59.7 & 81.5 & 4.8 & 44.3 & 46.1 & 77.5 & 36.9 & 67.8 \\
Qwen2.5-Coder-7B & \multicolumn{1}{r|}{7B} & 67.3 & 86.1 & 13.2 & \fst{55.2} & 58.4 & 83.9 & 46.3 & 75.1 \\
\graybg \ourmodel-8B-Base & \multicolumn{1}{r|}{8B} & \fst{72.5} & \fst{88.7} & \fst{19.3} & 53.0 & \fst{60.8} & \fst{85.0} & \fst{50.8} & \fst{75.6} \\ \midrule
\multicolumn{10}{c}{\grayt 13B+ Models} \\ \midrule
\grayt StarCoder2-15B & \multicolumn{1}{r|}{\grayt 15B} & \grayt 30.9 & \grayt 62.5 & \grayt 5.5 & \grayt 43.7 & \grayt 21.7 & \grayt 60.3 & \grayt 19.4 & \grayt 55.5 \\
\grayt DeepSeek-Coder-33B-Base & \multicolumn{1}{r|}{\grayt 13B} & \grayt 66.5 & \grayt 86.6 & \grayt 10.3 & \grayt 52.9 & \grayt 54.2 & \grayt 83.5 & \grayt 43.7 & \grayt 74.3 \\
\grayt DeepSeek-Coder-V2-Lite-Base & \multicolumn{1}{r|}{\grayt 2.4B/16B} & \grayt 66.5 & \grayt 85.4 & \grayt 10.8 & \grayt 53.9 & \grayt 53.1 & \grayt 81.3 & \grayt 43.4 & \grayt 73.5 \\
\grayt Qwen2.5-Coder-14B & \multicolumn{1}{r|}{\grayt 14B} & \grayt 74.3 & \grayt 90.1 & \grayt \fst{14.1} & \grayt \fst{59.5} & \grayt 63.4 & \grayt 87.3 & \grayt 50.6 & \grayt \fst{79.0} \\
\grayt Qwen2.5-Coder-32B & \multicolumn{1}{r|}{\grayt 32B} & \grayt \fst{76.1} & \grayt \fst{90.5} & \grayt 13.6 & \grayt 57.5 & \grayt \fst{65.1} & \grayt \fst{87.6} & \grayt \fst{51.6} & \grayt 78.5 \\
\bottomrule
\end{tabular}
}
\caption{Performance of base models on RepoEval.}
\label{tab:fim_repoeval}
\end{table}

\noindent \textbf{CrossCodeEval}~\citep{ding2023cceval} is composed of single-line infilling tasks across Python, TypeScript, C\# and Java. In addition to the prefix and suffix context in the same file as the reference, it provides cross-file-related codes to create a self-contained code snippet. For evaluation parameters, we set the maximum sequence length to $8,192$ tokens, limited the maximum generated length to 50 tokens, and added cross-file context of no more than $2,048$ tokens, computed under Retrieval w/ Ref. mode as in the original paper of CrossCodeEval. Table~\ref{tab:fim_cceval} summarizes the performance of base models under the Exact Match (EM) and Edit Similarity (ES) metrics. The average is an unweighted mean of language-wise results.

\noindent \textbf{RepoEval}~\citep{zhang2023repocoder} features API invocation and function-body completion alongside single-line completion tasks. It is a Python-oriented benchmark reorganized from selected high-quality GitHub repositories. We set the maximum sequence length to $8,192$ tokens and the maximum cross-file context length to $2,048$ tokens for all evaluation tasks. For API invocation and single-line completion, the number of predicted tokens was limited to 50, and for function-body completion, 256. Cross-file context was retrieved with the oracle method aligned to the original paper. The performance of base models on RepoEval is shown in Table~\ref{tab:fim_repoeval}. EM and ES remain as the metrics, and the average is computed with uniform weights from language-wise results.

\begin{table}[!h]
\centering
\resizebox{0.8\textwidth}{!}{
\begin{tabular}{lr|ccc|c}
\toprule
\multicolumn{1}{l}{\textbf{Model}} & \textbf{Size} & Python & Java & JavaScript & \textbf{Average}\\
\midrule
\multicolumn{6}{c}{\textasciitilde 8B Models}\\
\midrule
StarCoder2-7B & 7B & 70.8 & 86.0 & 84.4 & 82.0 \\
DeepSeek-Coder-6.7B-Base & 6.7B & 78.1 & 87.4 & 84.1 & 84.0 \\
CodeQwen1.5-7B & 7B & 75.8 & 85.7 & 85.0 & 83.3 \\
Qwen2.5-Coder-7B & 7B & \fst{79.7} & \fst{88.5} & \fst{87.6} & \fst{86.2} \\
\graybg \ourmodel-8B-Base & 8B & 77.1 & 84.5 & 83.2 & 82.4 \\
\midrule
\multicolumn{6}{c}{\grayt 13B+ Models}\\
\midrule
\grayt StarCoder2-15B & \grayt 15B & \grayt 74.2 & \grayt 85.2 & \grayt 84.6 & \grayt 82.6 \\
\grayt DeepSeek-Coder-33B-Base & \grayt 33B & \grayt 80.1 & \grayt 89.0 & \grayt 86.8 & \grayt 86.2 \\
\grayt DeepSeek-Coder-V2-Lite-Base & \grayt 2.4B/16B & \grayt 78.7 & \grayt 87.8 & \grayt 85.9 & \grayt 85.0 \\
\grayt Qwen2.5-Coder-14B & \grayt 14B & \grayt 80.5 & \grayt \fst{91.0} & \grayt 88.5 & \grayt 87.7 \\
\grayt Qwen2.5-Coder-32B & \grayt 32B & \grayt \fst{81.5} & \grayt \fst{91.0} & \grayt \fst{89.4} & \grayt \fst{88.3} \\
\bottomrule
\end{tabular}
}
\caption{Exact Match scores of base models on the task of single-line MultiPL-HumanEval FIM.}
\label{tab:fim_santacoder}
\end{table}

\begin{table}[!h]
\centering
\resizebox{0.68\textwidth}{!}{
\begin{tabular}{lrcc}
\toprule
\textbf{Model} & \multicolumn{1}{r|}{\textbf{Size}} & Input-CoT & Output-CoT \\ \midrule
\multicolumn{4}{c}{\textasciitilde 8B Models} \\ \midrule
StarCoder2-7B & \multicolumn{1}{r|}{7B} & 39.5 & 35.1 \\
DeepSeek-Coder-6.7B-Base & \multicolumn{1}{r|}{6.7B} & 39.0 & 41.0 \\
CodeQwen1.5-7B & \multicolumn{1}{r|}{7B} & 44.8 & 40.1 \\
OpenCoder-8B-Base & \multicolumn{1}{r|}{8B} & 43.3 & 43.9 \\
Qwen2.5-Coder-7B & \multicolumn{1}{r|}{7B} & \fst{56.5} & \fst{56.0} \\
\graybg \ourmodel-8B-Base & \multicolumn{1}{r|}{8B} & 52.0 & 54.8 \\ \midrule
\multicolumn{4}{c}{\grayt 13B+ Models} \\ \midrule
\grayt StarCoder2-15B & \multicolumn{1}{r|}{\grayt 15B} & \grayt 46.1 & \grayt 47.6 \\
\grayt CodeLlama-34B-Base & \multicolumn{1}{r|}{\grayt 34B} & \grayt 49.4 & \grayt 43.9 \\
\grayt DeepSeek-Coder-33B-Base & \multicolumn{1}{r|}{\grayt 33B} & \grayt 50.6 & \grayt 48.8 \\
\grayt DeepSeek-Coder-V2-Lite-Base & \multicolumn{1}{r|}{\grayt 2.4B/16B} & \grayt 53.4 & \grayt 46.1 \\
\grayt Qwen2.5-Coder-14B & \multicolumn{1}{r|}{\grayt 14B} & \grayt \fst{60.6} & \grayt \fst{66.4} \\
\bottomrule
\end{tabular}
}
\caption{Performance of base models on CRUXEval.}
\label{tab:base_cruxeval}
\end{table}

\noindent \textbf{Single-line MultiPL-HumanEval FIM}~\citep{allal2023santacoder} refines MultiPL-E~\citep{cassano2023multiple} into a set of single-line infilling challenges across Python, Java, and JavaScript, which serves as a classic benchmark for FIM evaluation. Performance is assessed using the EM metric~\citep{fried2022incoder}. Table~\ref{tab:fim_santacoder} presents the results on Single-line MultiPL-HumanEval FIM, comparing the models by size. The average is weighted by the number of samples from each language.

\noindent Although the EM and ES metrics may not fully capture true performance since semantically equivalent statements can be expressed in various ways, \ourmodel-8B-Base demonstrates superiority across the \textasciitilde8B model family.

\subsubsection{Code Reasoning} \label{subsubsec:eval_base_codereason}

\noindent \textbf{CRUXEval.} Compared to natural language, code is highly sensitive to even minor modifications, including single-token changes. Therefore, effective code reasoning demands precise token-level understanding and the capacity to internally simulate or trace the behavior of code. To evaluate the capability of code reasoning for base models, we utilized the CRUXEval~\citep{gu2024cruxeval} benchmark, which comprises 800 Python functions paired with corresponding input-output examples. CRUXEval is divided into two distinct tasks: CRUXEval-I, in which the model predicts the output from a given input, and CRUXEval-O, in which the model infers inputs from provided outputs. This structure tests the model's capability to comprehend and reason about Python code in both forward and backward directions.

\noindent Table~\ref{tab:base_cruxeval} presents the results for both tasks evaluated in the Chain-of-Thought (CoT) mode, where the model is prompted to reason step by step during simulated code execution. Notably, in both the Input-CoT and Output-CoT settings, \ourmodel-8B-Instruct achieves top-tier performance among \textasciitilde 8B models, only marginally trailing Qwen2.5-Coder-7B. Note that both our model and Qwen2.5-Coder-7B surpass some much larger models such as DeepSeek-Coder-33B-Base.

\subsubsection{Long-Context Capability}
\begin{figure}[]
    \centering
    \includegraphics[width=0.65\linewidth]{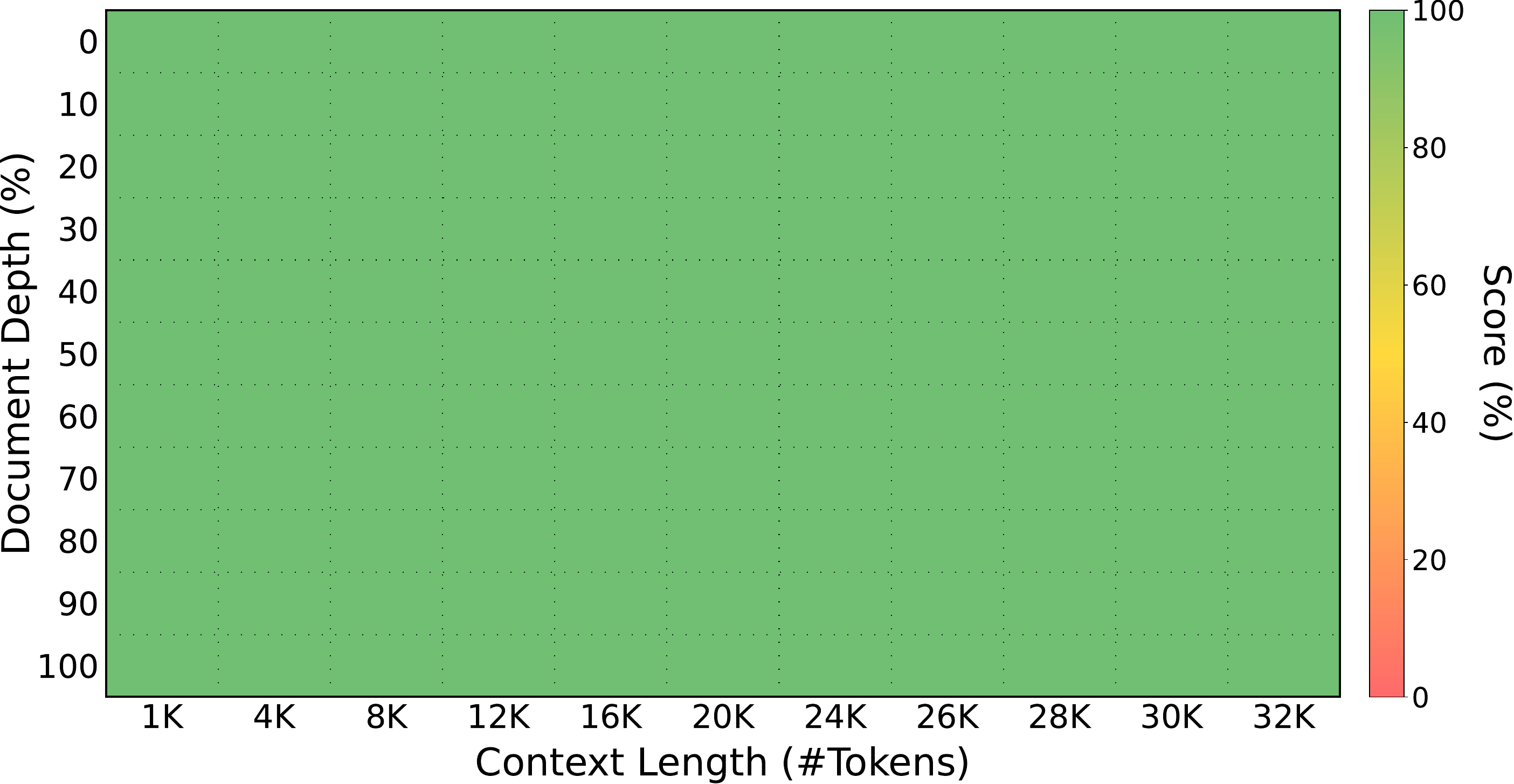}
    \caption{Evaluation results of the ``Needle in the Code'' pressure test.}
    \label{fig:needle}
\end{figure}

\noindent \textbf{Needle in the Code.} To unlock the potential of our code model in understanding repository-level code contexts and benefit real-world software development, it is vital to involve long-context capability in the pretraining phase. To this end, we build a ``Needle in the Code'' pressure test. The needle construction method is a specialized evaluation technique designed to assess LLM's ability to handle long contexts. This approach involves carefully placing a relevant code snippet (the ``needle'') in broader code functions (the ``haystack''). The model is then presented with a specific query related to the embedded code and is required to find and identify the particular code snippet within the larger functional context. Figure~\ref{fig:needle} shows that \ourmodel-8B-Base achieves 100\% accuracy in the ``Needle in the Code'' pressure test under 32K context length.

\subsection{Evaluation of Instruct Models}

In this section, we report comprehensive evaluation results of our instruct model \ourmodel-8B-Instruct and previous state-of-the-art open-source models on a wide range of coding tasks, including code generation, code reasoning, code editing, and software engineering. To guarantee reproducibility, we utilized either official benchmark leaderboards or the open-source evaluation suites provided by Qwen2.5-Coder~\citep{qwen25coder}, incorporating their published performance metrics whenever available, and performing our own evaluations for metrics not previously reported.

\subsubsection{Code Generation}

\begin{table}[]
\centering
\resizebox{\textwidth}{!}{
\begin{tabular}{lr|cc|cc|cc|c|c}
\toprule
\multirow{2}{*}{\textbf{Model}} & \multicolumn{1}{r|}{\multirow{2}{*}{\textbf{Size}}} & \multicolumn{2}{c|}{\textbf{HumanEval}} & \multicolumn{2}{c|}{\textbf{MBPP}} & \multicolumn{1}{c|}{\textbf{MHPP}} & \multicolumn{2}{c|}{\textbf{BigCodeBench}} & \textbf{LiveCodeBench} \\
 & \multicolumn{1}{c|}{} & \textit{HE} & \multicolumn{1}{c|}{\textit{HE+}} & \textit{MBPP} & \multicolumn{1}{c|}{\textit{MBPP+}} & \multicolumn{1}{c|}{\textit{pass@1}} & \multicolumn{1}{c}{\textit{\ \ \ Full}} & \multicolumn{1}{c|}{\textit{Hard}} & \textit{2410 -- 2502} \\ \midrule
\multicolumn{10}{c}{\textasciitilde 8B Models} \\ \midrule
CodeLlama-7B-Instruct & \multicolumn{1}{r|}{7B} & 40.9 & \multicolumn{1}{c|}{33.5} & 54.0 & \multicolumn{1}{c|}{44.4} & \multicolumn{1}{c|}{6.7} & \multicolumn{1}{c}{\ \ \ 25.7} & \multicolumn{1}{c|}{4.1} & \multicolumn{1}{c}{3.6} \\
DeepSeek-Coder-6.7B-Instruct & \multicolumn{1}{r|}{6.7B} & \multicolumn{1}{c}{74.4} & \multicolumn{1}{c|}{71.3} & \multicolumn{1}{c}{74.9} & \multicolumn{1}{c|}{65.6} & \multicolumn{1}{c|}{20.0} & \multicolumn{1}{c}{\ \ \ 43.8} & \multicolumn{1}{c|}{15.5} & \multicolumn{1}{c}{9.6} \\
CodeQwen1.5-7B-Chat & \multicolumn{1}{r|}{7B} & \multicolumn{1}{c}{83.5} & \multicolumn{1}{c|}{78.7} & \multicolumn{1}{c}{77.7} & \multicolumn{1}{c|}{67.2} & \multicolumn{1}{c|}{17.6} & \multicolumn{1}{c}{\ \ \ 43.6} & \multicolumn{1}{c|}{15.5} & \multicolumn{1}{c}{3.0} \\
Yi-Coder-9B-Chat & \multicolumn{1}{r|}{9B} & \multicolumn{1}{c}{82.3} & \multicolumn{1}{c|}{74.4} & \multicolumn{1}{c}{82.0} & \multicolumn{1}{c|}{69.0} & \multicolumn{1}{c|}{26.7} & \multicolumn{1}{c}{\ \ \ 49.0} & \multicolumn{1}{c|}{17.6} & \multicolumn{1}{c}{17.5} \\
Llama-3.1-8B-Instruct & \multicolumn{1}{r|}{8B} & \multicolumn{1}{c}{68.3} & \multicolumn{1}{c|}{59.8} & \multicolumn{1}{c}{70.1} & \multicolumn{1}{c|}{59.0} & \multicolumn{1}{c|}{17.1} & \multicolumn{1}{c}{\ \ \ 40.5} & \multicolumn{1}{c|}{13.5} & \multicolumn{1}{c}{11.5} \\
OpenCoder-8B-Instruct & \multicolumn{1}{r|}{8B} & \multicolumn{1}{c}{83.5} & \multicolumn{1}{c|}{78.7} & \multicolumn{1}{c}{79.1} & \multicolumn{1}{c|}{69.0} & \multicolumn{1}{c|}{30.5} & \multicolumn{1}{c}{\ \ \ 50.9} & \multicolumn{1}{c|}{18.9} & \multicolumn{1}{c}{17.1} \\
Qwen2.5-Coder-7B-Instruct & \multicolumn{1}{r|}{7B} & \multicolumn{1}{c}{\fst{88.4}} & \multicolumn{1}{c|}{\fst{84.1}} & \multicolumn{1}{c}{83.5} & \multicolumn{1}{c|}{\fst{71.7}} & \multicolumn{1}{c|}{26.7} & \multicolumn{1}{c}{\ \ \ 48.8} & \multicolumn{1}{c|}{20.3} & \multicolumn{1}{c}{17.3} \\
Qwen3-8B & \multicolumn{1}{r|}{8B} & \multicolumn{1}{c}{84.8} & \multicolumn{1}{c|}{80.5} & \multicolumn{1}{c}{77.0} & \multicolumn{1}{c|}{67.2} & \multicolumn{1}{c|}{32.8} & \multicolumn{1}{c}{\ \ \ 51.7} & \multicolumn{1}{c|}{23.0} & \multicolumn{1}{c}{23.5} \\
\graybg \ourmodel-8B-Instruct & \multicolumn{1}{r|}{8B} & \multicolumn{1}{c}{84.8} & \multicolumn{1}{c|}{78.7} & \multicolumn{1}{c}{\fst{85.2}} & \multicolumn{1}{c|}{71.2} & \multicolumn{1}{c|}{\fst{36.2}} & \multicolumn{1}{c}{\ \ \ \fst{53.3}} & \multicolumn{1}{c|}{\fst{26.4}} & \multicolumn{1}{c}{\fst{24.7}} \\ \midrule
\multicolumn{10}{c}{\grayt 13B+ Models} \\ \midrule
\grayt StarCoder2-15B-Instruct & \multicolumn{1}{r|}{\grayt 15B} & \multicolumn{1}{c}{\grayt 67.7} & \multicolumn{1}{c|}{\grayt 60.4} & \multicolumn{1}{c}{\grayt 78.0} & \multicolumn{1}{c|}{\grayt 65.1} & \multicolumn{1}{c|}{\grayt 19.0} & \multicolumn{1}{c}{\ \ \ \grayt 45.1} & \multicolumn{1}{c|}{\grayt 14.9} & \multicolumn{1}{c}{\grayt 5.3} \\
\grayt Codestral-22B & \multicolumn{1}{r|}{\grayt 22B} & \multicolumn{1}{c}{\grayt 81.1} & \multicolumn{1}{c|}{\grayt 73.2} & \multicolumn{1}{c}{\grayt 78.2} & \multicolumn{1}{c|}{\grayt 62.2} & \multicolumn{1}{c|}{\grayt 25.2} & \multicolumn{1}{c}{\ \ \ \grayt 52.5} & \multicolumn{1}{c|}{\grayt 24.3} & \multicolumn{1}{c}{\grayt 20.5} \\
\grayt CodeLlama-70B-Instruct & \multicolumn{1}{r|}{\grayt 70B} & \multicolumn{1}{c}{\grayt 72.0} & \multicolumn{1}{c|}{\grayt 65.9} & \multicolumn{1}{c}{\grayt 77.8} & \multicolumn{1}{c|}{\grayt 64.6} & \multicolumn{1}{c|}{\grayt 19.5} & \multicolumn{1}{c}{\ \ \ \grayt 49.6} & \multicolumn{1}{c|}{\grayt 15.5} & \multicolumn{1}{c}{\grayt 14.5} \\
\grayt DeepSeek-Coder-33B-Instruct & \multicolumn{1}{r|}{\grayt 33B} & \multicolumn{1}{c}{\grayt 81.1} & \multicolumn{1}{c|}{\grayt 75.0} & \multicolumn{1}{c}{\grayt 80.4} & \multicolumn{1}{c|}{\grayt 70.1} & \multicolumn{1}{c|}{\grayt 32.9} & \multicolumn{1}{c}{\ \ \ \grayt 51.1} & \multicolumn{1}{c|}{\grayt 20.9} & \multicolumn{1}{c}{\grayt 14.5} \\ 
\grayt DeepSeek-Coder-V2-Lite-Instruct & \multicolumn{1}{r|}{\grayt 2.4B/16B} & \multicolumn{1}{c}{\grayt 81.1} & \multicolumn{1}{c|}{\grayt 75.6} & \multicolumn{1}{c}{\grayt 82.8} & \multicolumn{1}{c|}{\grayt 70.4} & \multicolumn{1}{c|}{\grayt 30.5} & \multicolumn{1}{c}{\ \ \ \grayt 47.6} & \multicolumn{1}{c|}{\grayt 18.2} & \multicolumn{1}{c}{\grayt 14.2} \\
\grayt DeepSeek-Coder-V2-Instruct & \multicolumn{1}{r|}{\grayt 21B/236B} & \multicolumn{1}{c}{\grayt 85.4} & \multicolumn{1}{c|}{\grayt 82.3} & \multicolumn{1}{c}{\grayt 89.4} & \multicolumn{1}{c|}{\grayt \fst{75.1}} & \multicolumn{1}{c|}{\grayt 31.9} & \multicolumn{1}{c}{\ \ \ \grayt \fst{59.7}} & \multicolumn{1}{c|}{\grayt \fst{33.1}} & \multicolumn{1}{c}{\grayt 28.9} \\
\grayt Qwen2.5-Coder-14B-Instruct & \multicolumn{1}{r|}{\grayt 14B} & \multicolumn{1}{c}{\grayt 89.6} & \multicolumn{1}{c|}{\grayt \fst{87.2}} & \multicolumn{1}{c}{\grayt 86.2} & \multicolumn{1}{c|}{\grayt 72.8} & \multicolumn{1}{c|}{\grayt 36.7} & \multicolumn{1}{c}{\ \ \ \grayt 52.2} & \multicolumn{1}{c|}{\grayt 16.2} & \multicolumn{1}{c}{\grayt 19.3} \\
\grayt Qwen2.5-Coder-32B-Instruct & \multicolumn{1}{r|}{\grayt 32B} & \multicolumn{1}{c}{\grayt \fst{92.7}} & \multicolumn{1}{c|}{\grayt \fst{87.2}} & \multicolumn{1}{c}{\grayt \fst{90.2}} & \multicolumn{1}{c|}{\grayt \fst{75.1}} & \multicolumn{1}{c|}{\grayt \fst{42.4}} & \multicolumn{1}{c}{\ \ \ \grayt 52.3} & \multicolumn{1}{c|}{\grayt 20.9} & \multicolumn{1}{c}{\grayt \fst{30.7}} \\
\bottomrule
\end{tabular}
}
    \captionsetup{justification=justified, singlelinecheck=false}
\caption{Performance of instruct models on HumanEval($^+$), MBPP($^+$), MHPP, BigCodeBench-Completion and LiveCodeBench (2410 -- 2502).}
\label{tab:instruct-codegen}
\end{table}

\begin{table}[t]
\vspace{-10pt}
\centering
\resizebox{\textwidth}{!}{
\begin{tabular}{lrccccccccccccc|c}
\toprule
\textbf{Model} & \multicolumn{1}{r|}{\textbf{Size}} & Python & Java & C++ & C\# & TS & JS & PHP & Go & Kotlin & Perl & Ruby & Scala & Swift & \textbf{Average} \\ \midrule
\multicolumn{16}{c}{\textasciitilde 8B Models} \\ \midrule
CodeLlama-7B-Instruct & \multicolumn{1}{r|}{7B} & 54.0 & 38.8 & 32.9 & 50.0 & 42.3 & 45.5 & 36.6 & 48.8 & 47.2 & 50.1 & 36.9 & 40.2 & 33.2 & 42.8 \\
DeepSeek-Coder-6.7B-Instruct & \multicolumn{1}{r|}{6.7B} & 74.9 & 52.2 & 30.9 & 55.9 & 64.8 & 64.7 & 25.8 & 93.8 & 59.6 & 3.3 & 65.9 & 54.8 & 47.4 & 53.4 \\
CodeQwen1.5-7B-Chat & \multicolumn{1}{r|}{7B} & 77.7 & 66.6 & 66.8 & 64.4 & 66.7 & 67.5 & 67.3 & 55.1 & 60.9 & 61.1 & 65.9 & 60.0 & 54.7 & 64.2 \\
Yi-Coder-9B-Chat & \multicolumn{1}{r|}{9B} & 82.0 & 73.4 & 79.1 & 70.3 & 74.1 & 73.3 & 76.4 & 90.9 & 64.4 & 60.9 & 67.3 & 63.5 & 57.3 & 71.8 \\
Llama-3.1-8B-Instruct & \multicolumn{1}{r|}{8B} & 70.1 & 59.8 & 59.1 & 56.6 & 59.1 & 59.1 & 62.5 & 85.7 & 52.2 & 42.6 & 55.9 & 44.5 & 31.8 & 56.8 \\
OpenCoder-8B-Instruct & \multicolumn{1}{r|}{8B} & 79.1 & 68.1 & 71.3 & 71.0 & 67.6 & 61.4 & 68.1 & 94.4 & 66.4 & 56.1 & 70.5 & 63.1 & 56.7 & 68.8 \\
Qwen2.5-Coder-7B-Instruct & \multicolumn{1}{r|}{7B} & 83.5 & 70.5 & 74.1 & 71.5 & 72.2 & 74.1 & 74.2 & \fst{96.0} & 65.5 & 64.4 & 75.5 & 64.2 & \fst{62.0} & 72.9 \\
Qwen3-8B & \multicolumn{1}{r|}{8B} & 77.0 & 69.0 & 72.8 & 68.9 & 73.0 & 73.8 & 72.3 & 92.9 & 62.0 & 64.6 & 69.0 & 63.1 & 42.2 & 69.3 \\
\graybg \ourmodel-8B-Instruct & \multicolumn{1}{r|}{8B} & \fst{85.2} & \fst{72.7} & \fst{77.0} & \fst{74.2} & \fst{72.8} & \fst{78.8} & \fst{74.7} & 95.5 & \fst{73.4} & \fst{72.5} & \fst{78.0} & \fst{70.3} & 54.2 & \fst{75.3} \\ \midrule
\multicolumn{16}{c}{\grayt 13B+ Models} \\ \midrule
\grayt StarCoder2-15B-Instruct & \multicolumn{1}{r|}{\grayt 15B} & \grayt 78.0 & \grayt 25.1 & \grayt 25.9 & \grayt 21.7 & \grayt 20.7 & \grayt 59.8 & \grayt 53.5 & \grayt 90.4 & \grayt 46.7 & \grayt 31.9 & \grayt 56.1 & \grayt 43.2 & \grayt 42.0 & \grayt 45.8 \\
\grayt Codestral-22B & \multicolumn{1}{r|}{\grayt 22B} & \grayt 78.2 & \grayt 73.6 & \grayt 77.3 & \grayt 70.1 & \grayt 71.7 & \grayt 68.5 & \grayt 74.9 & \grayt 97.1 & \grayt 71.0 & \grayt 66.6 & \grayt 74.2 & \grayt 64.4 & \grayt 50.1 & \grayt 72.1 \\
\grayt CodeLlama-70B-Instruct & \multicolumn{1}{r|}{\grayt 70B} & \grayt 77.8 & \grayt 66.6 & \grayt 68.6 & \grayt 69.2 & \grayt 47.8 & \grayt 62.5 & \grayt 70.5 & \grayt 77.7 & \grayt 57.2 & \grayt 51.1 & \grayt 67.0 & \grayt 51.3 & \grayt 48.7 & \grayt 62.8 \\
\grayt DeepSeek-Coder-33B-Instruct & \multicolumn{1}{r|}{\grayt 33B} & \grayt 80.4 & \grayt 71.8 & \grayt 76.8 & \grayt 69.9 & \grayt 72.4 & \grayt 69.8 & \grayt 75.1 & \grayt 96.4 & \grayt 70.1 & \grayt 66.6 & \grayt 75.1 & \grayt 64.6 & \grayt 54.3 & \grayt 72.6 \\
\grayt DeepSeek-Coder-V2-Lite-Instruct & \multicolumn{1}{r|}{\grayt 2.4B/16B} & \grayt 82.8 & \grayt 73.3 & \grayt 75.3 & \grayt 72.4 & \grayt 72.4 & \grayt 73.1 & \grayt 75.1 & \grayt 95.1 & \grayt 69.9 & \grayt 61.6 & \grayt 74.5 & \grayt 63.5 & \grayt 55.0 & \grayt 72.6 \\
\grayt DeepSeek-Coder-V2-Instruct & \multicolumn{1}{r|}{\grayt 21B/236B} & \grayt 89.4 & \grayt 78.2 & \grayt 77.6 & \grayt 72.6 & \grayt 74.8 & \grayt \fst{80.5} & \grayt 75.8 & \grayt 89.1 & \grayt 74.5 & \grayt 70.7 & \grayt 80.2 & \grayt 67.9 & \grayt 59.0 & \grayt 76.2 \\
\grayt Qwen2.5-Coder-14B-Instruct & \multicolumn{1}{r|}{\grayt 14B} & \grayt 86.2 & \grayt 77.5 & \grayt 84.8 & \grayt \fst{80.1} & \grayt 77.6 & \grayt 77.7 & \grayt 79.7 & \grayt \fst{97.1} & \grayt 75.3 & \grayt \fst{76.2} & \grayt 79.3 & \grayt \fst{73.1} & \grayt \fst{67.2} & \grayt 79.4 \\
\grayt Qwen2.5-Coder-32B-Instruct & \multicolumn{1}{r|}{\grayt 32B} & \grayt \fst{90.2} & \grayt \fst{80.4} & \grayt \fst{86.3} & \grayt 73.5 & \grayt \fst{78.3} & \grayt 79.3 & \grayt \fst{87.6} & \grayt 96.4 & \grayt \fst{75.6} & \grayt 74.7 & \grayt \fst{83.4} & \grayt 63.3 & \grayt 66.7 & \grayt \fst{79.7} \\
\bottomrule
\end{tabular}
}
\caption{Performance of instruct models on MBXP.}
\label{tab:instruct_mbxp}
\end{table}

\noindent \textbf{HumanEval and MBPP.} As introduced in Sec~\ref{subsubsec:eval_base_codegen}, we also evaluated basic code generation abilities of instruct models on HumanEval($^+$) and MBPP($^+$) using EvalPlus~\citep{liu2023evalplus}. As shown in Table~\ref{tab:instruct-codegen}, \ourmodel-8B-Instruct achieves competitive performance on these conventional code generation benchmarks.

\noindent \textbf{MHPP.} With the rapid advancement of code LLMs, traditional coding benchmarks such as HumanEval and MBPP are no longer sufficient to effectively differentiate between state-of-the-art models. High overall pass rates on these conventional benchmarks alone cannot reliably indicate whether a model possesses the complex reasoning abilities required to solve difficult coding problems.

\noindent Recently, a challenging benchmark named Mostly Hard Python Problems (MHPP) was introduced~\citep{dai2024mhppexploringcapabilitieslimitations}, consisting of 210 human-curated problems accompanied by unit tests. MHPP focuses on evaluating LLMs' abilities to tackle various challenges in code generation: handling variance in natural language inputs, understanding newly defined contexts, demonstrating commonsense reasoning, handling edge cases, following intricate instructions, applying mathematical and algorithmic knowledge, and exhibiting familiarity with coding principles. To compare the code generation performance of code LLMs under these challenges, we evaluated various instruct models on MHPP and report the results in Table~\ref{tab:instruct-codegen}. Our \ourmodel-8B-Instruct model achieves a striking 36.2\% pass@1 score, surpassing all the other \textasciitilde 8B models by a significant margin and even outperforming some much larger models such as DeepSeek-Coder-V2-Instruct with 236B total parameters.

\noindent \textbf{BigCodeBench.} To assess how well LLMs can solve challenging and real-world programming tasks, the recently proposed benchmark BigCodeBench~\citep{zhuo2025bigcodebench} challenges LLMs to invoke multiple function calls as tools from 139 libraries across 7 domains, encompassing 1,140 Python tasks with rich context. These high-quality programming tasks demand compositional reasoning and precise comprehension of complex instructions. Each task includes an average of 5.6 test cases, achieving 99\% branch coverage on average, thus ensuring rigorous and comprehensive evaluation of LLMs. We evaluated various instruct models on both the full set and the hard set of BigCodeBench-Completion, which is designed to showcase the ability of LLMs to complete coding tasks based on natural language instructions.

\noindent As shown in Table~\ref{tab:instruct-codegen}, our model \ourmodel-8B-Instruct surpasses other instruct models of comparable size, achieving superior pass@1 scores on both the full set (53.3\%) and the hard set (26.4\%). Even when compared to models with over 13B parameters, our 8B model outperforms some much larger models, demonstrating the strong code generation capabilities of \ourmodel-8B-Instruct on challenging and practical coding tasks.

\noindent \textbf{LiveCodeBench.} Static benchmarks for code generation may be subject to potential contamination or overfitting, leading to a skewed or misleading picture of LLMs' actual coding capabilities. To mitigate these issues, LiveCodeBench~\citep{livecodebench} introduces live updates by continuously collecting new problems from coding contests across prominent competitive programming platforms -- LeetCode, AtCoder, and CodeForces -- and tagging each problem with a release date. Using LiveCodeBench, we can evaluate LLMs on problems selected within a specified, up-to-date time window, effectively preventing contamination and ensuring fair comparisons. We set the LiveCodeBench time window to ``2410 -- 2502'' as it was the most recent available during our evaluation.

\noindent Table~\ref{tab:instruct-codegen} shows the results of various instruct models on LiveCodeBench within the same specified time window. Our model \ourmodel-8B-Instruct achieves a standout 24.7\% pass@1 score, demonstrating its exceptional performance among the \textasciitilde 8B models. Surprisingly, despite its smaller model size, \ourmodel-8B-Instruct even surpasses Qwen2.5-Coder-14B-Instruct and some other larger models on LiveCodeBench, highlighting the superior capability of our model in the field of competitive programming.

\noindent \textbf{MBXP.} The original MBPP benchmark only contains Python tasks, while MBXP~\citep{athiwaratkun2023multilingual} transforms the problems and unit tests in MBPP to 10+ programming languages. To thoroughly assess the multilingual code generation capabilities of LLMs, we evaluated various instruct models on MBXP. Table~\ref{tab:instruct_mbxp} presents a comprehensive performance overview of MBXP across 13 widely used programming languages. \ourmodel-8B-Instruct shows prominent results, achieving the highest scores in most languages among the \textasciitilde 8B models. With an average score of 75.3\%, our model even outperforms or matches the performance of some much larger models.

\begin{table}[t]
\centering
\resizebox{0.9\textwidth}{!}{
\begin{tabular}{lr|cccccc|c}
\toprule
\multirow{2}{*}{\textbf{Model}} & \multicolumn{1}{r|}{\multirow{2}{*}{\textbf{Size}}} & \multicolumn{3}{c}{NCB (zh)} & \multicolumn{3}{c|}{NCB (en)} & \multirow{2}{*}{\textbf{Total}} \\ \cmidrule(lr){3-5} \cmidrule(lr){6-8}
 & \multicolumn{1}{r|}{} & \textit{Python} & \textit{Java} & \multicolumn{1}{l}{\textit{Total}} & \textit{Python} & \textit{Java} & \multicolumn{1}{c}{\textit{Total}} & \multicolumn{1}{|c}{} \\ \midrule
\multicolumn{9}{c}{\textasciitilde 8B Models} \\ \midrule
CodeLlama-7B-Instruct & 7B & 18.6 & 8.6 & 13.6 & 17.1 & 14.3 & 15.7 & 14.6 \\
DeepSeek-Coder-6.7B-Instruct & 6.7B & 38.6 & 31.4 & 35.0 & 32.9 & 32.9 & 32.9 & 33.9 \\
CodeQwen1.5-7B-Chat & 7B & 30.0 & 28.6 & 29.3 & 30.0 & 27.1 & 28.6 & 25.7 \\
Yi-Coder-9B-Chat & 9B & 41.4 & \fst{45.7} & 43.6 & 38.6 & 44.3 & 41.5 & 42.5 \\
Llama-3.1-8B-Instruct & 8B & 27.1 & 24.3 & 25.7 & 22.9 & 22.9 & 22.9 & 24.3 \\
OpenCoder-8B-Instruct & 8B & 40.0 & 30.0 & 35.0 & 35.7 & 24.3 & 30.0 & 32.5 \\
Qwen2.5-Coder-7B-Instruct & 7B & 34.3 & 37.1 & 35.7 & 34.3 & 35.7 & 35.0 & 35.4 \\
Qwen3-8B & 8B & 37.1 & 32.9 & 35.0 & 34.3 & 38.6 & 36.5 & 35.7 \\
\graybg \ourmodel-8B-Instruct & 8B & \fst{55.7} & \fst{45.7} & \fst{50.7} & \fst{50.0} & \fst{47.1} & \fst{48.6} & \fst{49.6} \\ \midrule
\multicolumn{9}{c}{\grayt 13B+ Models} \\ \midrule
\grayt StarCoder2-15B-Instruct & \grayt 15B & \grayt 44.3 & \grayt 30.0 & \grayt 37.2 & \grayt 38.6 & \grayt 42.9 & \grayt 40.8 & \grayt 39.0 \\
\grayt Codestral-22B & \grayt 22B & \grayt 40.0 & \grayt 44.3 & \grayt 42.2 & \grayt 41.4 & \grayt \fst{45.7} & \grayt 43.6 & \grayt 42.9 \\
\grayt CodeLlama-70B-Instruct & \grayt 70B & \grayt 35.1 & \grayt 32.1 & \grayt 33.6 & \grayt 32.8 & \grayt 30.5 & \grayt 31.7 & \grayt 32.6 \\
\grayt DeepSeek-Coder-33B-Instruct & \grayt 33B & \grayt 44.3 & \grayt 38.9 & \grayt 41.6 & \grayt \fst{44.3} & \grayt 44.3 & \grayt \fst{44.3} & \grayt 43.0 \\
\grayt DeepSeek-Coder-V2-Lite-Instruct & \grayt 2.4B/16B & \grayt 41.4 & \grayt 47.1 & \grayt 44.3 & \grayt 41.4 & \grayt 37.1 & \grayt 39.3 & \grayt 41.8 \\
\grayt Qwen2.5-Coder-14B-Instruct & \grayt 14B & \grayt \fst{48.6} & \grayt \fst{48.6} & \grayt \fst{48.6} & \grayt 42.9 & \grayt \fst{45.7} & \grayt \fst{44.3} & \grayt \fst{46.4} \\
\bottomrule
\end{tabular}
}
\caption{Performance of instruct models on NaturalCodeBench.}
\label{tab:instruct_naturalcodebench}
\end{table}

\noindent \textbf{NaturalCodeBench.} Traditional code generation benchmarks, such as HumanEval and MBPP, are often limited to well-defined coding problems in algorithmic and basic programming, which may fall short in sufficiently capturing the wide range of needs and complexity in practical software engineering problems. To bridge this gap, NaturalCodeBench~\citep{zhang2024naturalcodebenchexaminingcodingperformance} is designed to mirror the complexity and variety of scenarios in real-world coding tasks. NaturalCodeBench (NCB) meticulously curates a collection of 402 high-quality problems derived from authentic user queries on popular online coding platforms. These problems, available in Python and Java, span six domains: front-end development, algorithms, data science, artificial intelligence, software engineering, and system administration. Beyond basic data structures such as lists and numbers, the test inputs for NCB problems incorporate diverse file types and complex structures, providing a greater challenge for code LLMs. We evaluate various instruct models on NCB to investigate their abilities to solve real-world coding problems.

\noindent We evaluated the performance of various instruct models on both the Chinese (zh) and the English (en) versions of NaturalCodeBench, with the results summarized in Table~\ref{tab:instruct_naturalcodebench}. As shown in the table, \ourmodel-8B-Instruct consistently exhibits dominant performance in the Python split, outperforming not only all the other listed \textasciitilde 8B models but also larger models with 13B+ parameters, such as Qwen2.5-Coder-14B-Instruct. Similarly, for the Java split, \ourmodel-8B-Instruct achieves remarkable results compared to existing open-source code LLMs. Notably, \ourmodel-8B-Instruct attains an overall score of 49.6\% with 8B parameters, exceeding the 46.4\% achieved by Qwen2.5-Coder-14B-Instruct with 14B parameters, which highlights the strong capability of our model in solving challenging real-world coding problems.

\begin{table}[]
\centering
\resizebox{\textwidth}{!}{
\begin{tabular}{lrcccccccccccc|c}
\toprule
\textbf{Model} & \multicolumn{1}{r|}{\textbf{Size}} & BP & AP & SE & DA & MA & DW & ML & SC & DB & MM & OS & Others & \textbf{Overall} \\ \midrule
\multicolumn{15}{c}{\textasciitilde 8B Models} \\ \midrule
CodeLlama-7B-Instruct & \multicolumn{1}{r|}{7B} & 21.4 & 21.7 & 30.5 & 34.3 & 20.3 & 40.5 & 8.8 & 11.8 & 34.9 & 15.0 & 50.0 & 29.3 & 27.1 \\
DeepSeek-Coder-6.7B-Instruct & \multicolumn{1}{r|}{6.7B} & 34.2 & 43.4 & 38.5 & 58.1 & 38.1 & 43.9 & 33.8 & 23.9 & 46.0 & 38.3 & 60.3 & 44.2 & 41.9 \\
CodeQwen1.5-7B-Chat & \multicolumn{1}{r|}{7B} & 36.7 & 44.9 & \fst{46.0} & 51.8 & 29.7 & 40.8 & 26.3 & 24.3 & 42.1 & 41.7 & 48.5 & 44.7 & 40.5 \\
Yi-Coder-9B-Chat & \multicolumn{1}{r|}{9B} & 39.1 & 46.0 & 39.5 & 65.0 & 46.5 & \fst{49.7} & 42.5 & 34.9 & 48.4 & 41.7 & 58.8 & 49.5 & 47.1 \\
Llama-3.1-8B-Instruct & \multicolumn{1}{r|}{8B} & 22.8 & 39.1 & 36.0 & 52.3 & 43.0 & 35.7 & 27.5 & 22.4 & 45.2 & 38.3 & 45.6 & 37.8 & 36.8 \\
OpenCoder-8B-Instruct & \multicolumn{1}{r|}{8B} & 39.5 & 49.1 & 38.0 & 55.6 & 36.0 & 45.9 & 27.5 & 26.5 & 47.6 & 46.7 & 45.6 & 45.7 & 43.6 \\
Qwen2.5-Coder-7B-Instruct & \multicolumn{1}{r|}{7B} & 38.6 & 53.2 & 39.0 & 63.2 & 49.7 & 44.6 & 37.5 & 33.5 & 46.8 & 55.0 & 63.2 & 54.3 & 48.0 \\
Qwen3-8B & \multicolumn{1}{r|}{8B} & 41.6 & 47.4 & 38.0 & 64.0 & 64.0 & 45.2 & 37.5 & 28.7 & 46.8 & 48.3 & 60.3 & 48.9 & 47.7 \\
\graybg \ourmodel-8B-Instruct & \multicolumn{1}{r|}{8B} & \fst{52.8} & \fst{64.1} & 37.0 & \fst{70.6} & \fst{52.8} & \fst{49.7} & \fst{51.3} & \fst{44.1} & \fst{56.3} & \fst{58.3} & \fst{67.6} & \fst{58.5} & \fst{55.8} \\ \midrule
\multicolumn{15}{c}{\grayt 13B+ Models} \\ \midrule
\grayt StarCoder2-15B-Instruct & \multicolumn{1}{r|}{\grayt 15B} & \grayt 38.4 & \grayt 42.2 & \grayt 29.0 & \grayt 59.9 & \grayt 37.1 & \grayt 41.0 & \grayt 42.5 & \grayt 28.7 & \grayt 54.8 & \grayt 33.3 & \grayt 42.7 & \grayt 45.7 & \grayt 41.8 \\
\grayt Codestral-22B & \multicolumn{1}{r|}{\grayt 22B} & \grayt 39.3 & \grayt 46.8 & \grayt 41.5 & \grayt 55.8 & \grayt 42.0 & \grayt 43.4 & \grayt 32.5 & \grayt 30.1 & \grayt 49.2 & \grayt 43.3 & \grayt 45.6 & \grayt 54.8 & \grayt 44.3 \\
\grayt CodeLlama-70B-Instruct & \multicolumn{1}{r|}{\grayt 70B} & \grayt 31.4 & \grayt 33.3 & \grayt 36.0 & \grayt 47.5 & \grayt 34.6 & \grayt 41.3 & \grayt 25.0 & \grayt 36.8 & \grayt 43.7 & \grayt 28.3 & \grayt 36.8 & \grayt 39.4 & \grayt 37.2 \\
\grayt DeepSeek-Coder-33B-Instruct & \multicolumn{1}{r|}{\grayt 33B} & \grayt 38.4 & \grayt 50.6 & \grayt 35.5 & \grayt 66.0 & \grayt 50.0 & \grayt \fst{49.5} & \grayt 43.8 & \grayt 39.7 & \grayt 49.2 & \grayt 53.3 & \grayt 54.4 & \grayt 48.4 & \grayt 48.6 \\
\grayt DeepSeek-Coder-V2-Lite-Instruct & \multicolumn{1}{r|}{\grayt 2.4B/16B} & \grayt 45.8 & \grayt 57.2 & \grayt 38.5 & \grayt 56.9 & \grayt 52.8 & \grayt 44.6 & \grayt 42.5 & \grayt 33.8 & \grayt 52.4 & \grayt 33.3 & \grayt 50.0 & \grayt 51.6 & \grayt 48.7 \\
\grayt DeepSeek-Coder-V2-Instruct & \multicolumn{1}{r|}{\grayt 21B/236B} & \grayt 52.8 & \grayt \fst{63.6} & \grayt \fst{43.0} & \grayt 71.6 & \grayt \fst{75.9} & \grayt 47.5 & \grayt 46.3 & \grayt \fst{52.9} & \grayt 54.0 & \grayt 51.7 & \grayt 63.2 & \grayt 59.6 & \grayt \fst{58.1} \\
\grayt Qwen2.5-Coder-14B-Instruct & \multicolumn{1}{r|}{\grayt 14B} & \grayt \fst{53.3} & \grayt 58.5 & \grayt 41.0 & \grayt 69.5 & \grayt 69.2 & \grayt 46.3 & \grayt 51.3 & \grayt 43.0 & \grayt 49.2 & \grayt 60.0 & \grayt \fst{69.1} & \grayt 57.5 & \grayt 55.3 \\
\grayt Qwen2.5-Coder-32B-Instruct & \multicolumn{1}{r|}{\grayt 32B} & \grayt 51.9 & \grayt 60.9 & \grayt \fst{43.0} & \grayt \fst{73.1} & \grayt 69.9 & \grayt 47.1 & \grayt \fst{55.0} & \grayt 44.9 & \grayt \fst{56.4} & \grayt \fst{61.7} & \grayt 61.8 & \grayt \fst{60.6} & \grayt 56.9 \\
\bottomrule
\end{tabular}
}
    \captionsetup{justification=justified, singlelinecheck=false}
\caption{Performance of instruct models on FullStack Bench, which covers major real-world code development domains including Basic Programming (BP), Advanced Programming (AP), Software Engineering (SE), Data Analysis (DA), Mathematics (MA), Desktop and Web Development (DW), Machine Learning (ML), Scientific Computing (SC), Database (DB), Multimedia (MM), Operating System (OS), and Others.}
\label{tab:instruct_fullstackbench_domain}
\end{table}

\begin{table}[!t]
\centering
\resizebox{\textwidth}{!}{
\begin{tabular}{lrcccccccccccccccc|c}
\toprule
\textbf{Model} & \multicolumn{1}{r|}{\textbf{Size}} & Bash & C++ & C\# & D & Go & HTML & Java & JS & PHP & Python & R & Ruby & Rust & Scala & SQL & TS & \textbf{Overall} \\ \midrule
\multicolumn{19}{c}{\textasciitilde 8B Models} \\ \midrule
CodeLlama-7B-Instruct & \multicolumn{1}{r|}{7B} & 76.7 & 13.1 & 43.9 & 12.0 & 23.5 & 54.4 & 33.0 & 37.2 & 27.8 & 22.0 & 17.5 & 21.1 & 20.0 & 25.0 & 41.3 & 50.0 & 27.1 \\
DeepSeek-Coder-6.7B-Instruct & \multicolumn{1}{r|}{6.7B} & 76.7 & 23.4 & 56.1 & 13.0 & 52.9 & 65.0 & 41.5 & 65.4 & 50.0 & 39.3 & 31.3 & 47.4 & 43.3 & 28.6 & 57.5 & 60.5 & 41.9 \\
CodeQwen1.5-7B-Chat & \multicolumn{1}{r|}{7B} & 73.3 & 30.4 & 41.9 & 15.2 & 48.5 & 57.5 & \fst{44.8} & 60.3 & 44.4 & 37.4 & 15.0 & 52.6 & 58.3 & 46.4 & 52.5 & 60.5 & 40.5 \\
Yi-Coder-9B-Chat & \multicolumn{1}{r|}{9B} & 76.7 & 25.2 & \fst{75.3} & 13.0 & 41.2 & 67.5 & 42.6 & 74.4 & 55.6 & 45.6 & 27.5 & 63.2 & 51.7 & 46.4 & 56.3 & 65.8 & 47.1 \\
Llama-3.1-8B-Instruct & \multicolumn{1}{r|}{8B} & 73.3 & 17.3 & 49.0 & 16.3 & 44.1 & 58.8 & 34.1 & 60.3 & 27.8 & 36.8 & 13.8 & 26.3 & 35.0 & 10.7 & 57.5 & 47.4 & 36.8 \\
OpenCoder-8B-Instruct & \multicolumn{1}{r|}{8B} & 66.7 & 29.9 & 61.6 & 30.4 & 48.5 & 61.3 & 41.3 & 70.5 & 58.3 & 40.1 & 25.0 & 39.5 & 50.0 & 51.8 & 61.3 & 60.5 & 43.6 \\
Qwen2.5-Coder-7B-Instruct & \multicolumn{1}{r|}{7B} & \fst{93.3} & 34.6 & 58.1 & 23.9 & 54.4 & 65.6 & 43.0 & 73.1 & \fst{63.9} & 47.2 & 18.8 & 52.6 & 56.7 & 35.7 & 60.0 & 68.4 & 48.0 \\
Qwen3-8B & \multicolumn{1}{r|}{8B} & 86.7 & 30.4 & 63.1 & 26.1 & 52.9 & 61.9 & 41.5 & 69.2 & 47.2 & 47.4 & 28.8 & 60.5 & 53.3 & 39.3 & 57.5 & 68.4 & 47.7 \\
\graybg \ourmodel-8B-Instruct & \multicolumn{1}{r|}{8B} & 80.0 & \fst{46.3} & 71.2 & \fst{32.6} & \fst{60.3} & \fst{67.5} & 43.7 & \fst{83.3} & 47.2 & \fst{55.5} & \fst{41.3} & \fst{73.7} & \fst{66.7} & \fst{60.7} & \fst{71.3} & \fst{76.3} & \fst{55.8} \\ \midrule
\multicolumn{19}{c}{\grayt 13B+ Models} \\ \midrule
\grayt StarCoder2-15B-Instruct & \multicolumn{1}{r|}{\grayt 15B} & \grayt 56.7 & \grayt 21.0 & \grayt 60.6 & \grayt 29.4 & \grayt 47.1 & \grayt 49.4 & \grayt 31.4 & \grayt 70.5 & \grayt 44.4 & \grayt 42.4 & \grayt 28.8 & \grayt \fst{71.1} & \grayt 35.0 & \grayt 32.1 & \grayt 63.8 & \grayt 52.6 & \grayt 41.8 \\
\grayt Codestral-22B & \multicolumn{1}{r|}{\grayt 22B} & \grayt 60.0 & \grayt 22.0 & \grayt 58.1 & \grayt 6.5 & \grayt 55.9 & \grayt 63.1 & \grayt 41.5 & \grayt 69.2 & \grayt 55.6 & \grayt 42.8 & \grayt 26.3 & \grayt 60.5 & \grayt 58.3 & \grayt 60.7 & \grayt 60.0 & \grayt 60.5 & \grayt 44.3 \\
\grayt CodeLlama-70B-Instruct & \multicolumn{1}{r|}{\grayt 70B} & \grayt 43.3 & \grayt 16.8 & \grayt 62.1 & \grayt 9.8 & \grayt 35.3 & \grayt 58.1 & \grayt 35.8 & \grayt 53.8 & \grayt 47.2 & \grayt 34.9 & \grayt 18.8 & \grayt 52.6 & \grayt 48.3 & \grayt 26.8 & \grayt 55.0 & \grayt 55.3 & \grayt 37.2 \\
\grayt DeepSeek-Coder-33B-Instruct & \multicolumn{1}{r|}{\grayt 33B} & \grayt 60.0 & \grayt 26.6 & \grayt 68.2 & \grayt 19.6 & \grayt 57.4 & \grayt \fst{71.3} & \grayt \fst{44.1} & \grayt 66.7 & \grayt 50.0 & \grayt 48.1 & \grayt 32.5 & \grayt 60.5 & \grayt 50.0 & \grayt 46.4 & \grayt 60.0 & \grayt 57.9 & \grayt 48.6 \\
\grayt DeepSeek-Coder-V2-Lite-Instruct & \multicolumn{1}{r|}{\grayt 2.4B/16B} & \grayt 50.0 & \grayt 36.5 & \grayt 52.5 & \grayt 27.2 & \grayt 61.8 & \grayt 64.4 & \grayt 41.3 & \grayt 73.1 & \grayt 58.3 & \grayt 49.4 & \grayt 38.8 & \grayt 55.3 & \grayt 53.3 & \grayt 42.9 & \grayt 57.5 & \grayt 60.5 & \grayt 48.7 \\
\grayt DeepSeek-Coder-V2-Instruct & \multicolumn{1}{r|}{\grayt 21B/236B} & \grayt 83.3 & \grayt \fst{43.5} & \grayt 72.7 & \grayt 28.3 & \grayt \fst{66.2} & \grayt 69.4 & \grayt 42.4 & \grayt \fst{80.8} & \grayt 61.1 & \grayt \fst{60.4} & \grayt \fst{41.3} & \grayt 65.8 & \grayt 68.3 & \grayt \fst{69.6} & \grayt 66.3 & \grayt \fst{68.4} & \grayt \fst{58.1} \\
\grayt Qwen2.5-Coder-14B-Instruct & \multicolumn{1}{r|}{\grayt 14B} & \grayt \fst{93.3} & \grayt 39.3 & \grayt 65.7 & \grayt 41.3 & \grayt 63.2 & \grayt 66.3 & \grayt 43.5 & \grayt \fst{80.8} & \grayt \fst{77.8} & \grayt 56.2 & \grayt 32.5 & \grayt \fst{71.1} & \grayt 73.3 & \grayt 42.9 & \grayt 62.5 & \grayt \fst{68.4} & \grayt 55.3 \\
\grayt Qwen2.5-Coder-32B-Instruct & \multicolumn{1}{r|}{\grayt 32B} & \grayt 83.3 & \grayt 36.9 & \grayt \fst{76.8} & \grayt \fst{46.7} & \grayt 54.4 & \grayt \fst{71.3} & \grayt 40.4 & \grayt 79.5 & \grayt 58.3 & \grayt 59.1 & \grayt 35.0 & \grayt 63.2 & \grayt \fst{76.7} & \grayt 50.0 & \grayt \fst{70.0} & \grayt 57.9 & \grayt 56.9 \\
\bottomrule
\end{tabular}
}
\caption{Performance of instruct models on FullStack Bench across programming languages.}
\label{tab:instruct_fullstackbench_language}
\end{table}

\noindent \textbf{FullStack Bench.} The limitations in conventional code generation benchmarks (e.g., HumanEval and MBPP) have been further revealed by systematic, tag-based analysis conducted in the recently proposed FullStack Bench~\citep{liu2024fullstackbenchevaluatingllms}. This work samples 500K questions from Stack Overflow\footnote{\url{https://stackoverflow.com/questions}} -- a widely used software development community -- and annotates each question's application domain using LLMs. Subsequently, eleven mainstream code development domains are identified from these annotated questions, including Software Engineering (SE), Data Analysis (DA), Machine Learning (ML), Database (DB), etc., covering 88.1\% of the questions in Stack Overflow. Through this structured tagging framework, conventional benchmarks such as HumanEval and MBPP are demonstrated to narrowly focus on a very limited set of domains, exhibiting inadequate coverage and low diversity across realistic coding scenarios.

\noindent To address the limitations of conventional code generation benchmarks, FullStack Bench is designed to comprehensively evaluate the coding capabilities of LLMs across diverse, real-world scenarios. Employing a meticulous annotation procedure complemented by rigorous cross-validation quality control, FullStack Bench comprises $3,374$ human-annotated coding problems, covering 11 distinct code development domains and 16 programming languages.

\noindent Table~\ref{tab:instruct_fullstackbench_domain} and \ref{tab:instruct_fullstackbench_language} present the results of various instruct models on FullStack Bench across code development domains and programming languages, respectively. As shown in both tables, \ourmodel-8B-Instruct achieves a remarkable 55.8\% pass@1 score, significantly outperforming other \textasciitilde 8B models and even surpassing some state-of-the-art open-source models of larger scale, such as Qwen2.5-Coder-14B-Instruct. On the diverse set of code development domains and programming languages in FullStack Bench, \ourmodel-8B-Instruct consistently achieves top-in-class performance, highlighting its exceptional generalization ability and robustness across various coding scenarios.

\begin{table}[]
\vspace{-5pt}
\centering
\resizebox{0.68\textwidth}{!}{
\begin{tabular}{lrcc}
\toprule
\textbf{Model} & \multicolumn{1}{r|}{\textbf{Size}} & Input-CoT & Output-CoT \\ \midrule
\multicolumn{4}{c}{\textasciitilde 8B Models} \\ \midrule
CodeLlama-7B-Instruct & \multicolumn{1}{r|}{7B} & 36.1 & 36.2 \\
DeepSeek-Coder-6.7B-Instruct & \multicolumn{1}{r|}{6.7B} & 42.6 & 45.1 \\
CodeQwen1.5-7B-Chat & \multicolumn{1}{r|}{7B} & 44.0 & 38.8 \\
Yi-Coder-9B-Chat & \multicolumn{1}{r|}{9B} & 47.5 & 55.6 \\
Llama-3.1-8B-Instruct & \multicolumn{1}{r|}{8B} & 35.6 & 37.8 \\
OpenCoder-8B-Instruct & \multicolumn{1}{r|}{8B} & 39.9 & 43.0 \\
Qwen2.5-Coder-7B-Instruct  & \multicolumn{1}{r|}{7B} & 65.8 & 65.9 \\
Qwen3-8B & \multicolumn{1}{r|}{8B} & \fst{73.8} & \fst{76.9} \\
\graybg \ourmodel-8B-Instruct & \multicolumn{1}{r|}{8B} & 63.3 & 67.1 \\ \midrule
\multicolumn{4}{c}{\grayt 13B+ Models} \\ \midrule
\grayt StarCoder2-15B-Instruct & \multicolumn{1}{r|}{\grayt 15B} & \grayt 45.5 & \grayt 60.9 \\
\grayt Codestral-22B & \multicolumn{1}{r|}{\grayt 22B} & \grayt 61.3 & \grayt 63.5 \\
\grayt CodeLlama-70B-Instruct & \multicolumn{1}{r|}{\grayt 70B} & \grayt 56.5 & \grayt 57.8 \\
\grayt DeepSeek-Coder-33B-Instruct & \multicolumn{1}{r|}{\grayt 33B} & \grayt 47.3 & \grayt 50.6 \\
\grayt DeepSeek-Coder-V2-Lite-Instruct & \multicolumn{1}{r|}{\grayt 2.4B/16B} & \grayt 53.0 & \grayt 52.9 \\
\grayt Qwen2.5-Coder-14B-Instruct & \multicolumn{1}{r|}{\grayt 14B} & \grayt \fst{69.5} & \grayt \fst{79.5} \\
\bottomrule
\end{tabular}
}
\caption{Performance of instruct models on CRUXEval.}
\label{tab:instruct_cruxeval}
\end{table}

\subsubsection{Code Reasoning}

\noindent \textbf{CRUXEval.} As introduced in Sec~\ref{subsubsec:eval_base_codereason}, CRUXEval~\citep{gu2024cruxeval} is designed to measure LLM's code reasoning capability by requiring the model to predict output from a given input (CRUXEval-O), or predict the input from a given output (CRUXEval-I). We further evaluated instruct models using the CRUXEval benchmark, with the results presented in Table~\ref{tab:instruct_cruxeval}. As shown in the table, \ourmodel-8B-Instruct delivers competitive performance in both the Input-CoT and the Output-CoT settings within the group of approximately 8B-parameter models, though lagging behind the recently released Qwen3-8B on this benchmark, indicating that the great potential of smaller-scale LLMs is yet to be explored.

\subsubsection{Code Editing}

\begin{table}[]
\centering
\resizebox{0.65\textwidth}{!}{
\begin{tabular}{lrcc}
\toprule
\multirow{2}{*}{\textbf{Model}} & \multicolumn{1}{r|}{\multirow{2}{*}{\textbf{Size}}} & \multicolumn{1}{c|}{\textbf{Aider}} & \multicolumn{1}{c}{\textbf{CanItEdit}} \\
 & \multicolumn{1}{c|}{} & \multicolumn{1}{c|}{\textit{tries=2}} & \textit{pass@1} \\ \midrule
\multicolumn{4}{c}{\textasciitilde 8B Models} \\ \midrule
CodeLlama-7B-Instruct & \multicolumn{1}{r|}{7B} & \multicolumn{1}{c|}{1.5} & 25.7 \\
DeepSeek-Coder-6.7B-Instruct & \multicolumn{1}{r|}{6.7B} & \multicolumn{1}{c|}{44.4} & 36.9 \\
CodeQwen1.5-7B-Chat & \multicolumn{1}{r|}{7B} & \multicolumn{1}{c|}{38.3} & 34.8 \\
Yi-Coder-9B-Chat & \multicolumn{1}{r|}{9B} & \multicolumn{1}{c|}{54.1} & 50.5 \\
Llama-3.1-8B-Instruct & \multicolumn{1}{r|}{8B} & \multicolumn{1}{c|}{33.1} & 39.5 \\
OpenCoder-8B-Instruct & \multicolumn{1}{r|}{8B} & \multicolumn{1}{c|}{30.8} & 39.0 \\
Qwen2.5-Coder-7B-Instruct & \multicolumn{1}{r|}{7B} & \multicolumn{1}{c|}{\fst{57.9}} & 49.5 \\
Qwen3-8B & \multicolumn{1}{r|}{8B} & \multicolumn{1}{c|}{55.6} & 45.7 \\
\graybg \ourmodel-8B-Instruct & \multicolumn{1}{r|}{8B} & \multicolumn{1}{c|}{57.1} & \fst{50.5} \\ \midrule
\multicolumn{4}{c}{\grayt 13B+ Models} \\ \midrule
\grayt StarCoder2-15B-Instruct & \multicolumn{1}{r|}{\grayt 15B} & \multicolumn{1}{c|}{\grayt 38.2} & \grayt 31.4 \\
\grayt Codestral-22B & \multicolumn{1}{r|}{\grayt 22B} & \multicolumn{1}{c|}{\grayt 51.1} & \grayt 52.4 \\
\grayt CodeLlama-70B-Instruct & \multicolumn{1}{r|}{\grayt 70B} & \multicolumn{1}{c|}{\grayt 15.0} & \grayt 40.5 \\
\grayt DeepSeek-Coder-33B-Instruct & \multicolumn{1}{r|}{\grayt 33B} & \multicolumn{1}{c|}{\grayt 54.5} & \grayt 46.2 \\
\grayt DeepSeek-Coder-V2-Lite-Instruct & \multicolumn{1}{r|}{\grayt 2.4B/16B} & \multicolumn{1}{c|}{\grayt 52.6} & \grayt 45.2 \\
\grayt Qwen2.5-Coder-14B-Instruct & \multicolumn{1}{r|}{\grayt 14B} & \multicolumn{1}{c|}{\grayt \fst{69.2}} & \grayt \fst{52.9} \\
\bottomrule
\end{tabular}
}
\caption{Performance of instruct models on Aider (``whole'' format) and CanItEdit.}
\label{tab:instruct_aider_canitedit}
\end{table}

\noindent \textbf{Aider.} We employed Aider's code editing benchmark\footnote{\url{https://aider.chat/docs/leaderboards/edit.html}} to assess the code editing capability of LLMs. This benchmark is based on a set of 133 coding exercises from Exercism\footnote{\url{https://github.com/exercism/python}}, testing the model by asking it to edit existing code and format the modifications so that all its changes to the source file can be successfully applied without human intervention. The comprehensive design of Aider's coding benchmark captures not only the coding proficiency of LLMs but also their consistency in generating code modifications aligned precisely with prompt specifications.

\noindent Aider uses different edit formats to collect code edits from different LLMs. For a fair comparison, we used the ``whole'' format with a default \textit{tries=2} setting for all evaluations. Table~\ref{tab:instruct_aider_canitedit} outlines the performance of various instruct models on Aider's code editing benchmark. We utilized scores from the official Aider leaderboard whenever available and adopted the reported scores for additional models provided by Qwen2.5-Coder~\citep{qwen25coder}. In cases where discrepancies arise between these two sources (e.g., Qwen2.5-Coder-7B-Instruct self-reported 68.4\%, whereas the official leaderboard lists 57.9\%), we defaulted to the official leaderboard scores. As shown in the table, \ourmodel-8B-Instruct achieves top-tier performance, comparable to Qwen2.5-Coder-7B-Instruct, and clearly outperforms other models of similar size, including Qwen3-8B, which was just released very recently.

\noindent \textbf{CanItEdit.} To assess LLM's proficiency in handling diverse code editing scenarios, we incorporated the CanItEdit benchmark~\citep{cassano2024can} in our code editing evaluations. CanItEdit serves as a benchmark for evaluating the performance of LLMs in instructional code editing, which comprises 105 hand-crafted instructional code editing problems, featuring both descriptive and lazy instructions. As shown in Table~\ref{tab:instruct_aider_canitedit}, \ourmodel-8B-Instruct also takes the lead among all the \textasciitilde 8B models, demonstrating the outstanding code editing capability of our model.

\noindent \textbf{CodeEditorBench.} A recent code editing benchmark CodeEditorBench~\citep{guo2025codeeditorbench} further categorizes the code editing task into four key scenarios: code debugging, code translation, code requirement switching, and code polishing. CodeEditorBench is designed to systematically assess the code editing capability of LLMs across these key scenarios. We employed CodeEditorBench to evaluate various instruct models, with the results presented in Table~\ref{tab:instruct_codeeditorbench}. As shown in the table, \ourmodel-8B-Instruct achieves top-notch performance across the four key code editing scenarios, even surpassing larger state-of-the-art open-source models such as Qwen2.5-Coder-14B-Instruct. The results obtained on these code editing benchmarks establish our model as the best-performing open-source code editing model at \textasciitilde 8B scale.

\begin{table}[]
\centering
\resizebox{0.8\textwidth}{!}{
\begin{tabular}{lrcccc}
\toprule
\textbf{Model} & \multicolumn{1}{r|}{\textbf{Size}} & Debug & \multicolumn{1}{l}{Translate} & \multicolumn{1}{l}{Switch} & Polish \\ \midrule
\multicolumn{6}{c}{\textasciitilde 8B Models} \\ \midrule
CodeLlama-7B-Instruct & \multicolumn{1}{r|}{7B} & 15.5 & 28.9 & 1.7 & 1.2 \\
DeepSeek-Coder-6.7B-Instruct & \multicolumn{1}{r|}{6.7B} & 24.5 & 33.8 & 13.7 & 1.5 \\
CodeQwen1.5-7B-Chat & \multicolumn{1}{r|}{7B} & 21.1 & 36.8 & 11.6 & 1.1 \\
Yi-Coder-9B-Chat & \multicolumn{1}{r|}{9B} & 27.0 & 36.1 & 15.6 & 1.2 \\
Llama-3.1-8B-Instruct & \multicolumn{1}{r|}{8B} & 20.8 & 25.0 & 2.5 & 1.5 \\
OpenCoder-8B-Instruct & \multicolumn{1}{r|}{8B} & 27.2 & 37.9 & 13.5 & 0.7 \\
Qwen2.5-Coder-7B-Instruct & \multicolumn{1}{r|}{7B} & 26.2 & 44.5 & 14.1 & 1.3 \\
Qwen3-8B & \multicolumn{1}{r|}{8B} & 27.9 & 44.9 & 11.2 & 1.5 \\
\graybg \ourmodel-8B-Instruct & \multicolumn{1}{r|}{8B} & \fst{30.7} & \fst{56.5} & \fst{17.7} & \fst{1.9} \\ \midrule
\multicolumn{6}{c}{\grayt 13B+ Models} \\ \midrule
\grayt StarCoder2-15B-Instruct & \multicolumn{1}{r|}{\grayt 15B} & \grayt 19.0 & \grayt 42.0 & \grayt 8.1 & \grayt 1.2 \\
\grayt Codestral-22B & \multicolumn{1}{r|}{\grayt 22B} & \grayt 27.6 & \grayt 39.5 & \grayt 14.1 & \grayt 1.0 \\
\grayt CodeLlama-70B-Instruct & \multicolumn{1}{r|}{\grayt 70B} & \grayt 3.5 & \grayt 21.0 & \grayt 5.5 & \grayt 0.3 \\
\grayt DeepSeek-Coder-33B-Instruct & \multicolumn{1}{r|}{\grayt 33B} & \grayt 27.7 & \grayt 41.4 & \grayt \fst{17.2} & \grayt 1.1 \\
\grayt DeepSeek-Coder-V2-Lite-Instruct & \multicolumn{1}{r|}{\grayt 2.4B/16B} & \grayt 28.0 & \grayt 34.1 & \grayt 15.0 & \grayt 1.1 \\
\grayt Qwen2.5-Coder-14B-Instruct & \multicolumn{1}{r|}{\grayt 14B} & \grayt \fst{29.7} & \grayt \fst{50.8} & \grayt 16.1 & \grayt \fst{1.7} \\
\bottomrule
\end{tabular}
}
\caption{Performance of instruct models on CodeEditorBench.}
\label{tab:instruct_codeeditorbench}
\end{table}

\begin{table}[t]
\centering
\resizebox{0.85\textwidth}{!}{
\begin{tabular}{lr|cc|c}
\toprule
\multirow{2}{*}{\textbf{Model}} & \multicolumn{1}{r}{\multirow{2}{*}{\textbf{Size}}} & \multicolumn{2}{|c}{\textbf{SWE-bench Verified}} & \multicolumn{1}{|c}{\textbf{Multi-SWE-bench mini}} \\ \cmidrule(lr){3-4} \cmidrule(lr){5-5}
 & \multicolumn{1}{c}{} & \multicolumn{1}{|c|}{\textit{Agentless}} & \multicolumn{1}{c}{\textit{OpenHands}} & \multicolumn{1}{|c}{\textit{Agentless}} \\ \midrule
\multicolumn{5}{c}{\textasciitilde 8B Models} \\ \midrule
Yi-Coder-9B-Chat & \multicolumn{1}{r|}{9B} & \multicolumn{1}{c|}{0.0} & 1.6 & 0.0\\
Llama-3.1-8B-Instruct & \multicolumn{1}{r|}{8B} & \multicolumn{1}{c|}{1.0} & 1.2 & 0.5\\
Qwen2.5-Coder-7B-Instruct & \multicolumn{1}{r|}{7B} & \multicolumn{1}{c|}{4.2} & 1.0 & 0.5 \\
Qwen3-8B & \multicolumn{1}{r|}{8B} & \multicolumn{1}{c|}{14.6} & 3.4 & 2.3 \\
Qwen3-8B-Thinking & \multicolumn{1}{r|}{8B} & \multicolumn{1}{c|}{12.5} & 8.6 & 2.5 \\
\graybg \ourmodel-8B-Instruct & \multicolumn{1}{r|}{8B} & \multicolumn{1}{c|}{\fst{19.2}} & \fst{11.2} & \fst{4.0} \\ \midrule
\multicolumn{5}{c}{\grayt 13B+ Models} \\ \midrule
\grayt StarCoder2-15B-Instruct & \multicolumn{1}{r|}{\grayt 15B} & \multicolumn{1}{c|}{\grayt 0.0} & \grayt 0.0 & 0.0 \\
\grayt CodeLlama-34B-Instruct & \multicolumn{1}{r|}{\grayt 34B} & \multicolumn{1}{c|}{\grayt 0.2} & \grayt 0.8 & 0.5 \\
\grayt DeepSeek-Coder-V2-Lite-Instruct & \multicolumn{1}{r|}{\grayt 2.4B/16B} & \multicolumn{1}{c|}{\grayt 4.4} & \grayt 1.0 & 0.5 \\
\grayt Qwen2.5-Coder-14B-Instruct & \multicolumn{1}{r|}{\grayt 14B} & \multicolumn{1}{c|}{\grayt 19.4} & \grayt 1.4 & 3.8 \\
\grayt Qwen2.5-Coder-32B-Instruct & \multicolumn{1}{r|}{\grayt 32B} & \multicolumn{1}{c|}{\grayt \fst{30.2}} & \grayt 5.6 & \fst{4.5} \\
\grayt QwQ-32B & \multicolumn{1}{r|}{\grayt 32B} & \multicolumn{1}{c|}{\grayt 18.4} & \grayt \fst{15.8} & \fst{4.5} \\ \bottomrule
\end{tabular}
}
\caption{Performance of instruct models on SWE-bench Verified and Multi-SWE-bench mini.}
\label{tab:instruct_swebench}
\end{table}

\subsubsection{Software Engineering}
\label{sec:software_engineering}

We evaluated \ourmodel and competitor models on two benchmarks to assess their effectiveness in realistic software engineering scenarios:
\begin{itemize} [leftmargin=*, topsep=0pt, itemsep=5pt]
    \item \textbf{SWE-bench Verified}~\citep{jimenez2023swe} is a benchmark designed to evaluate the ability of LLMs to solve real-world GitHub issues. It consists of $500$ Python instances, each manually verified to ensure the accuracy of both the issue description and the corresponding patch.
    \item \textbf{Multi-SWE-bench}~\citep{zan2025multiswe} is a recently proposed new benchmark, aiming to evaluate the ability of LLMs to resolve issues across multiple programming languages.
    It comprises $1,632$ manually verified instances across $8$ languages: Python, Java, TypeScript, JavaScript, Go, Rust, C, and C++.
    We evaluated on the \emph{mini} version -- a balanced subset of Multi-SWE-bench consisting of $400$ instances, with $50$ instances per language.
    Multi-SWE-bench serves as a practical benchmark for testing the multilingual capabilities of LLMs in real-world software engineering tasks.
\end{itemize}

\noindent Two representative methods of agent scaffolding were used in our evaluation:
\begin{itemize}[leftmargin=*]
    \item \textbf{Agentless}~\citep{agentless} is a method that follows a fixed, manually designed workflow. 
    It decomposes the issue resolving task into pre-defined standard operating procedure consisting of fault localization, code repair, and patch validation.
    \item \textbf{OpenHands}~\citep{openhands} is a fully autonomous agent platform that has no requirements of any fixed workflow.
    It relies entirely on LLM's own planning and reasoning capability, using minimal prompting and external tools to plan, act, and iterate in solving issues.
\end{itemize}

\noindent As shown in Table~\ref{tab:instruct_swebench}, \ourmodel-8B-Instruct achieves the strongest performance among all \textasciitilde 8B models on SWE-bench Verified and Multi-SWE-bench mini, with $19.2\%$ and $4.0\%$ resolved rate under Agentless.
Despite its smaller size, \ourmodel-8B-Instruct matches or exceeds the performance of much larger models such as QwQ-32B ($18.4\%$) and Qwen2.5-Coder-14B-Instruct ($19.4\%$), highlighting its parameter efficiency and strong code generation capabilities.

\noindent We further noticed a consistent pattern:
Agentless outperforms OpenHands in most cases.
This gap is primarily due to the limited capacity of smaller models, which typically rely on pre-defined workflows (Agentless) to perform reliably and often fail when required to operate autonomously (OpenHands).
However, as the model capability continuously improves, agent-based methods like OpenHands are expected to surpass fixed workflows, as reasoning and control flow become increasingly internalized within the model's parameters, enabling more dynamic and adaptive solutions.

\noindent Notably, \ourmodel-8B-Instruct breaks the norm. 
It achieves $11.2\%$ resolved rate under OpenHands, significantly outperforming all the other \textasciitilde 8B baselines, further demonstrating its ability to operate effectively without pre-defined workflow.
On Multi-SWE-bench mini, \ourmodel-8B-Instruct also achieves the highest score among all models of comparable scale. 
This further demonstrates the model's ability to generalize across languages in software engineering scenarios.
We attribute this effectiveness to two key factors:
(1) strong instruction-following capabilities, reinforced by applying LLM filter to ensure consistent formatting across training data;
(2) enhanced software engineering skills, achieved by incorporating commit data during training to support fault localization and code repair.
Overall, \ourmodel delivered strong code generation performance on software engineering tasks, demonstrating the potential of small models to tackle complex, real-world scenarios.

\subsection{Evaluation of Reasoning Models}
To evaluate the effectiveness of our Reasoning model, we conducted experiments on several mainstream and challenging tasks, including LiveCodeBench (2410 -- 2502), the International Olympiad in Informatics\footnote{\url{https://github.com/huggingface/ioi}} (IOI), and Codeforces\footnote{\url{https://codeforces.com}}. For IOI, we adopted the evaluation protocol from Open R1~\citep{openr1}, performing tests on 41 subtasks from the IOI'2024 competition and reporting results based on full submissions. For Codeforces, we followed the scoring methodology and used contest problems consistent with the standardized competition-level code generation benchmark CodeElo~\citep{quan2025codeelo}, implemented within our self-developed submission system, enabling direct comparisons.

\noindent As shown in Table~\ref{reasoning-lcb}, our distilled model DeepSeek-R1-Distill-\ourmodel-8B, trained solely with warmup data, achieves an overall pass@1 score of 39.0\% on LiveCodeBench, surpassing similarly distilled models such as DeepSeek-R1-Distill-Qwen-7B and OlympicCoder-7B. Although we observed that scaling the distillation dataset could lead to further performance improvements, we chose not to further expand the dataset size in order to leave room for reinforcement learning (RL). Extensive distillation would significantly alter the distribution of the base model and might obscure the model's true upper bound capabilities.

\begin{table}[]
\centering
\resizebox{\textwidth}{!}{
\begin{tabular}{l|ccccccccc|c}
\toprule
\multirow{2}{*}{\textbf{Model}} & \multicolumn{3}{c}{Hard} & \multicolumn{3}{c}{Medium} & \multicolumn{3}{c|}{Easy} & \multirow{2}{*}{\textbf{Overall}} \\ \cmidrule(lr){2-4} \cmidrule(lr){5-7} \cmidrule(lr){8-10}
 & \textit{4-mon} & \textit{3-mon} & \textit{2-mon} & \textit{4-mon} & \textit{3-mon} & \textit{2-mon} & \textit{4-mon} & \textit{3-mon} & \textit{2-mon} &  \\
\midrule
\multicolumn{11}{c}{\textasciitilde 8B Models} \\
\midrule
DeepSeek-R1-Distill-Qwen-7B & 11.3 & 10.7 & 9.6 & 39.6 & 37.2 & 37.1 & 76.2 & 77.1 & 67.1 & 36.5 \\
DeepSeek-R1-Distill-Seed-Coder-8B & 13.6 & 13.9 & 13.4 & 39.6 & 38.7 & 39.3 & 79.8 & 80.2 & 73.2 & 39.0 \\
OlympicCoder-7B & 12.7 & 11.8 & 12.5 & 40.8 & 39.0 & 38.7 & 78.0 & 77.1 & 67.8 & 37.9 \\
Qwen3-8B-Thinking & 27.5 & 23.5 & 19.7 & 65.7 & \fst{59.7} & \fst{58.5} & \fst{98.0} & \fst{98.1} & \fst{97.3} & \fst{57.4} \\
\graybg Seed-Coder-8B-Reasoning & \fst{27.6} & \fst{28.0} & \fst{31.0} & \fst{65.8} & 59.2 & 57.5 & 87.8 & 88.0 & 80.1 & 53.6 \\
\midrule
\multicolumn{11}{c}{13B+ Models} \\
\midrule
DeepSeek-R1-Distill-Qwen-14B & 21.3 & 20.5 & 16.1 & 58.1 & 53.4 & 51.4 & 93.3 & 94.2 & 93.7 & 51.9 \\
Claude-3.7-Sonnet-Thinking & 27.3 & 30.8 & \fst{31.0} & 54.5 & 55.1 & 51.4 &  96.2 & \fst{100.0} & \fst{100.0} & 53.3 \\
o3-mini-low & \fst{30.3} & \fst{32.3} & 28.6 & \fst{69.6} & \fst{61.2} & \fst{54.1} &  \fst{98.7} & \fst{100.0} & \fst{100.0} & \fst{59.4} \\
\bottomrule
\end{tabular}
}
    \captionsetup{justification=justified, singlelinecheck=false}
\caption{Performance (pass@1) of reasoning models on LiveCodeBench evaluated over different time windows: 4-mon (2410 -- 2502), 3-mon (2411 -- 2502), and 2-mon (2412 -- 2502).}
\label{reasoning-lcb}
\end{table}

\noindent After RL training, our model \ourmodel-8B-Reasoning markedly improves the overall pass@1 score on LiveCodeBench by 14.6 points. With an overall pass@1 score of 53.6\%, our model surpasses larger models such as DeepSeek-R1-Distill-Qwen-14B, and even slightly outperforms Claude-3.7-Sonnet-Thinking. Notably, during the RL training, our model exhibits substantial pass@1 improvements for medium and hard questions, increasing by 22.8 and 14.0 points, respectively. Furthermore, the scores on hard questions remain stable across different evaluation periods, with a distribution closely matching that of Claude-3.7-Sonnet-Thinking.

\begin{figure}[t]
    \vspace{6pt}
    \centering
    \begin{subfigure}[t]{0.49\textwidth}
        \centering
        \includegraphics[height=5cm]{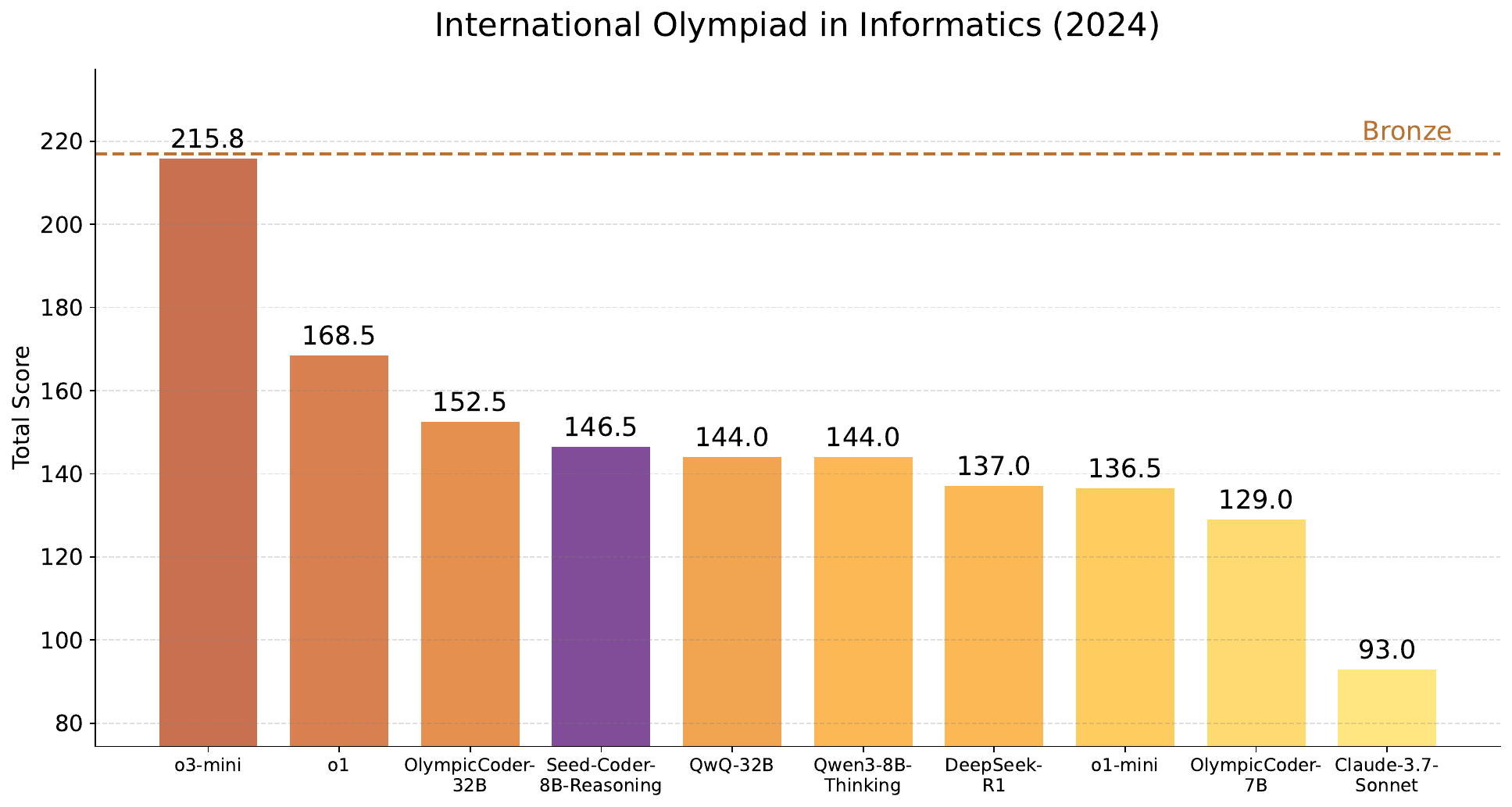}
        \label{fig:fig1}
    \end{subfigure}
    \hfill
    \begin{subfigure}[t]{0.38\textwidth}
        \centering
        \includegraphics[height=5cm]{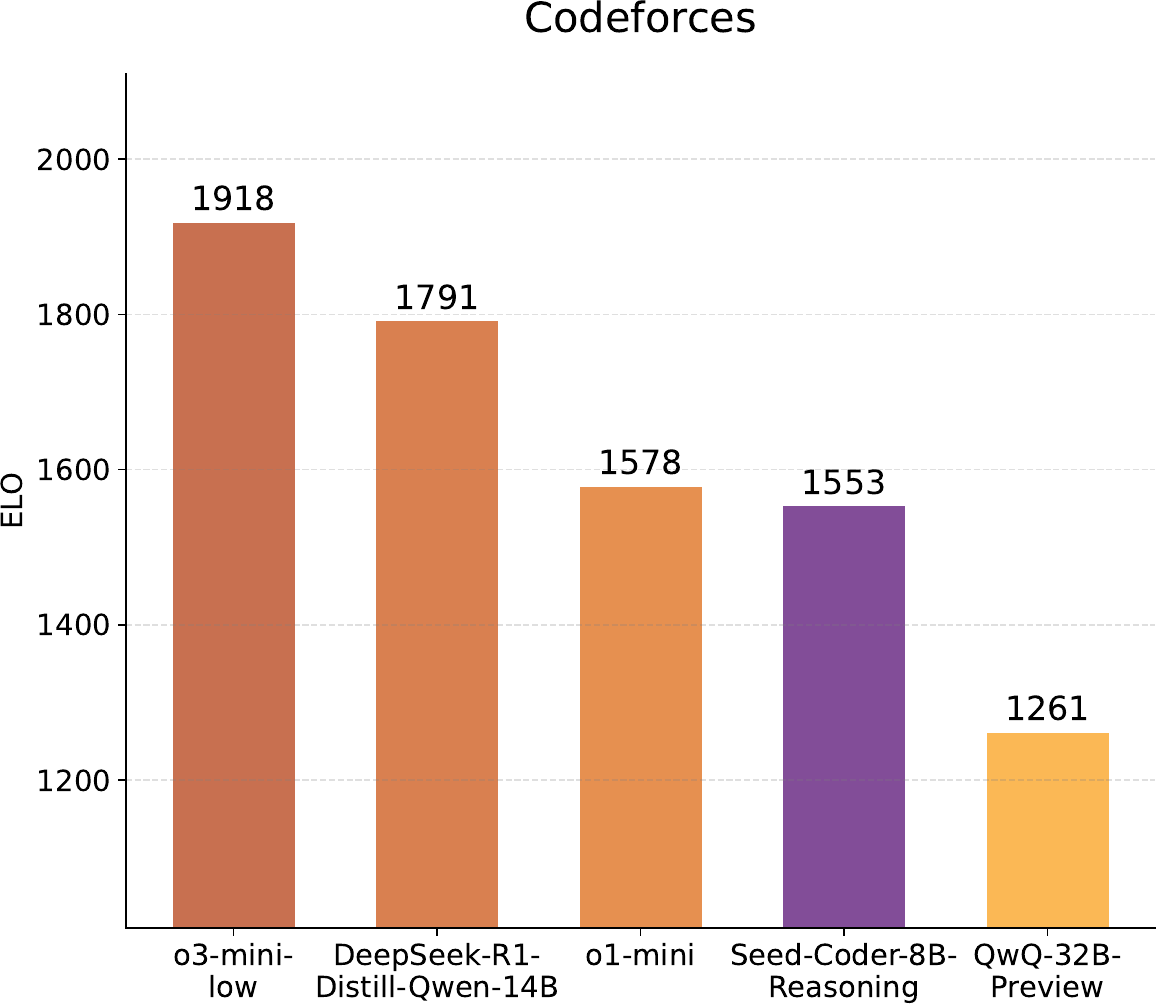}
        \label{fig:fig2}
    \end{subfigure}
    \vspace{-15pt}
    \caption{Performance of reasoning models on IOI'2024 (left) and Codeforces (right).}
    \label{fig:ioi_codeforces}
\end{figure}

\noindent The evaluation results of IOI and Codeforces are shown in Figure~\ref{fig:ioi_codeforces}. On the IOI benchmark, our model achieves higher scores than QwQ-32B and the 671B-sized DeepSeek-R1, also outperforming similarly-sized thinking models. In the Codeforces evaluations, our model scores 1553 points, closely aligning with o1-mini and significantly surpassing the powerful QwQ-32B-Preview model. These results are largely consistent with the IOI rankings. Additionally, we noticed that more than 10\% of the samples were truncated at the 64K sequence length, suggesting a potentially higher performance ceiling of our model.

\noindent However, we also observed notable gaps between the models' scores and the IOI bronze medal threshold, indicating that substantial capability improvements are still required, particularly for solving difficult subtasks where models of similar sizes typically fail to generate correct solutions. Additionally, our model's accuracy on Codeforces problems rated above 2000 points remains very low, and its performance around the 88th percentile still exhibits a significant gap compared to top-tier competitors.

\section{Conclusion, Limitation, and Future Work}

In this report, we introduce \ourmodel, a family of lightweight yet powerful open-source code LLMs achieving state-of-the-art performance across diverse coding benchmarks. We demonstrate that, with minimal human effort, LLMs can effectively curate training data by themselves to significantly enhance code intelligence and reasoning capabilities.

\noindent \ourmodel represents our initial step towards contributing to the open-source LLM community. As \ourmodel primarily focuses on coding tasks and excludes general web data during training, 
its general natural language understanding and ability to handle broader tasks remain limited. Furthermore, the substantial token volume disparity compared to other models (e.g., Qwen3 is pretrained on 36 trillion tokens, while \ourmodel is only pretrained on 6 trillion tokens) stems from insufficient general knowledge and mathematical data, imposing inherent comprehension constraints.
We view this release as the foundation of a growing model family, with future iterations on further improving coding capabilities spanning a range of model sizes.

\bibliography{main.bib}

\clearpage
\newpage

\section{Contributions and Acknowledgments} \label{sec:contributions}

\setlength{\parindent}{0em}

\begin{multicols}{2}
\small
\creditsectionheader{Project Lead}
\corecontributor{Yuyu Zhang}
\\
\creditsectionheader{Core Contributors$^\dagger$}
\corecontributor{Jing Su}
\corecontributor{Yifan Sun}
\corecontributor{Chenguang Xi$^\star$}
\corecontributor{Xia Xiao}
\corecontributor{Yuyu Zhang}
\corecontributor{Shen Zheng}
\\
$^\dagger$ Core contributors sorted alphabetically.
\\
\\
\creditsectionheader{Pretraining}
\creditlistheader{Data Pipeline \& Model Training}
\corecontributor{Yuyu Zhang}
\corecontributor{Yifan Sun}
\corecontributor{Xia Xiao}
\corecontributor{Jing Su}
\corecontributor{Anxiang Zhang$^\star$}
\corecontributor{Chenguang Xi$^\star$}
\\
\creditlistheader{Data Quality Filter}
\corecontributor{Yifan Sun}
\corecontributor{Yuyu Zhang}
\corecontributor{Chenguang Xi$^\star$}
\corecontributor{Kaibo Liu$^\star$}
\\
\creditlistheader{Code-Related Data}
\corecontributor{Jing Su}
\corecontributor{Yuyu Zhang}
\corecontributor{Chenguang Xi$^\star$}
\corecontributor{Daoguang Zan}
\\
\creditlistheader{High-Quality Data}
\corecontributor{Xia Xiao}
\corecontributor{Jing Su}
\corecontributor{Yifan Sun}
\corecontributor{Yuyu Zhang}
\\
\creditlistheader{FIM Training}
\corecontributor{Yifan Sun}
\\
\creditlistheader{Long-Context Training}
\corecontributor{Jing Su}
\corecontributor{Tao Sun}
\columnbreak
\\
\creditsectionheader{Post-Training}
\creditlistheader{Instruct Model}
\corecontributor{Shen Zheng}
\corecontributor{Xia Xiao}
\corecontributor{Jing Su}
\corecontributor{Yifan Sun}
\corecontributor{Yuyu Zhang}
\corecontributor{Kaibo Liu$^\star$}
\\
\creditlistheader{Reasoning Model}
\corecontributor{Xia Xiao}
\corecontributor{Jinhua Zhu$^\star$}
\corecontributor{Shen Zheng}
\corecontributor{Yuyu Zhang}
\\
\creditsectionheader{Evaluation}
\corecontributor{Shulin Xin}
\corecontributor{Kaibo Liu$^\star$}
\corecontributor{Dong Huang}
\corecontributor{Daoguang Zan}
\corecontributor{Shen Zheng}
\corecontributor{Xia Xiao}
\corecontributor{Yifan Sun}
\\
\creditsectionheader{Data Support}
\corecontributor{Yetao Bai}
\corecontributor{Lixin Dong$^\star$}
\corecontributor{Chao Li}
\corecontributor{Jianchong Chen}
\\
\creditsectionheader{Infrastructure}
\corecontributor{Hanzhi Zhou$^\star$}
\corecontributor{Yifan Huang}
\\
\creditsectionheader{Additional Contributors}
\corecontributor{Guanghan Ning$^\star$}
\corecontributor{Xierui Song$^\star$}
\corecontributor{Jiaze Chen}
\corecontributor{Siyao Liu}
\\
\creditsectionheader{Team Management}
\corecontributor{Kai Shen}
\corecontributor{Liang Xiang}
\corecontributor{Yonghui Wu}
\end{multicols}

\vspace{-10pt}
Names marked with $^\star$ denote individuals who have departed from the team.

\newpage
\noindent \textbf{Acknowledgments}

We gratefully acknowledge and thank every Seed-LLM-Code team member not explicitly mentioned above. We also acknowledge and thank every Seed team member for the valuable support. We thank Xuwu Wang, Rui Long, Zihan Wang, Yurong Wu, Aoyan Li, Qi Liu, Xiaojian Zhong, Ran Xin, Wentao Chen, Chen Zheng, Deyi Liu, Yuan Yang, Yiyuan Ma, Ke Sun, Hang Wu, Peng Wu, Runpeng Chen, Tong Wu, Jay Yang, Meiji Wang, Defa Zhu, Yutao Zeng, Ji Li, Xun Zhou, Ke Shen, Pengyang Gao, Haoyuan Guo, Jiacheng Pan, Yunzhe Tao, Shijie Geng, Linyi Li, Liyu Chen, Yite Wang, Boyi Liu, Zhengyu Chen, Yongsheng Xiao, Jianpeng Jiao, Ningyuan Sun, Taifeng Wang, Haibin Lin, and Wenjia Zhu for insightful technical discussions. We thank Huifeng Sun for pointing out the inconsistent BigCodeBench evaluation setting in the previous report version. We have updated the evaluation results with the aligned setting.

\clearpage
\newpage

\appendix

\section{Appendix}
\subsection{Programming Languages in GitHub Data}
\label{appd:supported_langs}
89 programming languages were used in our GitHub Data for pretraining. The full list is as follows:

ANTLR, Ada, Agda, Alloy, AppleScript, Assembly, Augeas, AWK, Batchfile, Bluespec, C, C\#, C++, CMake, CSS, Clojure, CoffeeScript, Common Lisp, CUDA, Dart, Dockerfile, Elixir, Elm, Emacs Lisp, Erlang, F\#, Fortran, GLSL, Go, Groovy, HTML, Haskell, Idris, Isabelle, JSON, Java, Java Server Pages, JavaScript, Julia, Kotlin, Lean, Literate Agda, Literate CoffeeScript, Literate Haskell, Lua, Makefile, Maple, Markdown, Mathematica, MATLAB, OCaml, PHP, Pascal, Perl, PowerShell, Prolog, Protocol Buffer, Python, R, RMarkdown, Racket, Ruby, Rust, SAS, SPARQL, SQL, Scala, Scheme, Shell, Smalltalk, Solidity, Stan, Standard ML, Stata, Swift, SystemVerilog, Tcl, Tcsh, TeX, Thrift, TypeScript, VHDL, Verilog, Visual Basic, XSLT, YAML, Yacc, Zig, reStructuredText.

\subsection{Configuration of Quality Scorer}
\label{appd:pretraining_quality_filter}
In this section, we provide the implementation details of the quality scorer described in Section~\ref{sec:pretraining_quality_filter}.
\subsubsection{Prompt for GitHub Code Quality Scoring}
\label{appd:quality_prompt}
This is the original prompt we used to evaluate the quality of individual code files. It remains consistent throughout the entire pipeline, from collecting ground-truth data to training the quality scorer and applying it across all GitHub data during inference.

\noindent{\footnotesize\texttt{You are an expert of coding. Please carefully evaluate the quality of the \{LANGUAGE\} code file below based on the specific quality criteria essential for its potential use in pretraining a large language model.
\\
Begin your assessment with a brief explanation that addresses the key factors listed below. Following your explanation, assign a numerical rating to the code file on a scale from 1 to 10, where 1 indicates the lowest quality and 10 indicates the highest quality. Please adhere strictly to the following format for your rating: ``Rating: [[X]]'', where X is your numerical rating. Note that the zero score policy should be firstly considered in your analysis, and skip the other criteria if the code meets any zero score conditions.
\\
Criteria for Evaluation:
\\
* Readability:
\\
- Presence of a reasonable amount of comments.
\\
- Inclusion of classes or functions, better with reasonable docstrings that describe the functionality.
\\
- Neat and consistent formatting that adheres to common practice.
\\
- Good naming conventions and well-structured code.
\\
* Modularity:
\\
- Avoidance of overly complicated / very long functions through modularization.
\\
- Clear separation of logic and functionality, using classes and functions.
\\
- Design of each module or component to perform a clear and coherent task.
\\
* Clarity:
\\
- Minimization of excessively repeated code and code blocks, such as repeatedly calling the same function for many times.
\\
- Avoidance of massive commented-out code blocks.
\\
- Avoidance of many random printing statements for debugging.
\\
- Clear communication of intentions behind code blocks.
\\
* Reusability:
\\
- Absence of syntax or logical errors.
\\
- Avoidance of embedding lots of hard-coded data directly within the code.
\\
- Provision of complete and meaningful functionality, not overly simplistic.
\\
- Design that facilitates easy reuse of functions or classes in other projects.
\\
* Zero Score Policy:
\\
- If the code is mostly configurations, such as very long json objects with many numbers or strings, rate 0 score.
\\
- If the code is essentially a data file which includes lots of hard-coded data, such as too many lines of numbers or strings, rate 0 score.
\\
- If the code has little to none effective logic, or is dominated by literals or assignments without any complexity, rate 0 score.
\\
- If the code is auto-generated, with any comments like ``generated by Django'', rate 0 score.
\\
After your analysis, provide your explanation for the aspects evaluated. Then, conclude with the rating in the specified format. For example, if you rate the code quality as 5 out of 10, you should write: ``Rating: [[5]]''.
\{LANGUAGE\} code to be assessed:
\{CONTENT\}.

}
}
\\
\newline The \texttt{LANGUAGE} parameter specifies the primary language of a file, inferred from its extension. The \texttt{CONTENT} parameter contains the file's original content as a string.

\subsubsection{Ground-Truth Quality Scores and Comparison of Oracles}
We collected ground-truth scores from 21 representative programming languages. Table~\ref{tab:quality_gt_lans} shows the number of files for each language.

\begin{table}[h]
\centering
\resizebox{0.65\textwidth}{!}{
\begin{tabular}{lc|lc|lc}
\toprule
Language & Count & Language & Count & Language & Count \\
\midrule
Python & $26,924$ & C\# & $10,112$ & reStructuredText & $9,951$ \\
Shell & $10,194$ & SQL & $10,110$ & C++ & $9,938$ \\
Ruby & $10,187$ & JavaScript & $10,110$ & HTML & $9,422$ \\
Go & $10,185$ & CSS & $10,053$ & R & $9,379$ \\ 
TypeScript & $10,179$ & Kotlin & $10,051$ & TeX & $9,156$ \\
MATLAB & $10,165$ & PHP & $10,049$ & Markdown & $8,845$ \\
Java & $10,161$ & C & $10,020$ & R Markdown & $6,875$ \\
\bottomrule
\end{tabular}
}
\caption{Language distribution of the quality scorer training set.}
\label{tab:quality_gt_lans}
\end{table}

\begin{figure}[]
    \centering
    \includegraphics[width=0.48\textwidth]{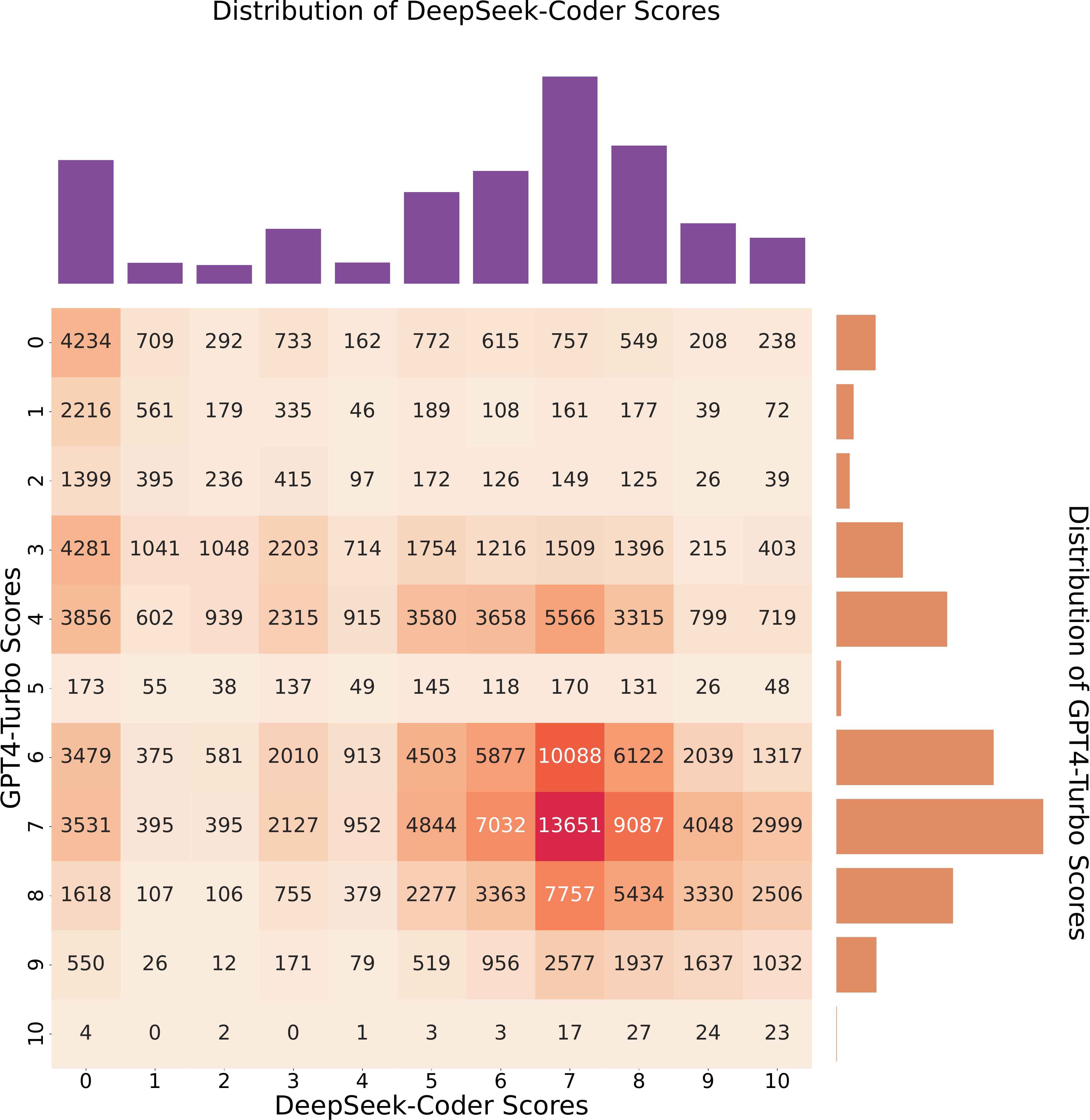}
    \hfill
    \includegraphics[width=0.48\textwidth]{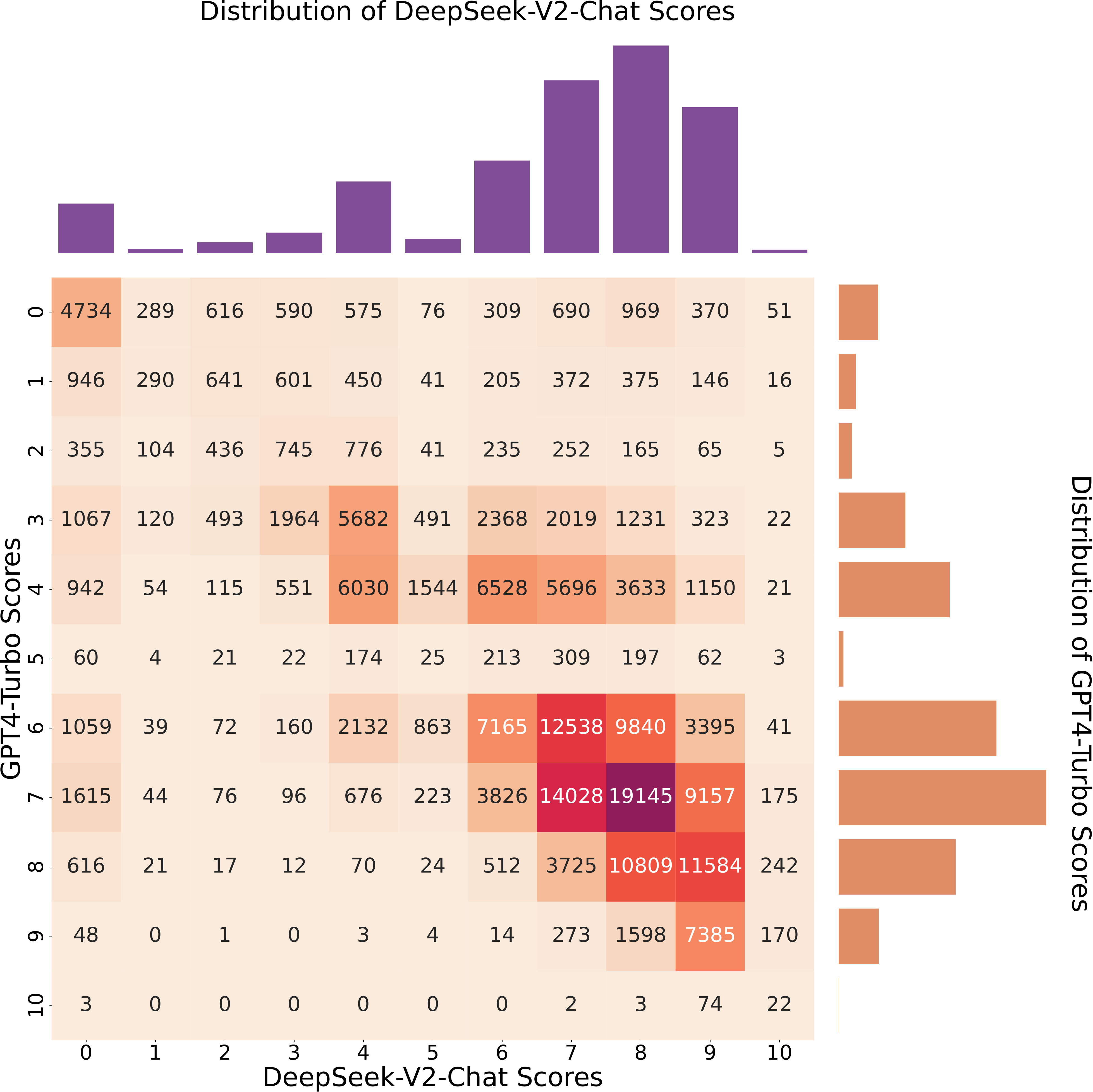}
        \captionsetup{justification=justified, singlelinecheck=false}
    \caption{Comparison of quality scores from three oracles. Score distributions from individual oracles are presented along the top and right edges of the heatmap.}
    \label{fig:oracle_comparison}
\end{figure}

Figure~\ref{fig:oracle_comparison} compares the ground-truth scores from the responses of three oracles. We used GPT-4 Turbo as the baseline, comparing it with DeepSeek-Coder-33B and DeepSeek-V2-Chat. We observed that despite minor discrepancies in determining whether a file meets the zero score conditions, all models produced consistent scores, both on a per-sample basis and in the overall distribution. In practice, we used DeepSeek-V2-Chat as the oracle for training the quality scorer, balancing efficiency and accuracy.

\begin{figure}[!h]
    \centering
    \includegraphics[width=\textwidth]{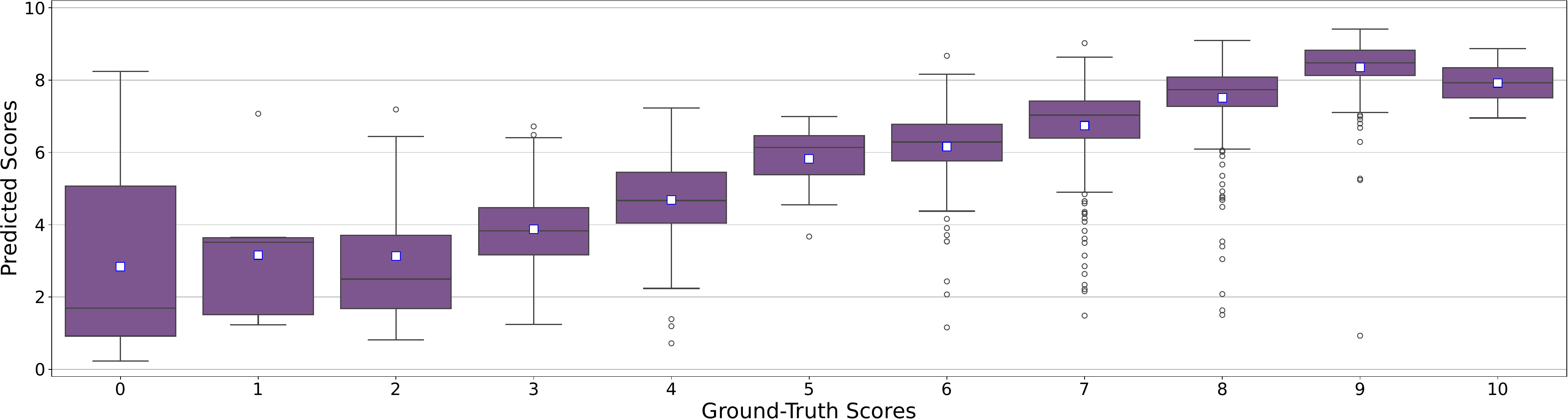}
        \captionsetup{justification=justified, singlelinecheck=false}
    \caption{Boxplot of predicted scores from the quality filter corresponding to each individual ground-truth score. Means are indicated with small white squares in the bars.}
    \label{fig:quality_scorer}
\end{figure}

\subsubsection{Evaluation of Quality Scorer}
The quality scorer is designed to filter out low-quality code files from the dataset. Once a threshold is set, the scorer effectively functions as a classifier. To ensure flexibility in threshold selection, the quality scorer must provide accurate predictions across arbitrary score intervals. During training, we used mean squared error (MSE) between predicted scores and ground-truth values as the loss function for the regression model. However, as shown in Figure~\ref{fig:oracle_comparison}, the ground-truth scores are not uniformly distributed, potentially leading to varying accuracy across different intervals under MSE-loss. To further assess the quality filter, we evaluated the mean absolute error (MAE) for samples at each ground-truth score. Figure~\ref{fig:quality_scorer} shows the predictions of the quality scorer over $1,178$ test samples, categorized by the ground-truth scores from DeepSeek-V2-Chat. While identifying zero-condition cases remains challenging, the overall MAE across categories is $\epsilon_{cMAE}=1.37$. In contrast, the MAE computed directly across all samples is $\epsilon_{MAE}=0.91$. The formulations for computing $\epsilon_{cMAE}$ and $\epsilon_{MAE}$ are
\begin{equation*}
\begin{split}
    \epsilon_{cMAE} = \frac{1}{11}\sum_{i=0}^{10}\left(\frac{1}{|C_i|}\sum_{j\in C_i}|\hat y_j - i|\right), \quad \epsilon_{MAE} = \frac{\sum_{i=0}^{10}\sum_{j\in C_i}|\hat y_j - i|}{\sum_{i=0}^{10} |C_i|},
\end{split}
\end{equation*}
where $C_i$ is the subset of samples with ground-truth score $i$ and $\hat y_j$ is the predicted fractional score of sample $j$. Both metrics indicate that the scorer performs reliably.

\end{document}